\documentclass{article}

\usepackage{PRIMEarxiv}
\usepackage[utf8]{inputenc} 
\usepackage[T1]{fontenc}    
\usepackage{hyperref}       
\usepackage{url}            
\usepackage{booktabs}       
\usepackage{amsfonts}       
\usepackage{nicefrac}       
\usepackage{microtype}      
\usepackage{lipsum}
\usepackage{fancyhdr}       
\usepackage{graphicx}       
\graphicspath{{media/}}     
\usepackage{svg}
\usepackage{authblk}
\usepackage[paper=portrait,pagesize]{typearea}
\usepackage{multirow} 
\usepackage{makecell}

\usepackage{geometry}
\usepackage{multirow}
\usepackage{booktabs}
\usepackage{makecell}
\usepackage{pdflscape}

\usepackage{svg}
\usepackage{enumitem}
\usepackage{caption}
\usepackage{bm}

\usepackage{float}
\usepackage{amsmath}
\usepackage{amssymb}

\usepackage{xcolor,pifont}
\newcommand*\colourcheck[1]{%
  \expandafter\newcommand\csname #1check\endcsname{\textcolor{#1}{\ding{52}}}%
}
\colourcheck{blue}
\colourcheck{green}
\colourcheck{red}
\colourcheck{orange}

\usepackage{array}
\usepackage{ragged2e}

\usepackage{subcaption} 

\usepackage{titlesec}
\titlespacing{\section}{0pt}{\parskip}{-\parskip}
\titlespacing{\subsection}{0pt}{\parskip}{-\parskip}
\titlespacing{\subsubsection}{0pt}{\parskip}{-\parskip}

\usepackage{tcolorbox}
\usepackage{enumitem}

\usepackage{amsmath}
\usepackage{algorithm}
\usepackage{algpseudocode}
\usepackage{amsfonts}
\usepackage{amsmath}
\usepackage{graphicx}
\usepackage{hyperref}

\usepackage{graphicx}
\DeclareGraphicsExtensions{.png}
\pdfpagewidth=\paperwidth
\pdfpageheight=\paperheight

\usepackage{graphicx}
\usepackage{float}
\usepackage{caption}

\usepackage{graphicx,float,caption}

\usepackage{authblk}
\usepackage{titlesec}
\titlelabel{\thetitle.\quad}
\title{Large Language Models for Healthcare Text Classification: A Systematic Review}

\author[ ]{\textbf{Hajar Sakai} and \textbf{Sarah S. Lam}}
\affil[ ]{School of Systems Science and Industrial Engineering}
\affil[ ]{State University of University at Binghamton}
\affil[ ]{Binghamton, NY, USA}
\affil[ ]{hsakai1, sarahlam@binghamton.edu}

\begin{document}
\maketitle

\begin{abstract}
Large Language Models (LLMs) have fundamentally transformed approaches to Natural Language Processing (NLP) tasks across diverse domains. In healthcare, accurate and cost-efficient text classification is crucial, whether for clinical notes analysis, diagnosis coding, or any other task, and LLMs present promising potential. Text classification has always faced multiple challenges, including manual annotation for training, handling imbalanced data, and developing scalable approaches. With healthcare, additional challenges are added, particularly the critical need to preserve patients’ data privacy and the complexity of the medical terminology. Numerous studies have been conducted to leverage LLMs for automated healthcare text classification and contrast the results with existing machine learning-based methods where embedding, annotation, and training are traditionally required. Existing systematic reviews about LLMs either do not specialize in text classification or do not focus on the healthcare domain. This research synthesizes and critically evaluates the current evidence found in the literature regarding the use of LLMs for text classification in a healthcare setting. Major databases (e.g., Google Scholar, Scopus, PubMed, Science Direct) and other resources were queried, which focused on the papers published between 2018 and 2024 within the framework of PRISMA guidelines, which resulted in 65 eligible research articles. These were categorized by text classification type (e.g., binary classification, multi-label classification), application (e.g., clinical decision support, public health and opinion analysis), methodology, type of healthcare text, and metrics used for evaluation and validation. This review reveals the existing gaps in the literature and suggests future research lines that can be investigated and explored.
\end{abstract}

\keywords{Large Language Models \and Text Classification \and Healthcare \and Natural Language Processing \and Neural Networks \and Systematic Review}

\section{Introduction}
Large Language Models (LLMs) are currently a major tech trend that nearly everyone can access. Their landscape is expanding rapidly, with new models introduced at an elevated pace. These models vary between open-source and closed-source, general-purpose and domain-specific, with some designed to be multilingual and/or multimodal. Certain foundation LLMs are also fine-tuned to solve specific tasks while targeting selected domains. Organizations, in addition to individuals, are perpetually exploring how to efficiently leverage these LLMs to extract insights and relevant information that could advance the decision-making processes from the continuously accumulating text data, and the healthcare industry is no exception. In this section, language modeling is reviewed, the importance of text classification is discussed, and the research objectives and structure are outlined.

\subsection{Language Modeling}
Language modeling involves building a mathematical model with statistical probabilities that represent the structure and sequence of tokens/words. The currently most widely used Language Models (LMs) are autoregressive, where the distribution over tokens is decomposed into conditional probabilities; in other words, these models can predict the next token given the past provided context and generated tokens. Language models are characterized by having both syntactic and semantic knowledge. The corresponding research of language modeling has evolved through four main generations. A specific task-solving capacity characterizes each generation, where the usefulness of the developed language models grows over time (Zhao et al., 2023). 
\\[0.25cm]
In the 90s, Statistical Language Models (SLM) were introduced. These models rely on the Markov assumption, stating that the probability of a word only depends on the previous words representing the context. The frequency and co-occurrence of words in a large corpus are used to calculate the probability of a word sequence. One famous case of these models is when the context length is fixed, and these models are the N-gram Language Models. These models, however, face challenges due to data sparsity occurring in the case of rare word sequences. Additionally, they suffer from their inability to capture long-term dependencies since the context length is fixed; moreover, when this size is large, the computed probability accuracy degrades, given the significantly large number of transition probabilities that need to be calculated. SLM applications are limited to assist and improve some Natural Language Processing (NLP) tasks. 
\\[0.25cm]
Around 2013, Neural Language Models (NLMs) were presented. Neural networks are used for word sequence probability calculations. These models aim to deal with natural language by learning word representations as continuous vectors. The representation’s learning captures the latent patterns and semantics in addition to long-term dependencies and, therefore, overcomes the challenges SLM suffers from. Word2vec is one of the most popular NLMs, introduced in 2013, and is commonly used to learn features from text. However, since these models are deep learning-based, the interpretability and explainability of the representation's vectors remain challenging. Besides, these models are computationally intensive and require large corpora for training. It is also worth mentioning that these representations that are learned are static. NLM applications consist of solving a range of standard NLP tasks. 
\\[0.25cm]
At the beginning of 2018, Pre-trained Language Models (PLMs) were unveiled. Given that NLMs provide static word representations, a strive towards learning context-aware word representations using PLMs was noticed. This was conducted by pre-training models (e.g., biLSTM for the case of ELMo) to generate task-agnostic representations followed by task-specific fine-tuning. As a result, a new learning paradigm was developed: ‘pre-training and fine-tuning'. BERT and GPTs 1 and 2 are regarded as part of the range of PLMs. Since these PLMs' pre-training comes with more computation requirements and a larger data scale, their limitations comprise the significant demand for computational resources for pre-training and fine-tuning. Additionally, these representations are highly dependent on the corpora used for training. Therefore, the collected data might introduce bias to the model’s outputs. These models, among others, demonstrated their strong performance when applied to multiple NLP tasks. 
\\[0.25cm]
In 2020, research around scaling the previously introduced PLMs, both at the model’s and data’s sizes levels, peaked with the release of Large Language Models (LLMs) such as GPT-3 (Brown et al., 2020) and LaMDA (Thoppilan et al., 2022), and later PaLM (Touvron et al., 2023) and LLaMA (Chowdhery et al., 2023). LLMs stand out from other language models by four attributes. First, LLMs are trained on significantly large corpora. Second, the LLM’s architecture size is substantially larger than the one corresponding to a PLM, which increases the number of parameters usually counted now in billions. Third, LLMs provide prompt-based completion, which makes them more accessible and intuitive, especially with applications such as ChatGPT, which was announced by OpenAI in 2022, enabling a conversational interaction with natural language (OpenAI, 2022). Fourth, LLMs not only serve as helpers in solving some tasks such as SLMs but also can solve multiple real-world tasks. The latest generation of LLMs is now considered “general purpose task solvers” and can be used for multiple downstream traditional NLP tasks such as text classification.

\begin{figure}[H]
    \centering
    \includegraphics[scale=0.75]{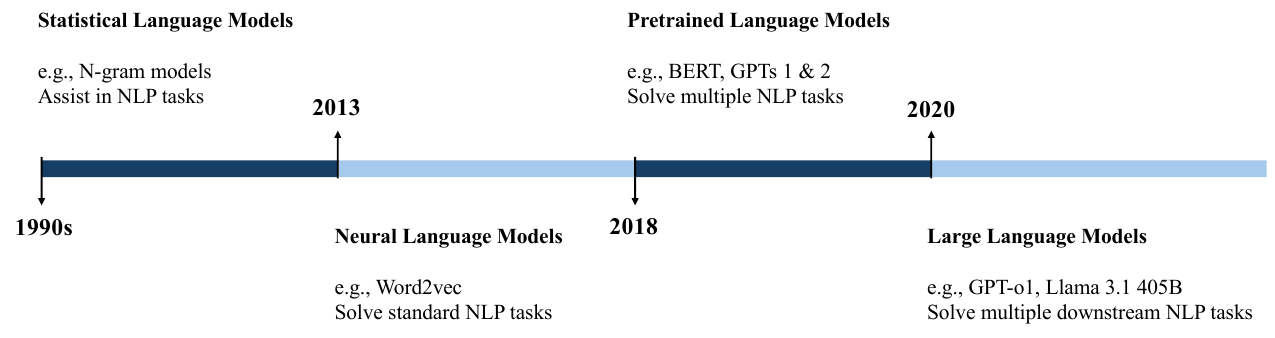}
    \caption*{\textbf{Figure 1.} The Evolution of Language Modeling}
    \label{fig:mltc-approaches}
\end{figure}

\newpage

\subsection{Large Language Models (LLMs)}
LLMs are deep neural networks, and most modern ones are transformer-based. They are trained on large amounts of text data and demonstrate an impressive ability to interact with and generate human-like natural language. These language models are large in terms of architecture, the number of parameters involved, and the size of pre-training data. The advent of the transformer is the basis of the revolution that NLP is witnessing, where the attention mechanism (Vaswani et al., 2017) incorporated in its architecture enables the LLM to focus on different parts of the text input when generating each part of the output. This is done by weighting the relative importance of each token in a text sequence. The transformer architecture was originally developed for machine translation and comprises two components: an encoder and a decoder, as shown in Figure 2. Both components consist of multi-head self-attention and Feedforward Neural Networks (FNN) layers. Two key LMs, BERT and GPT, are also based on variants of the transformer's architecture, where the former is built on the encoder component and trained for masked word prediction. While, the latter is founded on the decoder component and trained for text generation, one word at a time (Raschka, 2024).

\begin{figure}[H]
    \centering
    \includegraphics[scale=0.6]{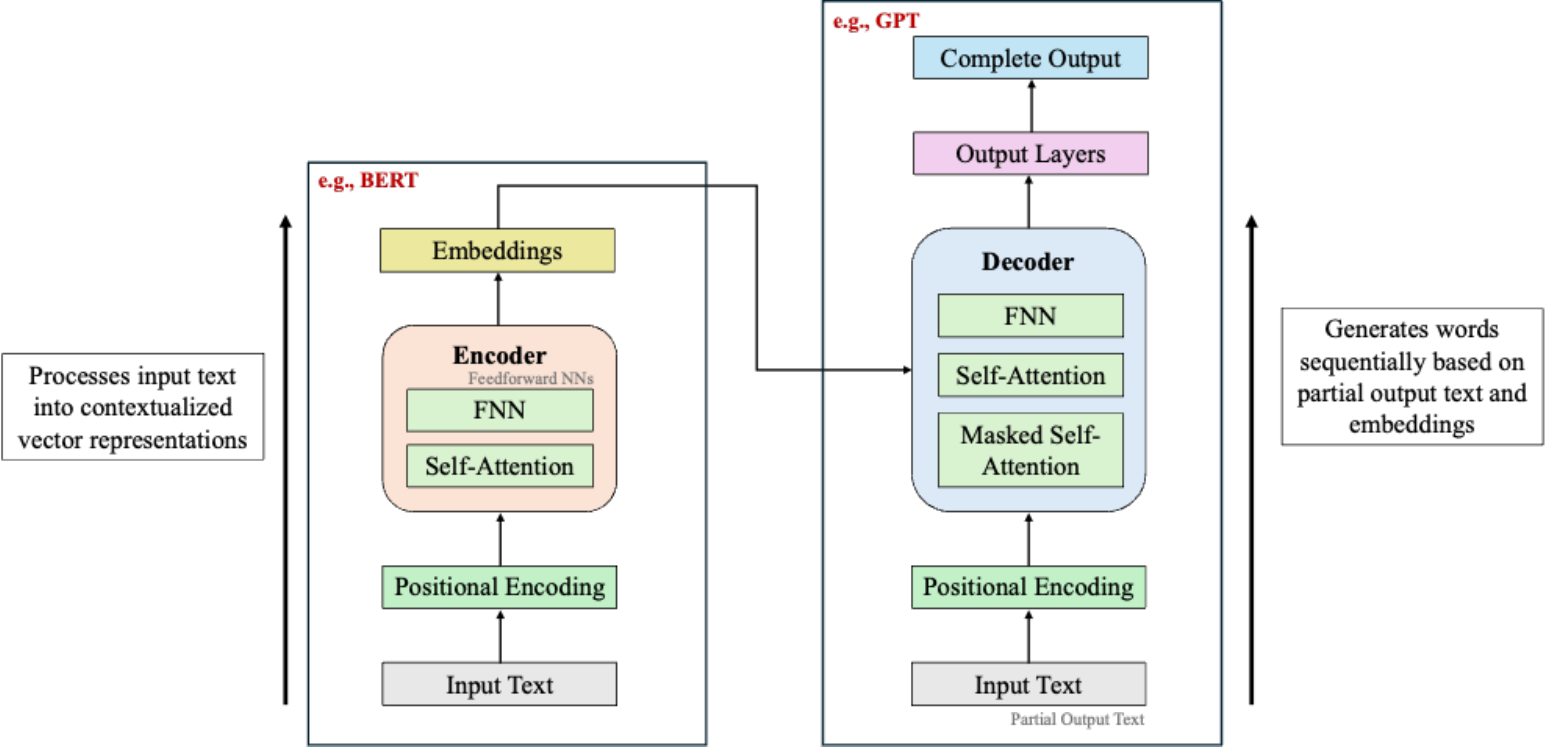}
    \caption*{\textbf{Figure 2.} High-level Transformer Architecture with BERT and GPT as examples}
    \label{fig:mltc-approaches}
\end{figure}

Developing an LLM from scratch is highly resource-intensive, which demands immense computational power, large datasets (where annotation is required for fine-tuning), and important financial investments. These resources are required for data sampling, pre-training the foundation model, and fine-tuning it for specific tasks such as text classification (Raschka, 2024). Consequently, numerous studies have chosen to leverage existing LLMs, which, although not originally designed for text classification, have shown the ability to carry out this task effectively, particularly in healthcare settings, where the resources for developing a tailored foundation model are usually limited.  
\\[0.25cm]
It is worth noting that in this systematic review, PLMs like BERT and its variants would be referred to as LLMs. 

\subsection{Text Classification}
\subsubsection{Motivation}
The expansion of the internet and complete digitalization of various domains, such as healthcare, triggered the continuous generation of textual data, and as a result, vast piles of textual data are accumulated. This presents text data management and analysis as not only challenges but also interesting research opportunities. The need to proficiently organize textual data and transform it into a structured format is guiding the research focus to dig deeper into the possibility of automatically assigning predefined categories to text, thus improving access to information through NLP tasks like text classification, named entity recognition (Ghali et al., 2024), and others.
 \\[0.25cm]
There are various motivations behind the research revolving around efficiently categorizing collected text data, especially since text classification constitutes a foundation task for multiple NLP applications. Sebastiani (2002) highlights its applications to include text indexing for Boolean Information Retrieval, document categorization, and filtering, in addition to Word Sense Disambiguation (WSD). In an information retrieval system, the text indexing is based on a controllable dictionary where documents are paired with one keyword or more. Considering the ensemble of vocabulary contained in the controllable dictionary as categories, text indexing, which facilitates information retrieval, can be viewed as text classification. Additionally, this results in an automated categorization of the documents considered, which can also be extended to web pages, and using the assigned labels, text filtering is permitted. Besides, WSD is another application of text classification, where the categories here are the sense of the ambiguous word given the context in which it appears. This can be particularly useful for other NLP tasks such as Machine Translation. Furthermore, text classification is useful for understanding and enhancing customer or user experience through Sentiment Analysis (SA) (Pang et al., 2002). SA is a special case of text classification where the categories considered are the sentiments (e.g., positive, negative, mixed, neutral). This spotlights the potential practical applications of text classification, which would consist of customer feedback analysis and understanding. Moreover, spam detection is another motivation behind developing adequate text classification techniques for email classification (Dada et al., 2019). In the same context, content moderation also relies on text classification (Nobata et al., 2016), which is an additional driver fueling this research. 
\\[0.25cm]
In healthcare, text classification is gaining importance, especially since medical textual data is growing at an exponential rate. Text classification automatically extracts valuable insights from various types of continuously generated healthcare narratives, such as clinical notes, patients’ feedback, and medical research papers. This is particularly crucial since it would improve patient care by exploiting available data that is usually challenging to extract, which results in more efficient healthcare systems (Spasic and Nenadic, 2020). There are diverse healthcare NLP tasks to which text classification can significantly contribute by providing clinical decision support. Examples include automated diagnosis coding (Karimi et al., 2017) and identifying at-risk patients for certain diseases (Bittar et al., 2019). These text classifications would be useful for early intervention and anticipated treatment planning. With the evolution of machine learning techniques applied to text classification, especially deep learning methods, there is promising potential to achieve accuracy rates comparable to human expert annotations without constant human intervention. 

\subsubsection{Text Classification Over Time}
Over the decades, the literature has witnessed various stages of the methodologies deployed to conduct text classification (Sakai et al. (2023a), Sakai et al. (2024a),  Sakai et al. (2024b), Sakai and Lam (2025)). It started with developing rule-based approaches (Apté et al., 1994) where dictionaries and rules were manually crafted and, therefore, lacked scalability. This was followed by statistical and machine learning techniques such as Multinomial Naive Bayes classifier (McCallum and Nigam, 1998), Support Vector Machines (Joachims, 1998), Decision Tree algorithm (Lewis and Ringuette, 1994), and ensemble techniques such as Random Forest (Xu et al, 2012). It is important to highlight that applying machine learning models for text classification would not have been enabled without the parallel development of representation learning and feature engineering techniques. These methods have, in turn, evolved from the simple Bag-of-Words technique and its extension N-Gram to Term Frequency and Inverse Document Frequency (TF-IDF) and later learning embedding methods comprising words’ semantic and syntactic relationships such as Word2Vec or capturing the analogies and linear relationships between words such as  GloVe (Rani et al., 2022). These methods resulted in the vectorization of text data to dense representations and, therefore, unlocked the potential of successfully applying neural networks and deep learning models for text classification. For instance, Socher et al. (2011) fitted a Feed-Forward Network (FNN), a recursive autoencoder, for binary polarity classification. A few years later, Zhang and LeCun (2015) showed that Convolutional Neural Networks (CNNs) can also be effectively used for text classification. Since textual data is categorized as sequential, Recurrent Neural Networks were also deployed for text classification (Lai et al., 2015). Subsequently, attention mechanisms were introduced, which enabled the models to detect the relevant parts of the embeddings. This naturally led to more research on the possibility of using transformers for text classification and resulted in the development of two of the most popular language models: BERT (Devlin et al., 2018) and GPT (Radford et al., 2018). Transfer learning is used in these cases to conduct text classification. As previously mentioned, the scaling of these PLMs led to the introduction of LLMs, and these models revolutionized NLP tasks, at the top of them, text classification.

\subsection{Research Objectives and Structure}
The evolution of LMs from assistants to self-sufficient models capable of conducting a wide range of NLP tasks redefines how textual data is mined. During the last couple of years, several systematic reviews have summarized and discussed research studies using LLMs in healthcare without focusing on specific tasks such as text classification (Li et al., 2024a; Busch et al., 2024; Wang et al., 2023; Sallam, 2023). In parallel, a few systematic reviews have explored current trends in healthcare text classification using NLP and machine learning techniques (Kesiku et al., 2022; Hossain et al., 2023). These systematic reviews either addressed the general impact of LLMs on healthcare-related tasks or targeted all the machine learning-based approaches used by researchers for healthcare text classification. 
\\[0.25cm]
LLMs have revolutionized healthcare text classification, which promises more cost-efficient and less time-consuming methodologies and potentially ensures accurate categorizations. Therefore, a systematic review is needed to combine the research studies where LLMs were leveraged for text classification in healthcare settings. This systematic review examines literature published in the last six years where LLMs and PLMs were developed, used as-is, or fine-tuned to address any type of healthcare text classification. The collected research studies are categorized, analyzed, and discussed, with the results revealing future research directions.
\\[0.25cm]
The remainder of this systematic review is organized as follows: Section 2 details the research methodology, which includes the paper collection process and inclusion/exclusion criteria. Section 3 presents the results of the surveyed literature, which focuses on text classification types, LLM-based methodologies, healthcare text types, and evaluation metrics. Section 4 discusses the results and highlights the current research gaps and limitations. Section 5 explores future research directions. Section 6 concludes this systematic research that focuses on healthcare text classification using LLMs. 

\section{Research Methodology}
This systematic review follows the Preferred Reporting Items for Systematic Reviews and Meta-Analyses (PRISMA) guidelines for its research methodology (Moher et al., 2009).

\subsection{Search Strategy}
This review paper considers five major databases for paper collection: Google Scholar, Scopus, Science Direct, Web of Science, and PubMed. It also includes references from collected papers and another available literature search engine (i.e., Elicit).  The peer-reviewed articles and conference proceedings surveyed were published between 2018 and 2024. The paper search was conducted between March and September 2024. Keywords used include ‘Large Language Models’, ‘LLMs’, ‘Healthcare’, ‘Medical’, ‘Bio’, ‘Biomedical’, ‘Classification’, ‘Text Classification’, ‘Text Categorization”, and ‘Sentiment Analysis’, while excluding terms such as ‘Image Classification’ and ‘Survey’. Besides, only papers written in English were considered. This systematic review focuses on research studies that involve healthcare text categorization and evaluation using LLM-based approaches.

\subsection{Inclusion and Exclusion Criteria}
The inclusion criteria for this systematic review are as follows: research studies (1) published in English; (2) within the last six years; (3) in peer-reviewed journals or conference proceedings; and (4) that utilized at least one PLM or LLM at any stage of the text classification, regardless of type. Acceptable models include BERT and its variants, open-source research LLMs such as Llama 2, or closed-source models such as GPT-3.5/4. GPT-based models could be used either through API requests or via ChatGPT, the inference model. The review considered various text classification types (i.e., binary classification, multi-class classification, and multi-label classification) and applications (e.g., clinical decision support, research/literature analysis, public health and opinion analysis, patient query analysis). Healthcare textual data sources vary, for instance, from clinical notes and discharge summaries to patient comments and medical literature. Methodologically, the included studies may have either used LLMs for direct classification, addressing the class imbalance, data augmentation, fine-tuned the model for specific healthcare text classification tasks, or pre-trained one. Furthermore, research papers where one or more healthcare datasets, among others, were used for evaluation are also included. 
\\[0.25cm]
The exclusion criteria of the research papers surveyed include cases where (1) the full text of the paper was inaccessible; (2) the focus is on NLP tasks other than classification (e.g., translation, extraction); (3) the evaluation approach consists of using synthetically generated textual data; (4) the manuscript is a thesis, dissertation, workshop preface, editorial, letter to the editor, seminal contribution, comment, or review; (5) the paper is primarily introducing a dataset or a library; (6) the paper deals solely with a non-healthcare application (e.g., finance, legal, social science) or focused on LLM-generated text detection; (7) the classification conducted is for non-textual data (e.g., images, time series); and (8) the study uses only traditional machine learning approaches without pre-training (i.e., simple text vectorization, model training, and classification) without involving an LLM at any level.

\section{Results}
\subsection{PRISMA Process}
At the ‘Identification’ stage, 826 research papers were collected, among which 257 studies from Google Scholar, 221 from Scopus, 174 from ScienceDirect, 44 from Web of Science, 29 from PubMed, and 101 from other resources that include some research surveys references and paper found through another available search engine. Of this batch of research papers, 127 were dropped because they were duplicates, and 156 were excluded following the aforementioned exclusion criteria. Moving forward to the ‘Screening’ phase, 405 studies were excluded based on the title and abstract review. As a result, 138 research papers were retained and assessed for ‘Eligibility’ based on the full-text review. Out of these studies, 73 were judged ineligible, resulting in 65 research papers ‘Included’ and carefully examined in this systematic review, as shown in Figure 3.

\begin{figure}[H]
    \centering
    \includegraphics[scale=1]{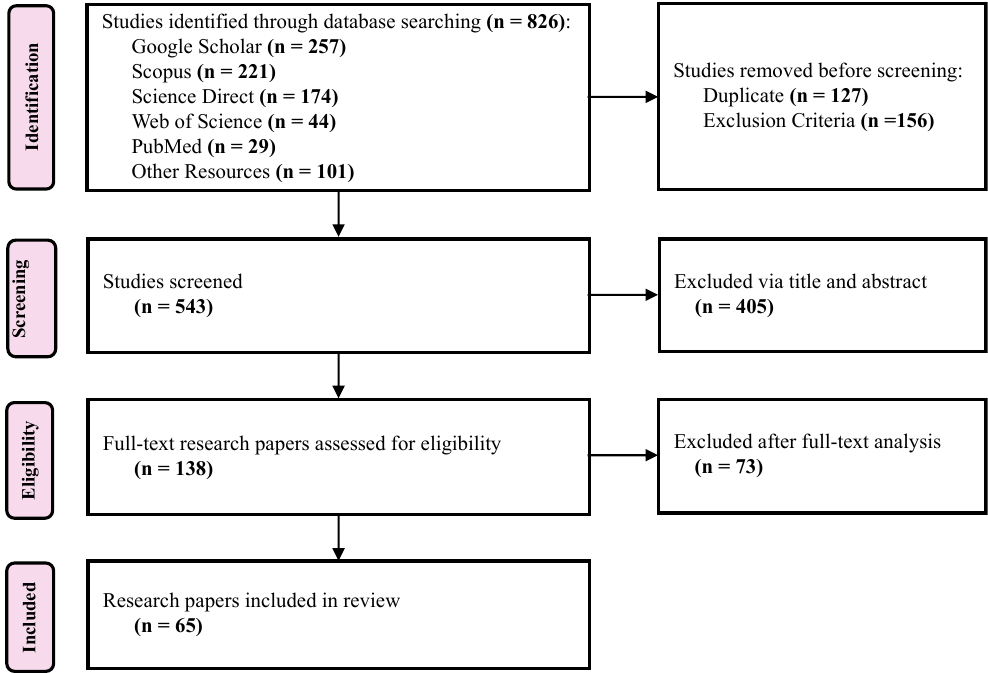}
    \caption*{\textbf{Figure 3.} PRISMA-based Search Strategy}
    \label{fig:mltc-approaches}
\end{figure}

\subsection{Overview of Selected Research Papers}
This systematic review includes 65 research papers published between 2018 to the third quarter of 2024. This timeframe was selected to capture the evolution and application of both PLMs and LLMs. For consistency with the literature, both models are referred to as LLMs in this paper. Following the exclusion criteria detailed before, the publications years ended up covering the period from 2020 to Q3 2024. Figure 4 shows the annual publication count during the last five years. A major shift in language modeling methods and applications marks this period. A significant increase is remarked in the first three quarters of 2024, which demonstrates a new trend in healthcare text classification where LLMs are leveraged. 

\begin{figure}[H]
    \centering
    \includegraphics[scale=1.25]{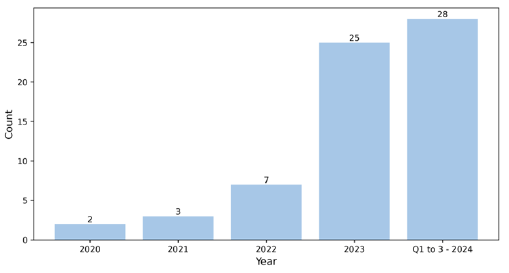}
    \caption*{\textbf{Figure 4.} Eligible Publications Per Year}
    \label{fig:mltc-approaches}
\end{figure}

Since the majority of available healthcare text data, regardless of its type, is English, more than 80\% of the research studies deemed eligible covered entirely or partially English datasets, as shown in Figure 5. However, this is not the only reason behind the illustrated text language distribution. Given the abundance of English text data, many popular LLMs (e.g., BERT, GPT) were initially developed and trained in English regardless of the application domain. Additionally, some researchers would choose to translate the healthcare dataset text into English, especially in the case of low-resource languages.
\begin{figure}[H]
    \centering
    \includegraphics[scale=1]{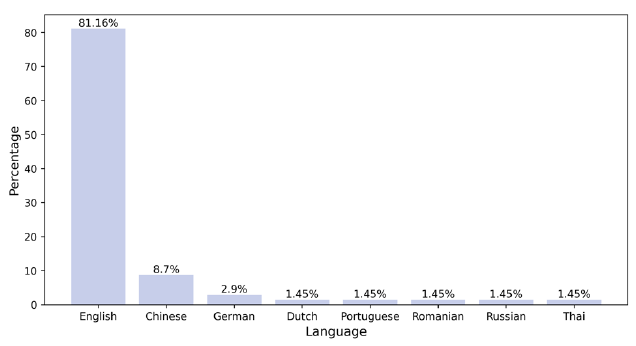}
    \caption*{\textbf{Figure 5.} Eligible Publications Text Data Languages}
    \label{fig:mltc-approaches}
\end{figure}

Healthcare text data used in classification studies can be categorized into three main types: Clinical Notes, Healthcare Communications, and Literature/Research. Clinical Notes comprise the largest volume of data, primarily due to their continuous generation through daily medical procedures and healthcare operations. Healthcare Communications data represents the second largest category, though significantly smaller in volume than Clinical Notes. This category has grown with the expansion of social media usage and healthcare facilities' increased use of patient satisfaction surveys. Research/Literature documents make up the smallest category, with a volume comparable to the Healthcare Communications category. 

\begin{figure}[H]
    \centering
    \includegraphics[scale=1]{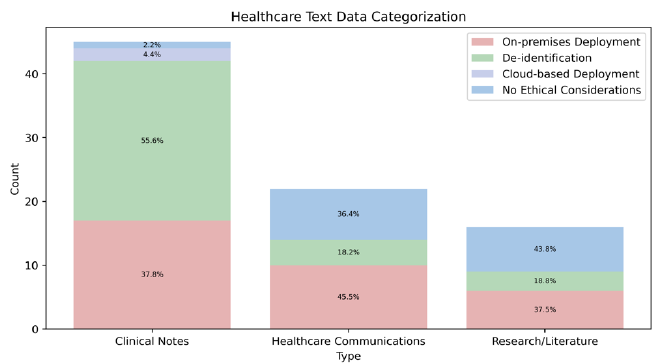}
    \caption*{\textbf{Figure 6.} Healthcare Text Data Categorization}
    \label{fig:mltc-approaches}
\end{figure}

Figure 6 illustrates both the distribution of healthcare text data types and the ethical strategies employed for each category. The data reveals three main protective approaches: on-premises deployment, patient de-identification, and cloud-based deployment. The Clinical Notes category, containing the most sensitive patient information, demonstrates the strongest ethical protocols. This category shows the highest rate of de-identification, closely followed by on-premises deployment, and is the only category utilizing cloud-based deployment (via Azure Services). Notably, the Clinical Notes category has the lowest number of papers lacking explicit ethical considerations. In contrast, both the Healthcare Communications and the Research/Literature categories frequently lack explicitly mentioned or deduced ethical considerations, with Research/Literature showing the highest occurrence, followed by Healthcare Communications. Both categories show similar de-identification rates of approximately 18\%, with the remaining papers implementing LLM-based on-premises deployment strategies.

\begin{figure}[H]
    \centering
    \begin{subfigure}[b]{0.48\textwidth}
        \centering
        \includegraphics[width=\textwidth]{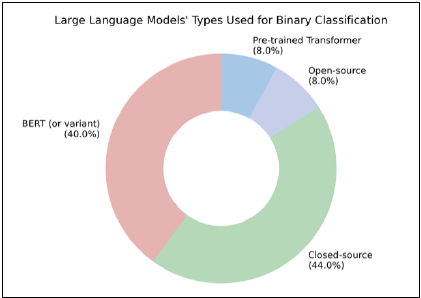}
        \caption*{\textbf{Figure 7.} LLMs Types Used for Binary Classification}
        \label{fig:mltc-approaches-a}
    \end{subfigure}
    \hfill
    \begin{subfigure}[b]{0.483\textwidth}
        \centering
        \includegraphics[width=\textwidth]{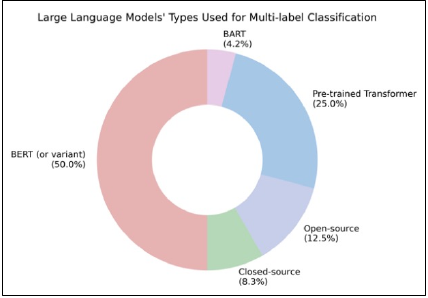} 
        \caption*{\textbf{Figure 8.} LLMs Types Used for Multi-label Classification}
        \label{fig:mltc-approaches-b}
    \end{subfigure}
    \caption*{} 
    \label{fig:mltc-approaches}
\end{figure}

\begin{figure}[H]
    \centering
    \begin{subfigure}[b]{0.48\textwidth}
        \centering
        \includegraphics[width=\textwidth]{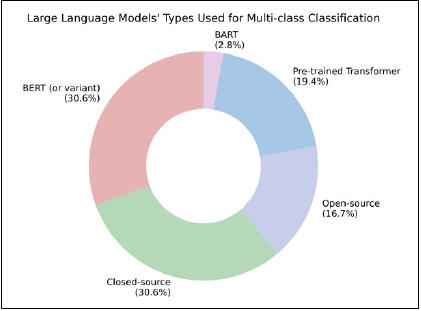}
        \caption*{\textbf{Figure 9.} LLMs Types Used for Multi-class Classification}
        \label{fig:mltc-approaches-a}
    \end{subfigure}
    \hfill
    \begin{subfigure}[b]{0.483\textwidth}
        \centering
        \includegraphics[width=\textwidth]{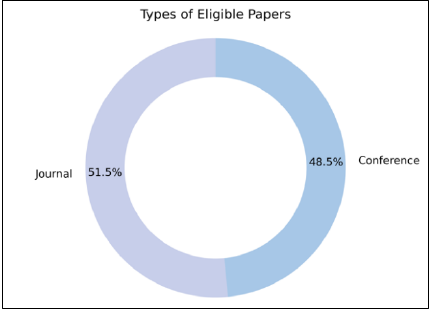} 
        \caption*{\textbf{Figure 10.} Types of Eligible Papers}
        \label{fig:mltc-approaches-b}
    \end{subfigure}
    \caption*{} 
    \label{fig:mltc-approaches}
\end{figure}

On one hand, Figures 7, 8, and 9 illustrate the distribution of LLM types used for different text classification tasks in the eligible papers of this systematic review. For Binary Classification, Closed-source LLMs were most frequently used, followed closely by BERT (or variant) approaches, while Open-source LLMs and Pre-trained Transformers were used least frequently. In Multi-label Classification, BERT (or variant) was predominant, followed by Pre-trained Transformers, then Open-source LLMs, and finally Closed-source LLMs. BART made its first appearance in this category and was also utilized in Multi-class Classification. For Multi-class Classification, Closed-source LLMs were most common, followed by BERT (or variant), Pre-trained Transformers, and Open-source LLMs. Across all three classification types, BERT (or variant) models maintained a significant presence, likely due to their feasibility for local implementation, particularly through fine-tuning approaches. The analysis also revealed the popularity of Closed-source LLMs, especially the GPT family, which were frequently employed in both multi-class and binary classifications. On the other hand, Figure 10 categorizes the reviewed eligible papers by publication type, distinguishing between journal articles and conference proceedings. The analysis reveals that these two categories have comparable representation in the literature.

\newpage

\subsection{Taxonomy of Selected Research Papers}
The research papers resulting from PRISMA’s eligibility phase are categorized according to seven dimensions: Healthcare Text Data Type, Ethical Considerations, Text Classification Type, Text Classification Application, Methodology Approach Type, Large Language Model Type, and Performance Evaluation. Figure 11 summarizes the different categories of each dimension considered in this systematic review. It is worth mentioning that the LLM Type only accounts for the best-performing model(s) in the research paper reviewed. Also, the papers’ count would usually not add up to the total of papers considered since one research study can include multiple classification tasks. 

\begin{figure}[H]
    \centering
    \includegraphics[scale=0.75]{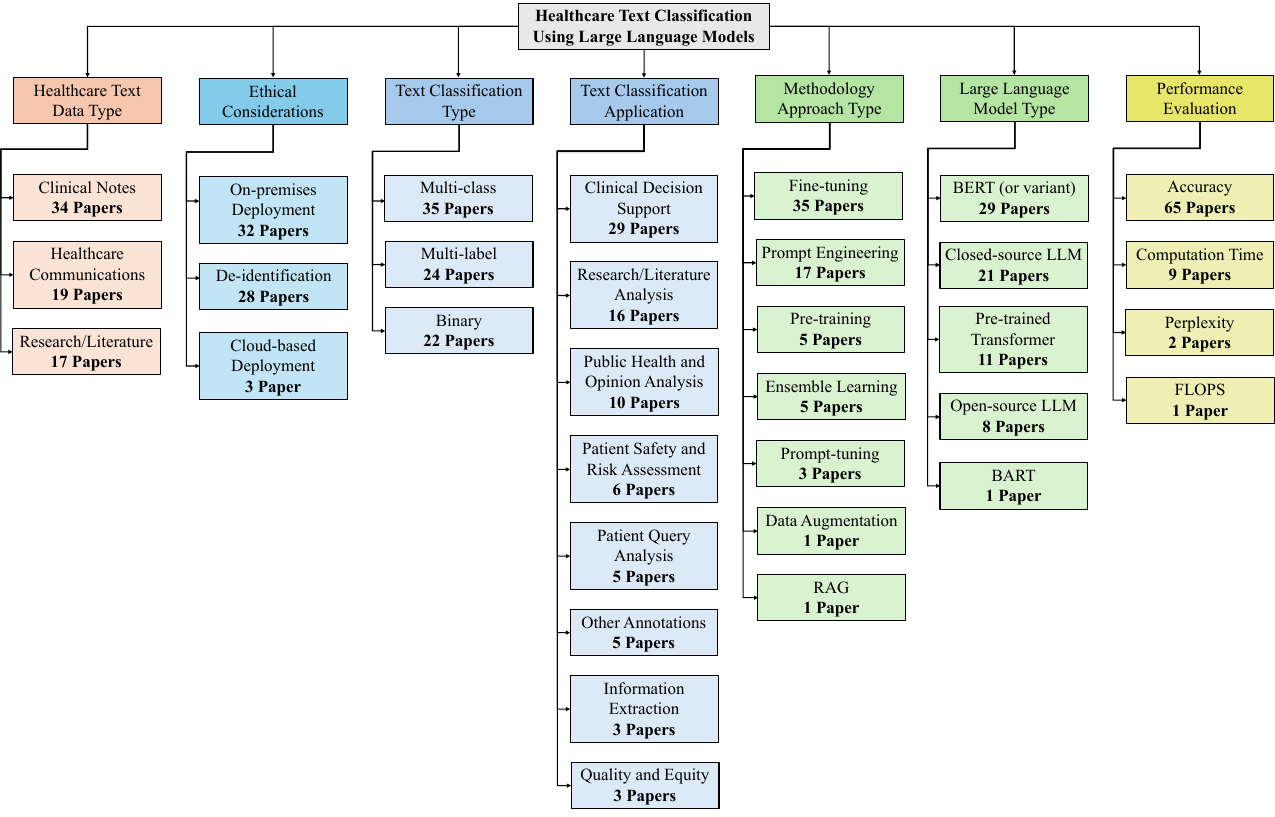}
    \caption*{\textbf{Figure 11.} Eligible Publications Categorization}
    \label{fig:mltc-approaches}
\end{figure}

\section{Discussion}
This section discusses and analyzes eligible research papers based on the seven previously introduced dimensions. Each dimension is detailed and further investigated and explored based on the different categories it encompasses. As a result, research gaps and limitations are deduced and highlighted.

\subsection{Healthcare Text Data Type and Ethical Considerations}
The type of healthcare-related text data varies depending on the context in which it was collected. It encompasses a wide range of information sources, ranging from clinical notes to healthcare communications, literature, and research. Depending on the type of text used for classification, advances in medical research, patient care improvement, and clinical decision-making support may vary fundamentally. Ethical considerations also vary from one research study to another and can be categorized into on-premises and cloud-based deployments as well as patients’ de-identification.
\\[0.25cm]
Clinical Notes represent one of the healthcare-related text data that is witnessing extremely rapid growth. This type of data covers, for instance, progress notes, admissions notes, discharge summaries, and treatment plans and is generated by healthcare professionals during a patient encounter or care. Clinical notes are characterized by including medical jargon and abbreviations and likely being inconsistent in formatting. Once digitally gathered and saved, they can be classified for multiple purposes, such as being the basis of a tool dedicated to diagnosis decision-support. Depending on availability and access, some researchers classified healthcare text using LLMs and a specific hospital’s clinical notes. However, other studies resorted to publicly available clinical notes. Another type of healthcare text data often categorized using LLMs is Healthcare Communications. These communications encompass patient-generated text data, such as feedback shared in hospitals’ surveys, comments collected from social media, or public opinion posted on different online healthcare platforms. Additionally, healthcare communications can include encounter messages shared between hospital professionals, feedback that medical students can receive from clinical mentors, and patient inquiries left in the hospital’s portal. A third type of healthcare text data found in the literature is Research/Literature, where healthcare research papers or medical blog articles are classified. Figure 11 categorizes the healthcare text classification types identified above in the eligible studies. Table 1 summarizes the research studies, in which each healthcare text data type was experimented with for LLM-based text classification, in addition to the ethical considerations resorted to, if any. 

\begin{landscape}
\begin{table}[h]
\captionsetup{labelformat=empty}
\centering
\caption*{\textbf{Table 1.} Healthcare Text Data-based Categorization of Reviewed Literature}
\label{tab:classification_models}
\begin{tabular}{c c c c c c c}
 \toprule
\multirowcell{2}[0pt][c]{\centering\makecell{\textbf{Reference}}} & 
\multicolumn{3}{c}{\makecell{\textbf{Healthcare Text Data Type}}} & 
\multicolumn{3}{c}{\makecell{\textbf{Ethical Considerations}}}
\\
\cmidrule(lr){2-7}
& \textbf{Clinical Notes} &  \makecell{\textbf{Healthcare} \\ \textbf{Communications}} & \textbf{Research/Literature} & \makecell{\textbf{On-premises} \\ \textbf{Deployment}} & \textbf{De-identification} & \makecell{\textbf{Cloud-based} \\ \textbf{Deployment}} \\
\hline
\textbf{Sushil et al.} (2024) & \checkmark & & & &  \checkmark &  \checkmark \\
\hline
\textbf{Lossio-Ventura et al.} (2024) & & \checkmark   &  & & \checkmark  & \\
\hline
\textbf{Chen et al.} (2024a) & &  &  \checkmark   & & & \\
\hline
\textbf{Ohse et al.} (2024) & \checkmark  &  &  & & \checkmark  & \\
\hline
\textbf{BT and Chen} (2024) & \checkmark  &  &  & & & \\
\hline
\textbf{Kim et al.} (2024) & & \checkmark   &  & & & \\
\hline
\textbf{Gu et al.} (2024) & & \checkmark   &  & \checkmark & & \\
\hline
\textbf{Raja et al.} (2024) & &  &  \checkmark   & & & \\
\hline
\textbf{Guo et al.} (2024) & &  &  \checkmark   & & & \\
\hline
\textbf{Yang et al.} (2024) & &  &  \checkmark   & & & \\
\hline
\textbf{Chang et al.} (2024) & \checkmark  &  &  & \checkmark & & \\
\hline
\textbf{Peng et al.} (2024) & \checkmark  &  &  & & & \\
\hline
\textbf{Williams et al.} (2024) & \checkmark  &  &  & & \checkmark  & \\
\hline
\textbf{Bețianu et al.} (2024) & &  &  \checkmark   & \checkmark & \checkmark  & \\
\hline
\textbf{Farruque et al.} (2024) & & \checkmark   &  & \checkmark & & \\
\hline
\textbf{Xu et al.} (2024) & & \checkmark   &  & & & \\
\hline
\textbf{Li et al.} (2024b) & \checkmark  &  &  & & \checkmark  & \\
\hline
\textbf{Guevara et al.} (2024) & \checkmark  &  &  & \checkmark & \checkmark  & \\
\hline
\textbf{Luo et al.} (2024) & & \checkmark   &  \checkmark   & & & \\
\hline
\textbf{Xie et al.} (2024) & \checkmark  &  &  & \checkmark & & \\
\hline
\textbf{Bumgardner et al.} (2024) & \checkmark  &  &  & \checkmark & & \\
\hline
\textbf{Wang et al.} (2024) & \checkmark  &  &  & & \checkmark  & \\
\hline
\textbf{Chen et al.} (2024b) & &  &  \checkmark   & \checkmark & & \\
\hline
\textbf{Silverman et al.} (2024) & \checkmark  &  &  & & \checkmark  & \\
\hline
\textbf{Uslu et al.} (2024) & \checkmark  &  &  & \checkmark & \checkmark  & \\
\hline
\textbf{Shi et al.} (2023) & & & \checkmark  & & &  \\
\hline
\textbf{Li et al.} (2023a) &\checkmark  & & & \checkmark  &  \checkmark & \checkmark   \\
\hline
\textbf{Aldeen et al.} (2023) & & \checkmark & & & &  \\
\hline
\textbf{Liu et al.} (2023) & \checkmark & & & & & \checkmark   \\
\hline
\textbf{Ramteke and Khandelwal} (2023) & & \checkmark & & & &  \\
\hline
\textbf{Carneros-Prado et al.} (2023) & & \checkmark & & & &  \\
\hline
\textbf{Wang et al.} (2023) & & \checkmark & \checkmark  & & \checkmark &  \\
\hline
\textbf{Alsentzer et al.} (2023) & \checkmark & & & \checkmark   & &  \\
\hline
\textbf{Sarkar et al.} (2023) & & & \checkmark  & & &  \\
\hline
\textbf{Yuan et al.} (2023) & \checkmark & & & & \checkmark &  \\
\bottomrule
\end{tabular}
\end{table}
\end{landscape}

\begin{landscape}
\begin{table}[h]
\captionsetup{labelformat=empty}
\centering
\caption*{\textbf{Table 1.} Cont.}
\label{tab:classification_models}
\begin{tabular}{c c c c c c c}
 \toprule
\multirowcell{2}[0pt][c]{\centering\makecell{\textbf{Reference}}} & 
\multicolumn{3}{c}{\makecell{\textbf{Healthcare Text Data Type}}} & 
\multicolumn{3}{c}{\makecell{\textbf{Ethical Considerations}}}
\\
\cmidrule(lr){2-7}
& \textbf{Clinical Notes} &  \makecell{\textbf{Healthcare} \\ \textbf{Communications}} & \textbf{Research/Literature} & \makecell{\textbf{On-premises} \\ \textbf{Deployment}} & \textbf{De-identification} & \makecell{\textbf{Cloud-based} \\ \textbf{Deployment}} \\
\hline
\textbf{Kementchedjhieva and Chalkidis} (2023) & \checkmark & & \checkmark  & & \checkmark &  \\
\hline
\textbf{Chen et al.} (2023) & & & \checkmark  & \checkmark & &  \\
\hline
\textbf{Wu et al. }(2023) & & \checkmark & & \checkmark & &  \\
\hline
\textbf{Li et al.} (2023b) & \checkmark & & & & \checkmark &  \\
\hline
\textbf{McMaster et al.} (2023) & \checkmark & & & \checkmark & &  \\
\hline
\textbf{Lehman et al. }(2023) & \checkmark & & & \checkmark & \checkmark &  \\
\hline
\textbf{Gretz et al.} (2023) & & & \checkmark  & \checkmark & & \\
\hline
\textbf{Savage et al.} (2023) & \checkmark & & & \checkmark & \checkmark &  \\
\hline
\textbf{Cui et al.} (2023) & \checkmark & & & \checkmark & \checkmark &  \\
\hline
\textbf{Van Ostaeyen et al.} (2023) & & \checkmark & & \checkmark & &  \\
\hline
\textbf{Jiang et al.} (2023) & & \checkmark & & & &  \\
\hline
\textbf{Ge et al.} (2023) & & \checkmark & & & &  \\
\hline
\textbf{Ren et al.} (2023) & & \checkmark & & \checkmark & & \\
\hline
\textbf{Qi et al.} (2023) & &  & \checkmark  & \checkmark & &  \\
\hline
\textbf{Ciobotaru and Dinu }(2023) & & \checkmark & & \checkmark & &  \\
\hline
\textbf{Tan et al.} (2023) & \checkmark & & & & \checkmark \\
\hline
\textbf{Bansal et al.} (2023) & & \checkmark & & \checkmark & \checkmark \\
\hline
\textbf{Kersting et al.} (2023) & & \checkmark & & \checkmark & \\
\hline
\textbf{Sivarajkumar and Wang} (2022) & \checkmark & & & \checkmark & \checkmark \\
\hline
\textbf{Yang et al.} (2022) & \checkmark & & \checkmark & \checkmark & \checkmark \\
\hline
\textbf{Yogarajan et al.} (2022a) & \checkmark & & & & \checkmark \\
\hline
\textbf{Shiju et al.} (2022) & & \checkmark & & \checkmark & \\
\hline
\textbf{Chen et al.} (2022) & & & \checkmark & \checkmark & \\
\hline
\textbf{Chaichulee et al.} (2022) & \checkmark & & & \checkmark & \checkmark \\
\hline
\textbf{Yogarajan et al.} (2022b) & \checkmark & & & & \checkmark \\
\hline
\textbf{Schneider et al.} (2021) & \checkmark & & \checkmark & & \checkmark \\
\hline
\textbf{Yogarajan et al.} (2021) & \checkmark & & & \checkmark & \checkmark \\
\hline
\textbf{Bressem et al.} (2020) & \checkmark & & & \checkmark & \\
\hline
\textbf{Pan et al.} (2020) & \checkmark & & & \checkmark & \\
\hline
\textbf{Blinov et al.} (2020) & \checkmark & & & & \checkmark \\
\bottomrule
\end{tabular}
\end{table}
\end{landscape}

\subsubsection{Clinical Notes}
Pathology reports have served as crucial data sources in clinical research. Sushil et al. (2024) analyzed breast cancer pathology reports from the University of California, San Francisco’s (UCSF) clinical data warehouse, manually labeled for 12 treatment-relevant categories. Bumgardner et al. (2024) investigated a large collection of surgical pathology reports from the University of Kentucky, focusing on cancer-related cases with International Classiﬁcation of Diseases (ICD) condition codes. Chang et al. (2024) used pathology reports from The Cancer Genome Atlas (TCGA) project of the National Cancer Institute (NCI) to extract pathologic tumor-node-metastasis (pTNM) staging information. This was translated into three multi-class text classifications, and their evaluation relied on existing annotations. Additionally, the prompt-based text classification was also validated using cancer-specific clinical reports (i.e., Breast Invasive Carcinoma (BRCA) and Lung Adenocarcinoma (LUAD)).
\\[0.25cm]
Radiology documentation has been extensively studied across multiple research papers as well. Bressem et al. (2020) leveraged 3.8 million radiology reports, including chest radiographs and CT scans, for pre-training and fine-tuning BERT models. A large size of reports was manually annotated for findings such as congestion, opacity, effusion, pneumothorax, and the presence of medical devices. This annotated set was split into a set of reports for fine-tuning and a smaller set for testing. Additionally, a small number of CT reports were used to evaluate the models' performance on longer texts. Tan et al. (2023) examined CT reports from the National Cancer Centre Singapore across four cancer types, employing an 80-10-10 split for training, development, and testing. Uslu et al. (2024) utilized the Medical Information Mart for Intensive Care Chest X-ray (MIMIC-CXR) dataset, focusing specifically on the 'FINDINGS' section from radiological reports of chest X-rays. These reports were generated by radiologists interpreting chest radiographs from patients admitted to the emergency department. The dataset contains detailed descriptions of radiological findings, which were used to be classify into 14 distinct impressions, including 13 specific abnormalities (e.g., atelectasis, cardiomegaly, consolidation, etc.) and a “no finding” category. Liu et al. (2023) evaluated various radiology-related text classification tasks (e.g., sentence similarity, disease classification) using radiology reports from MS-CXR-T, RadNLI, and Chest ImaGenome datasets.
\\[0.25cm]
Discharge summaries have provided rich data for various analyses. Li et al. (2023a) utilized MIMIC-III discharge summaries as part of their Silver dataset, annotated using Llama 65B. Alsentzer et al. (2023) examined obstetric-related discharge summaries from Mass General Brigham hospitals for Postpartum Hemorrhage classification. Wang et al. (2024) analyzed the “brief hospital course” sections from MIMIC-IV discharge summaries. These summaries include key events, diagnostics, and treatments during hospitalization. Cui et al. (2023) studied temporal relationships in n2c2 2012 challenge’s discharge summaries. These summaries compile patient hospital stay information, including treatments and their temporal relationships. The task consisted of classifying whether treatments occurred during hospitalization or not.
\\[0.25cm]
Progress and visit notes have also offered insights into patient care patterns. Williams et al. (2024) analyzed Emergency Department physician notes, focusing on chief concerns and illness histories. Schneider et al. (2021) examined progress notes from a Brazilian tertiary hospital. Savage et al. (2023) studied MIMIC-III history and physical notes for anticoagulant usage. Silverman et al. (2024) investigated outpatient clinical notes for inflammatory bowel disease from UCSF.
\\[0.25cm]
Specialty-specific documentation has provided targeted clinical insights as well. Xie et al. (2024) classified seizure status in University of Pennsylvania Health System epilepsy notes as either ‘seizure-free’ or ‘having recent seizures’. Guevara et al. (2024) analyzed both radiotherapy and immunotherapy treatment notes from multiple institutions. The first dataset compiles clinical notes of cancer patients receiving radiotherapy (RT)  at the Department of Radiation Oncology at Brigham and Women’s Hospital/Dana-Farber Cancer Institute in Boston, Massachusetts. At the same time, the second dataset comprises clinical notes of patients receiving immunotherapy treatment and not included in the first dataset. Chaichulee et al. (2022) examined Thai-English drug allergy records from Songklanagarind Hospital, covering 36 predefined symptom terms. Each record contained a free-text description of adverse drug reactions documented by healthcare professionals in a mixture of Thai and English languages.
\\[0.25cm]
Transcribed clinical interactions have offered unique perspectives on patient assessment. Ohse et al. (2024) analyzed GRID-HAMD-17 protocol interviews for depression classification, creating five distinct clusters. The interviews were conducted in German, but transcriptions were translated to English for depression assessment. These transcriptions constituted the original dataset. Additionally, a clustered dataset was created to provide more context to the text classification LLM. Each interview transcription was grouped into five clusters (i.e., depression, anxiety, somatic, insomnia, and unimportant) depending on the question. B.T and Chen (2024) examined transcribed “Cookie Theft Picture” descriptions for Alzheimer's Disease assessment. These transcriptions resulted from speech from both Alzheimer's Disease (AD) and Cognitively Normal (CN) subjects.
\\[0.25cm]
Administrative and coding documentation has supported various classification tasks. Yogarajan et al. (2021, 2022a) utilized clinical notes for ICD-9 code prediction while Kementchedjhieva and Chalkidis (2023) focused on MIMIC-III discharge summaries for ICD-9 coding.
\\[0.25cm]
Specialized clinical datasets have enabled focused research objectives. Lehman et al. (2023) used the Clinical Language Inference for Patient Monitoring (CLIP) dataset destined for key follow-up information in discharge summaries, besides MEDNLI. Peng et al. (2024) leveraged multiple clinical note datasets for text classification. The clinical abbreviation disambiguation task was evaluated using an abbreviation dataset elaborated by the University of Minnesota (UMN), while the Natural Language Inference (NLI) resorted to the MedNLI benchmark dataset, the medication attribute filling was validated using the Contextualized Medication Event Dataset (CMED), and progress note understanding was validated using the benchmark dataset developed by the 2022 n2c2 challenge (Track 3), derived from the MIMIC-III.
\\[0.25cm]
Critical care documentation has provided insights into acute care settings. Li et al. (2023b) analyzed ICU admission notes from MIMIC-AKI (Acute Kidney Injury). Guevara et al. (2024) included critical care unit inpatient notes in their multi-dataset study. Yogarajan et al. (2022b) examined eICU program records alongside MIMIC-III data.
\\[0.25cm]
Longitudinal patient records have offered comprehensive views of patient care. Li et al. (2023a) developed a Gold dataset from Longitudinal Electronic Health Records (EHRs) Notes of Alzheimer's Disease patients. Pan et al. (2020) analyzed EHRs containing eleven implicit symptoms or diseases. Sivarajkumar and Wang (2022) studied MIMIC-III patient notes for high-context phenotypes related to treatment and readmission risk. Yuan et al. (2023) resorted to ClinicalTrials.gov to collect six stroke clinical trials. Their content consists of inclusion and exclusion criteria. Additionally, UTHealth stroke patient database was used to retrieve patients’ EHRs containing diagnoses, procedures, and medications. Additional comprehensive studies include those by Yang et al. (2022), McMaster et al. (2023), and Li et al. (2024b), who utilized various MIMIC-III note types for model development and validation.

\subsubsection{Healthcare Communications}
Healthcare communications research demonstrates diverse focus areas across several key healthcare domains. In mental health and psychological communications, multiple studies have examined different aspects: Lossio-Ventura et al. (2024) analyzed mental health during COVID-19 through web-based surveys, while Aldeen et al. (2023) and Ramteke and Khandelwal (2023) focused on mental health manifestations in social media posts, particularly regarding anxiety, depression, and stress. Xu et al. (2024) conducted a comprehensive analysis of mental health datasets from social media platforms, examining stress, depression, and suicide ideation, complemented by Jiang et al.'s (2023) investigation of social anxiety and Farruque et al.'s (2024) analysis of depression-related tweets.
\\[0.25cm]
Within the framework of public health and vaccination communications, several researchers explored vaccination-related discourse: Kim et al. (2024) examined Human Papillomavirus (HPV) vaccination messages across social media platforms, while Carneros-Prado et al. (2023), Bansal et al. (2023), and Ciobotaru and Dinu (2023) focused on COVID-19 vaccination discussions, particularly analyzing public sentiment and concerns expressed on social media.
\\[0.25cm]
Clinical and medical services communications were examined through various lenses: Wang et al. (2023) and Luo et al. (2024) utilized the KUAKE-QIC dataset, where short texts representing patient inquiries are compiled and labeled into 11 intention classes, for patient inquiry analysis, Ren et al. (2023) examined Patient Portal Messages from clinical departments at Mayo Clinic, Shiju et al. (2022) analyzed drug reviews and medical conditions, and Kersting et al. (2023) investigated physician reviews and ratings. In medical education, Van Ostaeyen et al. (2023) uniquely focused on healthcare students' ePortfolio feedback across different healthcare programs.
\\[0.25cm]
Disease-specific communications were represented by studies focusing on diabetes-related communications, with both Ge et al. (2023) and Wu et al. (2023) analyzing diabetes-related questions and categorizing them into various classes. Finally, emotional and sentiment analysis in healthcare communications was explored by Gu et al. (2024), who analyzed six-class sentiment expressions on Weibo and was also incorporated into Jiang et al.'s (2023) analysis of therapy-related sentiments.

\subsubsection{Research/Literature}
Research in disease-specific medical literature classification has covered various medical conditions. Shi et al. (2023) focused on cardiovascular diseases using the Ohsumed dataset, implementing single-label classification. Chen et al. (2024a) developed a three-class categorization system for diabetes mellitus papers, while Raja et al. (2024) categorized ocular disease literature into 19 categories. COVID-19 research has also been present, with Guo et al. (2024) analyzing treatment-related papers and Yang et al. (2024) developing a binary classification for SARS-CoV-2 and Nipah virus literature for drug discovery purposes.
\\[0.25cm]
In the scope of clinical and medical topics, Wang et al. (2023) tackled the classification of clinical trial screening criteria, developing 44 semantic categories through the CHIP-CTC dataset, encompassing descriptive sentences. Sarkar et al. (2023) addressed the categorization of medical blog articles across 18 predefined topics, including headache, mental health, and heart health. Cancer research classification has been significant, with multiple studies, including Chen et al. (2022) and Chen et al. (2023), utilizing the Hallmarks of Cancer (HoC) dataset to classify cancer biology characteristics.
\\[0.25cm]
Other research studies focused on different research classifications. Chen et al. (2024b) developed a three-tier advice classification system (no advice, weak advice, strong advice) for medical research abstracts. Kementchedjhieva and Chalkidis (2023) worked with the BIOASQ dataset (consisting of biomedical articles from PubMed), implementing classification based on the Medical Subject Headings (MeSH) taxonomy. Qi et al. (2023) addressed industrial biomedical literature mining tasks, focusing on recognizing special biomedical phrases. The data was acquired with inherent label noise due to crowd-sourcing and labeling preferences. For the testing part, relabeling was conducted on a subset of data and assumed being clean.
\\[0.25cm]
General medical literature classification has been explored through various datasets. Yang et al. (2022) leveraged PubMed abstracts and Wikipedia articles for pre-training, while Schneider et al. (2021) utilized both PubMed and SciELO databases for fine-tuning. Gretz et al. (2023) contributed to this field by working with the Medical Abstracts dataset. Bețianu et al. (2024) and Luo et al. (2024) further expanded the research using PubMed datasets, with Luo et al. specifically incorporating multiple datasets, including BC7LitCovid, for comprehensive biomedical literature classification.

\subsubsection{Ethical Considerations}
These research studies employed various data sources, from social media platforms to EHR and literature databases, and used different categorization approaches ranging from binary to multi-class and multi-label classifications. The research spans multiple languages and formats, reflecting healthcare text classification research's global and diverse nature. However, leveraging LLMs to carry out this type of task in healthcare presents significant ethical concerns around patient privacy and data security without forgetting about the algorithmic bias that can emerge. The protection of sensitive health information is crucial under regulations that vary based on the country, such as the Health Insurance Portability and Accountability Act (HIPAA) in the U.S. This is particularly important when dealing with an LLM requiring API requests (e.g., GPT-4o) where the classification cannot be run locally, and when the text data consists of clinical notes where patients’ Protected Health Information (PHI) should be detected and de-identified in advance. To overcome the challenge resulting from the sensitive nature of healthcare textual data, researchers in the reviewed literature resorted to different ethical considerations. Some research papers conducted all their experiments locally; this is especially convenient when fine-tuning BERT (or variant). Others opted for anonymizing the textual data before providing it as input to the model or using already de-identified data (e.g., MIMIC datasets). In contrast, very few research studies used secure cloud-based deployments through services like Microsoft Azure OpenAI. However, in many cases, ethical considerations were minimal. This is the case for publicly available data such as medical literature and some cases of social media posts-based datasets. Table 1 details the specific approaches adopted by each study, reflecting varying ways for maintaining ethical awareness in the application of LLMs to healthcare text classification (Vayena et al., 2018).

\subsection{LLM-based Methodology Text Classification Approach Type and Evaluation}
This section categorizes the eligible research papers based on the methodology type, detailed in Figure 11, adopted by the best-performing approach. Additionally, the text classification type and application are provided for each study. The aim is to discuss and analyze these research studies. It is worth noting that some research papers evaluated their methodology using various datasets; only the healthcare-related ones are considered in this systematic review.
\\[0.25cm]
In each methodology’s category, two types of tables are included. These tables summarize each reviewed paper’s details and provide detailed performance evaluation accuracy metrics. The focus is on the accuracy (e.g., ACC@1), the F1 score (e.g., micro), the Precision (e.g., mean Average Precision (mAP)), the Recall (e.g., macro), and the AUC score (e.g., Precision-Recall (PR), Receiver Operating Characteristic (ROC)). The reported metrics can refer to the overall performance of each dataset, each class, or each classification task.

\subsubsection{Prompt Engineering}
Prompt engineering is a practice that consists of crafting input prompts that provide enough context to an LLM to maximize its performance on various NLP tasks. These tasks can range from text generation and translation to classification and summarization. This technique became particularly popular with the launch of products like ChatGPT and Claude. It involves designing the prompt’s format, wording, and structure to help the model better understand the task and improve the accuracy and efficiency in obtaining the desired output. Brown et al. (2020) demonstrated that carefully constructed prompts can significantly influence how an LLM interprets a task and produces results. They deduced that zero-shot (i.e., where no input-output example pairs are provided in the prompt) and few-shot learning (i.e., where typically one to five input-output example pairs are provided in the prompt) benefit greatly from prompt engineering. By providing contextual clues or examples in the prompt, LLMs can generalize and perform tasks they were not explicitly trained on; this is also known as an emergent behavior (Raschka, 2024). To help the LLMs better align with human expectations, multiple prompt engineering techniques were introduced in the literature. As discussed in the systematic review conducted by Liu et al., 2023, in the particular task of text classification, these prompt engineering techniques can include, for instance, cloze-prompts (e.g., “The topic is [Z]”) or prefix-prompts (e.g., “What is the sentiment?”). These techniques can always be enhanced through in-context learning (i.e., including examples in the prompt) and prompt ensembling (i.e., combining multiple prompts). The main goal is formulating the classification task to leverage the LLMs’ pre-trained knowledge best. As a result, the LLM’s capabilities are harnessed to perform text classification tasks without extensive pre-training (of further pre-training) or fine-tuning. Additionally, this technique efficiently eliminates the need for multiple specialized models, as a single LLM can adapt to various classification tasks via prompt design. For example, Schick and Schütze (2021) introduced the Pattern-Exploiting Training (PET) method, where task-specific patterns or templates are used to rephrase the input text so that the LLM can better understand the classification task. In healthcare particularly, the text to be classified can be complex, with nuanced medical terminology and varying contexts. In this case, prompt engineering can help guide the LLM and, therefore, contribute to providing a scalable way to implement advanced text classification systems for healthcare applications, facilitating more accurate and efficient healthcare decision-making.
\\[0.25cm]
The prompt engineering approach is the second most used approach in the covered literature in this systematic review, thanks to its accessibility and efficiency. As summarized in Table 3 and further detailed in Table 2, the best-performing LLM often belongs to the GPT family. These general-purpose LLMs trained on vast corpora continuously exhibit impressive capabilities in handling many NLP tasks, including text classification. In parallel, other research studies leveraged BARD, a variation of T5, or Clinical BERT.
\\[0.25cm]
Sushil et al. (2024) used GPT-4 through Azure OpenAI Studio, ensuring HIPAA compliance, for zero-shot classification of breast cancer pathology reports. The methodology involved zero-shot classification using single prompts that requested all classification tasks simultaneously, with outputs structured in JSON format for automated evaluation. This paper focused on extracting 12 key pathology features, including tumor characteristics, biomarkers, margin status, and lymph node involvement. This information can be useful for breast cancer diagnosis and treatment planning. Lossio-Ventura et al. (2024) used ChatGPT (based on GPT-3.5) for zero-shot sentiment analysis of COVID-19 survey responses. The method involved feeding individual documents to ChatGPT with the simple prompt “What is the sentiment of the following sentence 'x'?” where x was the text to be analyzed. Since no examples were provided in the prompt, a zero-shot setting can be deduced. This study focused on analyzing sentiment in free-text responses from two COVID-19 survey datasets, one from the National Institutes of Health (NIH) and one from Stanford, which captured people's experiences and attitudes during the pandemic lockdown. Shi et al. (2023) leveraged ChatGPT while combining it with graphs, resulting in proposing ChatGraph. ChatGPT was first used to refine input text (i.e., correcting grammar and improving readability) and then to extract knowledge graphs as triplets (head entity, relation, tail entity) using carefully designed prompts. These knowledge graphs were converted into text graphs where words became nodes and relationships became edges, which were then used to train an interpretable linear classifier, Graph Convolution Network (GCN). Always leveraging graphs, Chen et al. (2024a) explored using LLMs through two pipelines: LLMs-as-Enhancers and LLMs-as-Predictors. In the first pipeline, LLMs enhance node text attributes either at the feature level (by encoding text into embeddings) or at the text level (by augmenting text attributes), which are then used by Graph Neural Networks (GNNs) for predictions. The second pipeline directly uses LLMs to make predictions by converting graph structural information into natural language prompts. The LLM used in this approach is ChatGPT (GPT-3.5-turbo-0613). Ohse et al. (2024) investigated the potential of four NLP models (BERT, Llama2-13B, GPT-3.5, and GPT-4) for detecting depression through clinical interviews. They tested two main strategies: zero-shot learning with the LLMs and a clustering approach where interview data was segmented into depression-relevant categories (depression, anxiety, somatic, and insomnia). GPT-4 achieved the highest accuracy for depression classification for the original (i.e., not clustered) data. B.T and Chen (2024) assessed three LLM-based chatbots: ChatGPT-3.5, ChatGPT-4, and Bard. The objective was to detect Alzheimer's dementia (AD) versus Cognitively Normal (CN) individuals using textual input from spontaneous speech recordings. The approach employed zero-shot learning with two levels of independent queries, where the first consisted of one direct question: “Could the following transcribed speech be from a Cognitive Normal or Alzheimer’s Dementia subject?”, while the second query used Chain-of-Thought (CoT) prompting for more detailed information. The researchers analyzed recordings from the ADReSSo Challenge dataset, which were transcribed using Otter.ai. Aldeen et al. (2023) evaluated ChatGPT's capabilities in data annotation tasks across ten diverse datasets covering various subject areas and a number of classes, among which a Reddit-based mental health dataset can be found. The approach involved testing different GPT models (GPT-3.5 and GPT-4), exploring various prompt strategies, and comparing results against human expert annotations. Liu et al. (2023) assessed GPT-4's performance on text-based applications for radiology reports across various tasks, including Sentence Similarity Classification (Other Annotations Task 1), Natural Language Inference (NLI) (Other Annotations Task 2), Disease Classification (Clinical Decision Support Task 1), and Disease Progression Classification (Clinical Decision Support Task 2). The approach employed different prompting strategies such as zero-shot, few-shot, CoT, and example selection, comparing GPT-4's performance against state-of-the-art (SOTA) radiology-specific models. Among the datasets used for evaluation MS-CXR-T and RadNLI can be mentioned. Kim et al. (2024) demonstrated a successful application of ChatGPT (GPT-3.5-turbo-0613) for analyzing vaccination sentiment in healthcare-related social media content, specifically focusing on HPV vaccination discussions. Their best-performing methodology involved collecting human-evaluated social media messages in short format (SF) (i.e., Twitter) and long format (LF) (i.e., Facebook) about HPV vaccination, inputting them into GPT-3.5, and generating 20 response instances per message to determine the message stance (anti-vaccination, pro-vaccination, or neutral). This approach required no specific pre-training or fine-tuning for the healthcare domain, making it an accessible and efficient tool for researchers analyzing public health discourse on social media. Carneros-Prado et al. (2023) compared GPT-3.5 with IBM Watson for analyzing emotions and sentiments in COVID-19-related healthcare social media data. Their methodology involved processing COVID-19 tweets and using a specific prompt engineering approach where GPT-3.5 was instructed to “rate the sentiment between -1 (negative) and 1 (positive)” and classify emotions into five categories (i.e., Joy, Sadness, Fear, Anger, and Disgust). Without any specific training, GPT-3.5 demonstrated strong performance in detecting nuanced emotional expressions in healthcare-related tweets during the pandemic, particularly showing better capability than IBM Watson in recognizing irony and context-dependent sentiments in COVID-19 discussions. Williams et al. (2024) evaluated GPT-4's ability to assess clinical acuity in Emergency Department (ED) settings using a dataset of adult ED visits from UCSF. The methodology involved creating 10,000 balanced pairs of ED visits with different Emergency Severity Index (ESI) scores, where each pair had contrasting acuity levels (from immediate to nonurgent). GPT-4 was prompted to analyze de-identified text (specifically the chief concern, history of presenting illness, and review of systems sections) from ED physician notes to determine which patient in each pair had higher acuity. Besides performing well on this dataset, results comparable to a resident physician's assessment in a 500-pair subsample were achieved. Sarkar et al. (2023) achieved the best performance for healthcare text classification by using ChatGPT-3.5 with a prompt-based methodology rather than an embedding-based approach. The authors designed a specific prompt structure that included a system setup explaining the task and defined the possible health-related topic categories (e.g., “Addiction”, “Heart Health”, “Mental Health”). The prompt template included the instruction to classify the input medical article into predefined categories and required responses in a specific JSON format {“Topics”: [“List of topics”]}. Guo et al. (2024) developed a novel automated screening methodology using GPT-4 to evaluate titles and abstracts for inclusion/exclusion in clinical systematic reviews. The methodology involved crafting a specific prompt template that outlined the screening instructions, inclusion/exclusion criteria, and the abstract to be evaluated, requiring the model to respond with only “included” or “excluded”. They tested their approach on six review papers, including studies on COVID-19 treatments, Raynaud syndrome, postoperative pain management, and clinical machine learning applications.
\\[0.25cm]
Alsentzer et al. (2023) leveraged Flan-T5-XXL to perform zero-shot classification of Postpartum Hemorrhage (PPH) into four subtypes (i.e., uterine atony (Tone), retained products of conception and placenta accreta spectrum (Tissue), birth or surgical trauma (Trauma), and coagulation abnormalities (Thrombin)) from clinical discharge notes. When discharge summaries exceeded the model's input length limit, they split them into 512-token chunks with 128-token stride, generated predictions for each chunk, and aggregated the results. Raja et al. (2024) explored Bidirectional and Auto-Regressive Transformers (BART) for automatically classifying ophthalmology research papers. Five classifications were conducted, and the best-performing approach used zero-shot learning with BART to analyze titles and abstracts of ophthalmology articles from PubMed. This demonstrated BART's ability to understand complex medical terminology and concepts without requiring additional training data or fine-tuning for specific medical domains. Sivarajkumar and Wang (2022) introduced HealthPrompt, a zero-shot learning framework for clinical text classification using LLMs and prompt engineering, with Clinical BERT showing the best performance. The framework processes clinical texts by first using a chunk encoder to split long medical documents into smaller segments, then applies carefully designed prompt templates (either cloze or prefix prompts) to transform the input text into a classification. Chang et al. (2024) evaluated clinical LLMs (i.e., Med42-70B) for automatically classifying cancer staging from pathology reports using a Zero-Shot CoT (ZS-CoT) prompting strategy. The methodology involved feeding unstructured pathology report text into Med42-70B along with a system prompt asking for cancer staging review, followed by a “Let's think step by step” instruction that triggered the model to generate reasoning steps before making the final TNM (Tumor, Node, Metastasis) classification.

\begin{landscape}
\begin{table}[h]
\captionsetup{labelformat=empty}
\centering
\caption*{\textbf{Table 2.} Prompt Engineering-based Reviewed Literature Categorization}
\label{tab:classification_models}
\begin{tabular}{c c c c c c c c c c c}
 \toprule
\multirow{2}{*}{\centering\textbf{Reference}} & 
\multirow{2}{*}{\centering\textbf{Type}} & 
\multirow{2}{*}{\centering\textbf{Language}} & 
\multirow{2}{*}{\centering\textbf{Best LLM}} & 
\multicolumn{2}{c}{\multirow{2}{*}{\centering\textbf{Application}}} & 
\multicolumn{5}{c}{\centering\textbf{Performance Evaluation Metrics}} \\
\cmidrule(lr){7-11}
& & & & & & \textbf{Accuracy} & \textbf{F1 Score} & \textbf{Precision} & \textbf{Recall} & \textbf{AUC Score} \\

\hline
\makecell{\textbf{Sushil et al.} \\ (2024)} & Journal & English & GPT-4 & \multicolumn{2}{c}{\makecell{Information \\ Extraction}} & - & Macro: 86\% &	-	& -	& - \\
\hline
\makecell{\textbf{Lossio-Ventura} \\ \textbf{et al.} (2024)} & Journal & English & GPT-3.5 &  \multicolumn{2}{c}{\makecell{Public Health and \\ Opinion Analysis}} & \makecell{NIH: \\ 86\% \\ Stanford: \\ 87.40\%} & \makecell{NIH: \\ 86.68\% \\
Stanford: \\ 86.62\%} & \makecell{NIH: \\  89.26\% \\ Stanford: \\ 87.79\%}	& \makecell{NIH: \\ 85.26\%  \\
Stanford: \\ 86.32\%}	 & - \\
\hline
\makecell{\textbf{Chen et al.} \\ (2024a)} & Journal & English & \makecell{GPT-3.5- \\ turbo- \\ 0613} &  \multicolumn{2}{c}{\makecell{Research/Literature \\ Analysis}} & 90.75\%	& -	& -	& -	& - \\
\hline
\makecell{\textbf{Ohse et al.} \\ (2024)} & Journal & English & GPT-4 &  \multicolumn{2}{c}{\makecell{Clinical Decision \\ Support}} & -	& 73\%	& 72\% &	78\%	& - \\
\hline
\makecell{\textbf{BT and Chen} \\ (2024)} & Journal & English & \makecell{AD: BARD \\ CN: GPT-4} &  \multicolumn{2}{c}{\makecell{Clinical Decision \\ Support}} & \makecell{AD: 58\% \\
CN: 41\%} &	\makecell{AD: 71\% \\
CN: 61\%} &	\makecell{AD: 89\% \\
CN: 53\%}	& \makecell{AD: 60\% \\
CN: 73\%}	& - \\

\hline
\makecell{\textbf{Kim et al.} \\ (2024)} & Journal & English & \makecell{GPT-3.5- \\ turbo- \\ 0613} &  \multicolumn{2}{c}{\makecell{Public Health and \\ Opinion Analysis}} & \makecell{LF: 76.1\% \\
SF: 70.7\%} &	- &	- &	-	 & - \\

\hline
\makecell{\textbf{Guo et al.} \\ (2024)} & Journal & English & \makecell{GPT-3.5/ \\ GPT-4} &  \multicolumn{2}{c}{\makecell{Research/Literature \\ Analysis}} & 91\%	& Macro: 60\%	& \makecell{Included: \\ 76\% \\
Excluded: \\ 91\%}	& -	& - \\

\hline
\multirow{5}{*}{\makecell{\textbf{Raja et al.} \\ (2024)}} & 
\multirow{5}{*}{\centering Journal} & 
\multirow{5}{*}{\centering English} & 
\multirow{5}{*}{\centering BART} & 
\multirow{5}{*}{\makecell{ Research/ \\ Literature \\ Analysis}} & 
Article Type & 91\% & 92\% & 91\% & 93\% & 91\% \\
\cmidrule(lr){6-11}
 &  &  &  &  & Ocular Diseases & 85\% & 85\% & 86\% & 89\% & 92\% \\
\cmidrule(lr){6-11}
 &  &  &  &  & Automated Studies & 92\% & 92\% & 92\% & 94\% & 95\% \\
\cmidrule(lr){6-11}
 &  &  &  &  & DEye & 63\% & 79\% & 63\% & 67\% & 60\% \\
\cmidrule(lr){6-11}
 &  &  &  &  & DemL & 99\% & 99\% & 99\% & 99\% & 98\% \\
\hline
\makecell{\textbf{Chang et al.} \\ (2024)} &	Conference &	English	& Med42-70B &	\multicolumn{2}{c}{\makecell{Clinical Decision \\ Support}} &	- &	\makecell{T (Macro): \\ 78\% \\
N (Macro): \\ 82\% \\
M (Macro): \\ 62\%}	 & \makecell{T (Macro): \\ 77\% \\
N (Macro): \\ 81\% \\
M (Macro): \\ 62\%}	 & \makecell{T (Macro):  \\ 80\% \\
N (Macro): \\ 84\% \\
M (Macro): \\ 62\%	} & - \\
\hline
\makecell{\textbf{Williams et al.} \\ (2024)} &	Journal	& English	& GPT-4	& \multicolumn{2}{c}{\makecell{Clinical Decision \\ Support}} &	\makecell{10,000 \\ pairs: 89\% \\
500 pairs: \\ 88\%} &	- &	-	& - &	- \\
\bottomrule
\end{tabular}
\end{table}
\end{landscape}

\begin{landscape}
\begin{table}[h]
\captionsetup{labelformat=empty}
\centering
\caption*{\textbf{Table 2.} Cont.}
\label{tab:classification_models}
\begin{tabular}{c c c c c c c c c c}
 \toprule
\multirow{2}{*}{\centering\textbf{Reference}} & 
\multirow{2}{*}{\centering\textbf{Type}} & 
\multirow{2}{*}{\centering\textbf{Language}} & 
\multirow{2}{*}{\centering\textbf{Best LLM}} & 
\multirow{2}{*}{\centering\textbf{Application}} & 
\multicolumn{5}{c}{\centering\textbf{Performance Evaluation Metrics}} \\
\cmidrule(lr){6-10}
& & & & & \textbf{Accuracy} & \textbf{F1 Score} & \textbf{Precision} & \textbf{Recall} & \textbf{AUC Score} \\

\hline
\makecell{\textbf{Carneros-Prado} \\  \textbf{et al.} (2023)} & Conference & English & GPT-3.5 & \makecell{Public Health and Opinion \\  Analysis} & - & - &	-	& -	& - \\

\hline
\makecell{\textbf{Sarkar et al.} \\ (2023)} & Conference	& English &	ChatGPT-3.5	& \makecell{Research/ \\ Literature Analysis}	& - &	60.60\% &	-	& - &	- \\

\hline
\makecell{\textbf{Alsentzer et al.} \\ (2023)} & Journal	& English	& \makecell{Flan-T5- \\ XXL}	& Clinical Decision Support	& \makecell{Tone: \\  95.70\% \\ 
Tissue:  \\ 91.80\% \\ 
Trauma:  \\ 80.80\% \\ 
Thrombin:  \\ 95.20\%} &	\makecell{Tone:  \\ 94.70\% \\ 
Tissue:  \\ 88.70\% \\ 
Trauma: \\  74\% \\ 
Thrombin:  \\ 50\%}	& \makecell{Tone:  \\ 93.10\% \\ 
Tissue:  \\ 82.70\% \\ 
Trauma:  \\ 67.10\% \\ 
Thrombin:  \\ 38.50\%} &	-	& - \\

\hline
\textbf{Shi et al.} (2023)	& Conference	& English	& \makecell{ChatGraph \\ (w/ TF-IDF)} &	\makecell{Research/Literature \\ Analysis}	& 63.63\%	& - &	- &	-	& - \\

\hline
\makecell{\textbf{Aldeen et al.} \\ (2023)}	& Conference &	English	& GPT-4	& \makecell{Public Health and Opinion \\ Analysis}	& 89.9\%	& 89.8\%	& -	& - &	- \\

\hline
\multirow{4}{*}{\makecell{\textbf{Liu et al.} \\ (2023)}} &	\multirow{4}{*}{\makecell{Conference}}	& \multirow{4}{*}{\makecell{English}} &	\multirow{4}{*}{\makecell{GPT-4}} &  \makecell{Clinical Decision Support \\ Task 1} & -	& \makecell{Micro: \\ 97.9\% \\
Macro: \\ 97.5\%} &	- &	-	& - \\
\cmidrule(lr){5-10}
 & & & & \makecell{Clinical Decision Support \\ Task 2} & \makecell{Macro \\
Pl. eff.:  \\ 98.7\% \\ 
Cons.: \\  95.7\% \\ 
PNA:  \\ 96.4\% \\ 
PTX:  \\ 99.4\% \\ 
Edema:  \\ 96.8\%}	& - &	- &	- &	- \\
\cmidrule(lr){5-10}
 & & & & \makecell{Other Annotations Task 1} & \makecell{MS-CXR-T: \\ 97.2\% \\
RadNLI:  93.8\%} & 	-	& - &	- &	- \\
\cmidrule(lr){5-10}
 & & &  & \makecell{Other Annotations Task 2} & - &	RadNLI: 89.2\% &	- &	- &	- \\

\hline
\makecell{\textbf{Sivarajkumar} \\ \textbf{and Wang} \\ (2022)}	& Conference &	English	& \makecell{HealthPrompt \\ + Clinical \\ BERT} &	Clinical Decision Support &	85\%	& 86\% &	86\%	& 86\% &	- \\
	
\bottomrule
\end{tabular}
\end{table}
\end{landscape}

\begin{landscape}
\begin{table}[h]
\captionsetup{labelformat=empty}
\centering
\caption*{\textbf{Table 3.} Text Classification and LLM Type Categorization: The Case of Prompt Engineering}
\label{tab:classification_models}
\begin{tabular}{c c c c c c c c c}
 \toprule
\multirow{2}{*}{\centering\textbf{Reference}} & 
\multicolumn{3}{c}{\makecell{\textbf{Text Classification Type}}} & 
\multicolumn{5}{c}{\makecell{\textbf{Large Language Model Type}}}
\\
\cmidrule(lr){2-9}

& \textbf{Multi-class}	& \textbf{Multi-label}	& \textbf{Binary} & \textbf{	BERT (or variant)}	& \textbf{Closed-source} &	\textbf{Open-source}	& \makecell{\textbf{Pre-trained} \\ \textbf{Transformer}}	& \textbf{BART} \\

\hline
\makecell{\textbf{Sushil et al.} \\ (2024)} & \checkmark  &  \checkmark  & & &  \checkmark  &  &  &  \\
\hline
\makecell{\textbf{Lossio-Ventura et} \\ \textbf{al.} (2024)} & \checkmark  & & & &  \checkmark &   &  &  \\
\hline
\textbf{Chen et al.} (2024a) & \checkmark  & & & &  \checkmark  &  &  &  \\
\hline
\textbf{Ohse et al.} (2024) & &  & \checkmark & &  \checkmark  &  &  &   \\
\hline
\makecell{\textbf{BT and Chen} \\ (2024)} & \checkmark  &  & & &  \checkmark  &  &  &  \\
\hline
\textbf{Kim et al.} (2024) & \checkmark  & &  & &  \checkmark  &  &  &  \\
\hline
\textbf{Raja et al.} (2024) & \checkmark  & \checkmark &   & &  &  &  &  \checkmark  \\
\hline
\textbf{Guo et al.} (2024) & & & \checkmark & &  \checkmark  &  &  &  \\
\hline
\textbf{Chang et al.} (2024) & \checkmark  &  & & &  & \checkmark  &  &  \\
\hline
\makecell{\textbf{Williams et al.} \\ (2024)} & & & \checkmark  & &  \checkmark  &  &  &   \\
\hline
\makecell{\textbf{Alsentzer et al.} \\ (2023)} & \checkmark  & & & &  &  & \checkmark  &   \\
\hline
\textbf{Sarkar et al.} (2023) & &  \checkmark  & & &  \checkmark  &  &  &  \\
\hline
\textbf{Shi et al.} (2023) & \checkmark  & & & &  \checkmark  &  &  &  \\
\hline
\textbf{Aldeen et al.} (2023) & &  & \checkmark & &  \checkmark  &  &  &  \\
\hline
\textbf{Liu et al.} (2023) & \checkmark  & & \checkmark & &  \checkmark  &  &  &  \\
\hline
\makecell{\textbf{Carneros-Prado et} \\ \textbf{al.} (2023)} & \checkmark  & & & &   \checkmark &   &  &  \\
\hline
\makecell{\textbf{Sivarajkumar and} \\ \textbf{Wang} (2022)} & \checkmark  & &  & \checkmark &   &  &  &  \\

\bottomrule
\end{tabular}
\end{table}
\end{landscape}

\subsubsection{Pre-training, Fine-tuning, and Prompt Tuning}
\paragraph{Pre-training}\mbox{}\\[-1.5\baselineskip]

Pre-training or further pre-training an LLM for healthcare text classification involves adapting the model's language understanding to the specific healthcare domain. This approach enhances the model's ability to process specialized terminology, abbreviations, and context found in healthcare texts, improving classification performance. Pre-training typically occurs in two phases: the general transformer-based architecture (or existing LLM) is first trained on unlabeled curated data, and then further pre-training occurs using domain-specific datasets like MIMIC-III or PubMed. Pre-training is usually followed by fine-tuning. Fine-tuning then focuses on task-specific objectives, such as classifying patient diagnoses, symptoms, or treatments from clinical notes. This process allows the model to capture domain nuances, improving its ability to generalize and reduce error rates in healthcare-specific tasks. Multiple existing healthcare-related LLMs result from further pre-training of existing general-purpose LLMs such as BERT. These LLMs are referred to in this systematic review as BERT variants. For example, Lee et al. (2020) introduced BioBERT, a BERT model further pre-trained on biomedical text from PubMed and PubMed Central (PMC), demonstrating improved performance on Named Entity Recognition (NER) and Question-Answering tasks in the medical domain. Similarly, Alsentzer et al. (2019) adapted BERT to clinical notes, releasing ClinicalBERT, which showed significant improvements in healthcare NLP tasks such as NLI, which is a classification task.
\\[0.25cm]
Yang et al. (2022) developed GatorTron-large, an 8.9 billion parameter LLM specifically trained for healthcare applications using over 82 billion words of de-identified clinical text from UF Health's electronic health records, combined with additional medical text from PubMed, MIMIC-III, and Wikipedia. The model used a BERT-style architecture with 56 layers, 3584 hidden units, and 56 attention heads, trained using two self-supervised tasks, not requiring prior manual labeling, which are Masked Language Modeling (MLM) and Sentence-Order Prediction (SOP), as objectives. GatorTron-large demonstrated superior performance on different clinical NLP tasks critical for healthcare applications, among which NLI (to determine logical relationships between clinical statements) can be found. The model's extensive training on real clinical narratives from diverse healthcare settings (inpatient, outpatient, emergency) across various clinical departments enabled it to better understand the healthcare text data. McMaster et al. (2023) leveraged DeBERTa to automatically detect Adverse Drug Reactions (ADRs) in hospital discharge summaries. Their best-performing model, MeDeBERTa, used a two-stage training process: first, the base DeBERTa model was further pre-trained on a large amount of unlabeled clinical documents from the studied hospital to adapt it to their institutional context. Then, fine-tuning was carried out on this pre-trained model on annotated discharge summaries enriched with validated ADR cases to help the model distinguish between true ADRs and other drug-related adverse events. Li et al. (2023b) developed two clinical domain-specific language models, Clinical-Longformer and Clinical-BigBird, designed to handle long clinical texts from EHR. Starting with pre-trained Longformer and BigBird models, additional pre-training was performed using clinical notes from the MIMIC-III. The pre-training process enabled the models to extend their maximum input sequence length from 512 to 4096 tokens, allowing better capture of long-term dependencies in clinical narratives. Both models employed sparse attention mechanisms combining sliding windows and global attention to reduce computational costs. The pre-trained LLMs were evaluated on different NLP tasks following their respective fine-tuning. As a result, Clinical-Longformer outperformed conventional LLMs (e.g., ClinicalBERT). Bressem et al. (2020) leveraged BERT for classifying radiology text reports, specifically focusing on chest radiograph reports. The authors additionally pre-trained the German BERT base model on radiology reports from their institution, resulting in proposing RAD-BERT. Subsequently, this LLM was fine-tuned (i.e., FT RAD-BERT) to classify various radiological findings (e.g., congestion, effusion, consolidation, and pneumothorax). Blinov et al. (2020) focused on classifying clinical diagnoses from EHR text data using a modified BERT-based model called RuPool-BERT for an adaptation to Russian. The proposed methodology involved two key ideas. First, modifying RuBERT’s architecture (standard BERT architecture adapted for Russian) by concatenating the classification token output with two additional components, max and mean pooling over the last encoder states, prior to passing it through the final classification layer. Second, developing a domain-specific version (i.e., RuEHR-BERT) by pre-training BERT on medical records using MLM and a custom medical tokenizer trained on the same data. The resulting LLMs, RuPool-BERT and RuEHR-BERT outperformed baseline models, with RuPool-BERT having a slight edge in performance. It is worth noting that further pre-training of an LM is typically followed by fine-tuning the resulting model.

\paragraph{Fine-tuning}\mbox{}\\[-1.5\baselineskip]

Fine-tuning LLMs involves adapting pre-trained models to specific tasks or domains by training them on additional data. This method is, in some cases, carried out similarly to the conventional supervised learning paradigm since labeled data is required. This process builds on the LLM's acquired knowledge during pre-training while narrowing its focus to the nuances of a particular application. Fine-tuning is especially valuable in scenarios where the general-purpose language capabilities of LLMs need refinement to capture domain-specific jargon, structure, or patterns, such as in the case of healthcare. The main benefit of fine-tuning LLMs for healthcare is its ability to improve classification performance across complex text inputs without requiring massive amounts of labeled data from scratch. Additionally, fine-tuning provides more flexibility to healthcare-specific needs, such as identifying diagnoses and treatments or extracting structured information from unstructured text. The adaptation is done by updating the LLM’s parameters with task-specific data. There are several techniques used for fine-tuning. For instance, standard fine-tuning is the traditional method that updates all the LLM’s parameters for the downstream task (Devlin et al., 2018). It is highly effective but computationally expensive and requires a large amount of labeled data. There are also more efficient methods, Parameter-Efficient Fine-Tuning (PEFT). These methods selectively fine-tune a subset of the model’s parameters, reducing computational costs while maintaining performance. Adapter Layers and Low-Rank Adaptation (LoRA) can be cited among the developed PEFT techniques. Recently, LoRA particularly gained attraction and has been widely adopted for resource-constrained environments (Hu et al., 2021). Each fine-tuning approach aims to balance efficiency, generalization, and computational costs based on the specific use case and available resources. As a result, the model retains the general linguistic understanding gained in the pre-training phase while adapting to the particularities of healthcare terminology, abbreviations, and context-sensitive phrases.
\\[0.25cm]
Fine-tuning emerged as the most used LLM-based approach in the reviewed literature compared to pre-training and prompt engineering, contributing to the best-performing methodologies in 36 out of 65 research studies. Its popularity is due to its ability to balance computational efficiency and task-specific performance since it leverages pre-trained model foundations while adapting to healthcare data and text classification tasks. This leads to higher performance without the need for training an architecture from scratch or using potentially weaker prompt-based methods. Moreover, including early PLMs (e.g., BERT and GPT-2) since 2018 in this systematic review means that researchers had extensive time to experiment with these models. Besides, their smaller size made them more adequate for fine-tuning, given the lower computational requirements.
\\[0.25cm]
Ohse et al. (2024) evaluated LLMs for depression detection from clinical interview transcripts, with GPT-3.5 (after fine-tuning) achieving the highest F1 score. The methodology involved first collecting interviews from participants using the GRID-HAMD-17 protocol, with participants also completing the PHQ-8 questionnaire as a baseline measure of depression. The interviews were recorded, transcribed from German to English using the Whisper model, and then processed by segmenting the text into four clinically relevant clusters: depression, anxiety, somatic, and insomnia. The GPT-3.5 model was fine-tuned using 12 interview samples with the highest error rates in initial testing, with the model being trained to predict PHQ-8 scores from the clustered interview text. This fine-tuned approach significantly outperformed zero-shot implementations of other models (including GPT-4 and Llama2) in accurately detecting depression from clinical interview transcripts. Schneider et al. (2021) developed GPT2-Bio-Pt, a Portuguese language model specifically designed for biomedical and clinical text analysis, by fine-tuning an existing Portuguese GPT-2 model (GPorTuguese-2) on a large corpus of biomedical literature containing PubMed and Scielo databases. They evaluated the model's performance on a healthcare-specific task: detecting patient fall events in de-identified clinical progress notes from a Brazilian hospital. As a result, the fine-tuned GPT2-Bio-Pt achieved the best performance. It is worth noting that GPT-2 is not a fully open-source LLM and, therefore, is considered a closed-source LLM in this review paper.  
\\[0.25cm]
Xu et al. (2024) presented Mental-Alpaca and Mental-FLAN-T5, two instruction fine-tuned LLMs for mental health analysis using social media text data. The best-performing approach involved instruction fine-tuning base LLMs (i.e., Alpaca and FLAN-T5) simultaneously on multiple mental health datasets from social media platforms (primarily Reddit) covering different conditions, including stress, depression, and suicide risk. The finetuning process merged multiple datasets and tasks together in a single training iteration, allowing the models to learn various mental health prediction tasks concurrently. This approach enabled the models to outperform much larger models like GPT-3.5 and GPT-4 on tasks such as detecting signs of depression, stress, and suicide risk from user-generated text. The resulting two LLM-based approaches achieved higher performances when compared to other models, each on a list of tasks/datasets. Guevara et al. (2024) evaluated different configurations of fine-tuned Flan-T5 models to automatically extract Social Determinants of Health (SDoH) from clinical notes in EHR. The PEFT method, LoRA, was used for this purpose. The models were trained to identify six key SDoH categories: employment status, housing issues, transportation issues, parental status, relationship status, and social support. Using a dataset of clinical notes from cancer patients, the researchers experimented with different model architectures and synthetic data augmentation approaches. The addition of synthetic data, generated using GPT-3.5, proved beneficial primarily for smaller Flan-T5 models and rare SDoH categories with limited training examples. Gretz et al. (2023) achieved the best performance for healthcare text classification by fine-tuning Flan-T5-XXL using a leave-one-fold-out setup on healthcare datasets. This was done using LoRA for efficiency for three epochs with early stopping based on development set performance. Kementchedjhieva and Chalkidis (2023) proposed T5Enc, which is a T5 LM fine-tuned in a non-autoregressive fashion. The model was tasked with assigning multiple Medical Subject Headings (MeSH) terms to biomedical articles and ICD-9 diagnostic codes to clinical notes. The fine-tuning process involved using the Adafactor optimizer with a fixed learning rate after a warmup for one epoch, and the model was trained to minimize cross-entropy loss on the binary classification task for each medical label.
\\[0.25cm]
Li et al. (2024b) developed LlamaCare, a clinical domain-adapted LM created by instruction fine-tuning Llama 2 (7B chat version) on healthcare text data. Their methodology involved a two-step process. First, they used GPT-4 to generate diverse clinical instructions based on seed prompts for different medical service types (e.g., Radiology, Respiratory, Rehab). Second, they extracted corresponding input-output pairs from the MIMIC-III clinical database, which contains various medical notes, including discharge summaries, ECG reports, and nursing notes. The model was fine-tuned using LoRA to efficiently adapt Llama 2 to clinical tasks while minimizing computational resources. The proposed approach was tested on various tasks, among which Mortality (MOR) Prediction, Length of Stay (LOS) Prediction, Diagnoses Prediction (DIAG), and Procedures Prediction (PROC) can be categorized as text classifications. Bumgardner et al. (2024) demonstrated the successful application of fine-tuned LLaMA models for extracting structured condition codes from pathology reports in a healthcare setting, with the larger 13B parameter version showing superior overall performance. The researchers trained Path-LLaMA using surgical pathology reports containing gross descriptions, final written diagnoses, and ICD condition codes from clinical workflows at the University of Kentucky. They formatted their training data as instruction-based conversations in JSON format, where each pathology case's text was concatenated into a single input field with associated ICD codes as the model response. The researchers focused specifically on cases with cancer-related codes. Wang et al. (2024) study developed DRG-LLaMA, a fine-tuned version of the LLaMA language model, to predict Diagnosis-Related Groups (DRGs) from clinical discharge summaries in the MIMIC-IV dataset. The authors specifically extracted the “brief hospital course” section from discharge summaries and fine-tuned LLaMA using LoRA. The best-performing model used a 13B parameter version of LLaMA with a maximum input context length of 1024 tokens.  
\\[0.25cm]
Lehman et al. (2023) evaluated multiple LMs on three clinical tasks using healthcare text data from EHRs. Two of these three tasks are classification-based: NLI and the Identification of Follow-up Information in discharge summaries. The best-performing approach utilized BioClinRoBERTa followed by task-specific finetuning. This model achieved superior performance compared to larger general-purpose models, including GPT-3, demonstrating that domain-specific pre-training on the clinical text was more valuable than model size alone for this healthcare application. As previously discussed in the prompt engineering section, Raja et al. (2024) developed a text classification framework for ophthalmology research papers using the BART in a zero-shot learning approach. However, for clinical studies subclass grouping where the initial BART model had lower performance, they fine-tuned BioBERT, which performed better. Savage et al. (2023) developed and validated a fine-tuned BioMed-RoBERTa model to screen patients for appropriate Best Practice Alerts (BPAs) in EHRs, specifically focusing on identifying patients who should receive deep vein thrombosis prophylaxis. The model was fine-tuned using history and physical notes from the MIMIC-III database, where the notes were from patients who did not receive anticoagulation at admission. Two physicians labeled these notes as either indicating active bleeding or no active bleeding. The fine-tuned model was trained to perform binary classification of clinical notes to identify patients without active bleeding who would be appropriate for thromboembolism prophylaxis alerts. The researchers used a development set of notes to optimize hyperparameters and a separate test set of notes for evaluation. It is worth mentioning that due to token limits, patient notes were truncated to 2000 characters. Shiju et al. (2022) demonstrated that Bio\_ClinicalBERT best-classified drug reviews from Drugs.com when fine-tuned on a binary classification task. The methodology involved fine-tuning Bio\_ClinicalBERT on a dataset of drug reviews using a binary classification approach where ratings $ \geq $ 8 were considered "above average" and $ \textless $ 8 were considered "below average". Xie et al. (2024) developed an epilepsy specific LLM approach to identify seizure outcomes from clinical notes by fine-tuning Clinical\_BERT using manually annotated epileptologist notes. The model was trained to classify each clinical visit note as either “seizure-free” (no seizures since the last visit or within the past year) or “having recent seizures”. The fine-tuning process involved using a plurality voting system where model predictions were repeated 5 times using different random seeds. Chen et al. (2022) proposed LitMC-BERT, a transformer-based multi-label classification model specifically designed for biomedical literature classification. The architecture uses BioBERT as its shared transformer backbone. It introduces two novel components: label-specific modules that capture unique features for each medical topic (e.g., Treatment, Diagnosis, Prevention) through multi-head self-attention and a label pair module that models relationships between co-occurring medical topics through co-attention mechanisms. The training process involves multi-task training where the model simultaneously learns to predict labels and their co-occurrences, followed by fine-tuning of just the Label Module while keeping other components frozen. This approach demonstrated effectiveness in the medical domain by outperforming existing methods in classifying both COVID-19 literature and cancer research papers (i.e., HoC). Cui et al. (2023) focused on classifying temporal information about medical treatments in hospital discharge summaries, specifically determining whether treatments occurred during hospitalization periods. The best-performing approach used BERT with traditional fine-tuning. The methodology involved preprocessing clinical text data by extracting relevant sentences using a window-based approach (three sentences before and two after the target treatment mention), along with admission and discharge dates. The fine-tuning process adapted BERT's pre-trained weights to learn temporal relationships between medical treatments and hospitalization periods, helping healthcare professionals automatically track when specific medical interventions occurred during a patient's hospital stay. Van Ostaeyen et al. (2023) fine-tuned RobBERT, a Dutch BERT-based language model, to automatically analyze written feedback in healthcare education settings. Using a dataset of labeled feedback comments (split into sentences) collected from five healthcare educational programs (i.e., specialistic medicine, general practice, midwifery, speech therapy, and occupational therapy), the authors trained multiclass-multilabel classification models to identify both feedback quality criteria and Canadian Medical Education Directions for Specialists (CanMEDS) roles in the healthcare text. The fine-tuning process involved tokenizing the healthcare feedback data, padding/truncating sentences to a fixed length of 512 tokens, and optimizing hyperparameters through five optimization runs on a development dataset. Ren et al. (2023) showed the successful application of fine-tuned BERTweet, a RoBERTa-based language model pre-trained on Twitter data, for classifying Patient Portal Messages (PPMs) in healthcare settings. The model was fine-tuned on 2,239 annotated PPMs from three clinical departments (i.e., Cardiology, Gastroenterology, and Dermatology) to classify messages into four categories: Active Symptoms, Prescriptions, Logistics, and Update/Other. The fine-tuning process involved training for four epochs using the Adam optimizer and cross-entropy loss function. Chen et al. (2024b) evaluated GPT-4 using few-shot prompting with CoT reasoning to tackle biomedical text classification. They analyzed medical research literature to categorize health advice sentences into “no advice”, “weak advice”, and “strong advice” categories across different sections of medical papers (Discussion sections, Structured abstracts, and Unstructured abstracts). In parallel, BioBERT was fine-tuned in a standard supervised manner and outperformed the previously mentioned approach. Silverman et al. (2024) used two variants of UCSF-BERT, a clinical language model pre-trained on millions of clinical notes from UCSF's electronic health records, to identify Serious Adverse Events (SAEs) from Inflammatory Bowel Disease (IBD) clinical notes. The base hierarchical model (H-UCSF-BERT) could process longer sequences of up to 2,560 tokens by combining chunk representations through an additional transformer layer and performed best for detecting if a medication was mentioned/given prior to a hospitalization event (Task 1). A modified version (H-UCSF-BERT + only nearby SAEs), which restricted analysis to SAEs mentioned within a two-sentence window of hospitalization events, achieved the best results for identifying if a hospitalization was caused by/related to an adverse event (Task 2) and identifying if there was a medication given before a hospitalization that was caused by an adverse event (Task 3). Chen et al. (2023) used a standard supervised fine-tuning approach on task-specific biomedical datasets from the Biomedical Language Understanding \& Reasoning Benchmark (BLURB) benchmark, among which literature classification can be found. The model considered was BioLinkBERT-Large. Bețianu et al. (2024) proposed DALLMi for effective domain adaptation for healthcare text classification using BERT as the base LLM model. DALLMi employed a semi-supervised fine-tuning technique that combines three key components: a label-balanced sampling strategy ensuring at least one positive sample per label in each batch, a novel variational loss function that leverages both labeled and unlabeled medical text data, and a MixUp regularization technique that performs interpolation at the word embedding level of BERT to generate synthetic training samples. Farruque et al. (2024) developed a semi-supervised learning approach using Mental-BERT to detect depression symptoms from social media text data. The methodology involved first fine-tuning Mental-BERT on a clinician-annotated dataset of depression-related tweets, where tweets were labeled with specific depression symptoms according to guidelines. The researchers then employed an iterative data harvesting approach where this initial fine-tuned model, coupled with a Zero-Shot learning model (USE-SE-SSToT), was used to automatically label more depression-related tweets from a curated repository of tweets from self-disclosed depressed users. These newly labeled tweets are then used to retrain the model, creating an expanded training dataset that maintains the clinical distribution of depression symptoms. The research study includes the evaluation of two classification tasks: Depression Symptoms Detection (DSD) and Depression Post Detection (DPD). Pan et al. (2020) presented Feature-level Attention for Multi-label Classification on BERT (FAMLC-BERT), a fine-tuned BERT-based model designed to predict multiple medical diagnoses from unstructured clinical free text in EHRs. The methodology involved fine-tuning the BERT base model using a feature-level attention mechanism that captures semantic features from different BERT encoder layers. The attention mechanism assigns different weights to the [CLS] token embeddings from each of BERT's 12 encoder layers, allowing the model to focus on the most relevant semantic features for disease prediction. The fine-tuning process used binary cross-entropy loss and the Adam optimizer, with the input text being preprocessed to handle Chinese characters and standardize special tokens like dates and numbers. Bansal et al. (2023) used a fine-tuned DeBERTa Large model to classify healthcare-related social media posts, specifically tweets expressing concerns about COVID-19 vaccines. The model, which achieved the best performance, was fine-tuned using the CAVES dataset containing tweets. What makes this application particularly relevant for healthcare is its ability to automatically categorize vaccine-related concerns into 12 distinct categories, including side effects, efficacy, rushed development, and religious reasons. This enables public health officials to better monitor and address vaccine concerns promptly. Uslu et al. (2024) focused on using a BERT variant for classifying chest X-ray radiology reports from the MIMIC-CXR dataset into 14 distinct medical findings. The best-performing approach utilized CXR-BERT-GENERAL after fine-tuning it on the ‘FINDINGS’ section from radiological reports using binary cross-entropy loss and the Adam optimizer. Qi et al. (2023) developed SaFER, a two-stage fine-tuning framework for BERT that was successfully applied to biomedical literature mining tasks. The fine-tuning approach consists of two key stages. First, they fine-tune BERT using a label-agnostic early stopping strategy based on Local Intrinsic Dimensionality (LID) scores to determine the optimal stopping point before overfitting occurs. Second, they implement contrastive learning with a projection head alongside the classifier, using SimCSE to generate positive pairs and apply structural loss to maintain consistency between the classifier and the outputs. The proposed approach was evaluated based on two tasks that are Research/Literature Analysis related but were not explicitly described. Ciobotaru and Dinu (2023) used a fine-tuned Romanian BERT model to analyze public sentiment around COVID-19 vaccination in Romanian tweets. Specifically, they fine-tuned the romanian-bert-cased model by adding a linear layer with a softmax activation on top of the base model, training it for ten epochs using cross-entropy loss and AdamW optimizer. The model was trained on the SART dataset containing tweets. Kersting et al. (2023) developed an efficient approach for analyzing German physician reviews using XLM-RoBERTa-large, fine-tuned for Aspect-Based Sentiment Analysis (ABSA). The authors combined three traditionally separate steps (i.e., aspect term extraction, aspect category classification, and aspect polarity classification) into a single model through token classification. The model was fine-tuned on annotated physician review data to simultaneously identify relevant aspects (e.g., physician friendliness, competence, or waiting times), classify them into predefined categories, and determine their sentiment polarity. The methodology was evaluated on four datasets (i.e., A, B, C, and D). Yogarajan et al. (2021) explored the use of fine-tuned LLMs for automated medical code prediction from EHRs. PubMedBERT and BioMed-RoBERTa-base achieved the best results among the LLM approaches tested. The methodology involved fine-tuning these pre-trained models on medical text data from two distinct EHR databases (MIMIC-III and eICU) for multi-label classification of ICD-9 codes. Specifically, the researchers used standard fine-tuning on all layers without freezing, employing the Adam optimizer.
\\[0.25cm]
Ge et al. (2023) developed a fine-tuned LLM approach for classifying diabetes-related patient questions into six categories (i.e., diagnosis, treatment, common knowledge, healthy lifestyle, epidemiology, and others). Their methodology leveraged the Baichuan2-13B model and employed a two-stage fine-tuning process. First, they fine-tuned the model on a broader medical dataset (PromptCBLUE) using LoRA technique, then performed transfer learning on the specific diabetes question dataset. Tan et al. (2023) developed a fine-tuned LLM-based approach to automatically classify cancer disease responses from radiology reports. Specifically, they used the GatorTron transformer model and fine-tuned it on computed tomography reports from cancer patients, manually annotated into four categories: no evidence of disease, partial response, stable disease, or progressive disease. The best-performing approach combined traditional fine-tuning with a novel data augmentation technique using sentence permutation, where they generated synthetic training examples by randomly reordering sentences in the radiology reports while maintaining the same label. Yogarajan et al. (2022a) achieved the best results in automatic ICD-9 code prediction from clinical notes using TransformerXL, an LM capable of handling longer sequences. The fine-tuning was carried out using binary cross-entropy loss and the Adam optimizer without freezing any parameters. Yogarajan et al. (2022b) research’s best-performing LLM-based approach used TransformerXL to predict COVID-19 patient shielding from EHRs, specifically identifying patients who are clinically extremely vulnerable to coronavirus based on their medical conditions. The methodology involved fine-tuning TransformerXL on medical text data from the MIMIC-III dataset containing discharge summaries. The model was fine-tuned end-to-end on all layers using the Adam optimizer, with a sigmoid activation function for multi-label classification of ICD-9 medical codes that indicate high COVID-19 risk. Luo et al. (2024) proposed Taiyi, a bilingual fine-tuned LLM for biomedical NLP tasks. The methodology involved fine-tuning the Qwen-7B-base using a comprehensive collection of 140 biomedical text datasets (102 English and 38 Chinese) spanning over 10 task types, among which text classification can be found. The key innovation in their fine-tuning approach was a two-stage supervised instruction fine-tuning strategy using QLoRa (Quantized LoRA). First, fine-tuning on non-generation tasks (like information extraction) and tasks with smaller datasets, then combining all data for a second stage of fine-tuning that included question-answering and dialogue tasks. The training data was carefully curated and standardized using consistent schema templates, with particular attention paid to selecting high-quality healthcare datasets.

\paragraph{Prompt-tuning}\mbox{}\\[-1.5\baselineskip]

Prompt-tuning bridges the gap between prompt engineering and fine-tuning. On the one hand, unlike prompt engineering, which involves manually designing prompts to maximize the LLM’s efficiency and achieve the desired outputs, prompt-tuning uses trainable soft prompts (Lester et al., 2021). Soft prompts are the mechanism used in prompt-tuning, where instead of manually crafting prompts, as in traditional prompt engineering, the prompts are represented as learnable vectors. These vectors are trained along with the model on a specific task. These soft prompts guide the LLM in conducting healthcare text classification without needing extensive human-designed inputs, unlike traditional prompt engineering, which relies heavily on manual intervention. On the other hand, fine-tuning involves adjusting a large number of the LLM’s parameters, often requiring significant computational resources and large labeled datasets. However, prompt-tuning only updates the prompt embeddings while keeping the rest of the LLM frozen. Thus, prompt-tuning strikes a balance between the resource-intensive nature of fine-tuning and the simplicity of prompt engineering, allowing for task-specific adaptability in healthcare contexts with minimal computational cost.
\\[0.25cm]
Wang et al. (2023) used ERNIE-Health, a discriminative LLM specialized for healthcare, combined with prompt-tuning to classify healthcare text data. Rather than traditional fine-tuning, which adds extra classification layers, their best-performing method transforms the classification task into a mask prediction task. Specifically, they wrap medical text inputs into natural language templates where category labels are replaced by [UNK] tokens, then leverage ERNIE-Health's multi-token selection (MTS) pre-training task to predict the correct category label from candidate options. The prompt-tuning approach proved particularly effective for the smaller medical dataset (KUAKE-QIC), outperforming traditional fine-tuning by leveraging ERNIE-Health's pre-trained medical domain knowledge without requiring additional parameters or extensive training data. Peng et al. (2024) developed a unified approach to clinical text classification using GatorTronGPT, a GPT-3 further trained on a large amount of clinical text. The key element of this methodology was using soft prompting with a frozen LLM, where the model's original parameters remained unchanged while only trainable vector prompts were optimized during fine-tuning. This approach successfully tackled four text classification tasks: Clinical Abbreviation Disambiguation, NLI, Medication Attribute Filling, and Progress Note Understanding. Gu et al. (2024) developed AGCVT-Prompt, an LLM-based approach for analyzing healthcare-related social media content, specifically focusing on COVID-19 discussions. This method achieved good performance by combining three key components: automatically generating topic templates using T5 to cluster healthcare topics (the clustering is carried out using HDBSCAN using the embeddings obtained using BERT), creating sentiment prompt templates to identify emotional content, and utilizing soft prompt tokens; all integrated into a CoT reasoning approach.

\begin{landscape}
\begin{table}[h]
\captionsetup{labelformat=empty}
\centering
\caption*{\textbf{Table 4.} Fine-tuning-based Reviewed Literature Categorization}
\label{tab:classification_models}
\begin{tabular}{c c c c c c c c c c}
 \toprule
\multirow{2}{*}{\centering\textbf{Reference}} & 
\multirow{2}{*}{\centering\textbf{Type}} & 
\multirow{2}{*}{\centering\textbf{Language}} & 
\multirow{2}{*}{\centering\textbf{Best LLM}} & 
\multirow{2}{*}{\centering\textbf{Application}} & 
\multicolumn{5}{c}{\centering\textbf{Performance Evaluation Metrics}} \\
\cmidrule(lr){6-10}
& & & & & \textbf{Accuracy} & \textbf{F1 Score} & \textbf{Precision} & \textbf{Recall} & \textbf{AUC Score} \\

\hline
\makecell{\textbf{Ohse et al.} \\ (2024)} &	Journal	& English	 & GPT-3.5	& \makecell{Clinical Decision \\ Support}	& - &	82\%	& 84\%	& 85\%	& - \\

\hline
\makecell{\textbf{Raja et al.} \\ (2024)} &	Journal	& English	 & BioBERT	& \makecell{Research/ \\ Literature Analysis}	& - &	Micro: 67\%	& -	& -	& 70\% \\

\hline

\multirow{2}{*}{\centering\makecell{\textbf{Xu et al.} \\ (2024)}} &
\multirow{2}{*}{Conference} & \multirow{2}{*}{English} & \makecell{Mental- \\ Alpaca} & \multirow{2}{*}{\makecell{Clinical Decision \\ Support}} & \makecell{Task 1 (Dreaddit): \\ 81.60\% \\
Task 2 (DepSeverity): \\ 77.50\% \\
Task 4: 72.40\% \\
Task 2 (Red-Sam): \\ 60.40\% \\
Task 1 (SAD): 81.90\%} & - & - & - & - \\
\cmidrule(lr){4-4}
\cmidrule(lr){6-10}
& & & \makecell{Mental- \\ FLAN-T5} & & \makecell{Task 3: 75.56\% \\
Task 5: 86.80\% \\
Task 6: 48.10\% \\
Task 2 (Twt-60Users): \\ 73.60\%} & - & - & - & - \\

\hline
\multirow{6}{*}{\makecell{\textbf{Guevara et al.} \\ (2024)}} & 
\multirow{6}{*}{Journal} & 
\multirow{6}{*}{English} & 
\multirow{2}{*}{\makecell{Flan-T5 \\ XXL + \\ Synth. data}} & 
\multirow{6}{*}{Quality and Equity} & 
\makecell{RT (Any): \\ 70\%} & - & - & - & - \\
\cmidrule(lr){6-10}
& & & & & \makecell{Imm. (Any): \\ 71\%} & - & - & - & - \\
\cmidrule(lr){4-4} 
\cmidrule(lr){6-10}

& & & \makecell{Flan-T5 XL \\ + Synth.\\  data} & & 
\makecell{Imm. \\ (Adverse \\ SDoH): 66\%} & - & - & - & - \\
\cmidrule(lr){4-4} \cmidrule(lr){6-10}

& & & Flan-T5 XXL & & 
\makecell{MIMIC-III \\ (Any): 57\%} & - & - & - & - \\
\cmidrule(lr){4-4} \cmidrule(lr){6-10}

& & & \multirow{2}{*}{Flan-T5 XL} & & 
\makecell{RT \\ (Adverse): \\ 69\%} & - & - & - & - \\
\cmidrule(lr){6-10}

& & & & & 
\makecell{MIMIC-III \\ (Adverse): \\ 53\%} & - & - & - & - \\

\hline
\makecell{\textbf{Xie et al.} \\ (2024)} &	Journal	& English	& Clinical\_BERT	& \makecell{Clinical Decision \\ Support \\
Quality and Equity} &	- &	-	& -	& -	& - \\

\bottomrule
\end{tabular}
\end{table}
\end{landscape}

\begin{landscape}
\begin{table}[h]
\captionsetup{labelformat=empty}
\centering
\caption*{\textbf{Table 4.} Cont.}
\label{tab:classification_models}
\begin{tabular}{c c c c c c c c c c}
 \toprule
\multirow{2}{*}{\centering\textbf{Reference}} & 
\multirow{2}{*}{\centering\textbf{Type}} & 
\multirow{2}{*}{\centering\textbf{Language}} & 
\multirow{2}{*}{\centering\textbf{Best LLM}} & 
\multirow{2}{*}{\centering\textbf{Application}} & 
\multicolumn{5}{c}{\centering\textbf{Performance Evaluation Metrics}} \\
\cmidrule(lr){6-10}
& & & & & \textbf{Accuracy} & \textbf{F1 Score} & \textbf{Precision} & \textbf{Recall} & \textbf{AUC Score} \\

\hline
\multirow{4}{*}{\makecell{\textbf{Li et al.} \\  (2024b)}} & \multirow{4}{*}{Conference} & \multirow{4}{*}{English} & \multirow{4}{*}{LlamaCare} & \multirow{2}{*}{\makecell{Patient Safety and \\ Risk Assessment}} & - & - &	-	& -	& \makecell{MOR: \\ 77.62\%} \\
\cmidrule(lr){6-10}
 & &  & &  & - & - &	-	& -	& \makecell{LOS: \\ 68.76\%} \\
 \cmidrule(lr){5-10}
& & &  & \multirow{2}{*}{\makecell{Clinical Decision \\ Support}} & - & - &	-	& -	& \makecell{DIAG: \\ 79.16\%} \\
\cmidrule(lr){6-10}
 & &  &  & & - & - &	-	& -	& \makecell{PROC: \\ 90.76\%} \\

\hline
\makecell{\textbf{Bumgardner} \\ \textbf{et al.} (2024)}	& Conference &	English	& \makecell{Path- \\ LLaMA \\ 13B}	& \makecell{Clinical Decision \\ Support} &	74.80\%	& 77.50\%&	77.70\%	& 77.90\%	& 81.60\% \\

\hline
\makecell{\textbf{Wang et al.} \\ (2024)} & Journal & English & \makecell{DRG- \\ LLaMA- \\ 7B} & \makecell{Clinical Decision \\ Support} & \makecell{ACC@1: 52.00\% \\
ACC@5:84.80\% \\
ACC@10: 91.20\%} & Macro: 32.70\% & - &	- &	\makecell{Macro: \\ 98.60\% \\
Micro: \\ 99.40\%} \\

\hline
\makecell{\textbf{Chen et al.} \\ (2024b)} & Journal & English &	BioBERT & \makecell{Research/Literature \\ Analysis} & - & \makecell{Discussion: \\ 80\% \\
Unstructured: \\ 82.10\% \\
Structured: \\ 90.20\%} & -	& -	& - \\

\hline
\multirow{2}{*}{\makecell{\textbf{Silverman et} \\ \textbf{al.} (2024)}} & \multirow{2}{*}{Journal} & \multirow{2}{*}{English} & \makecell{H-UCSF- \\ BERT} & \multirow{2}{*}{\makecell{Patient Safety and \\ Risk Assessment}} & Task 1: 88\% &	Task 1: 62\%	& Task 1: 38\%	& Task 1: 27\% &	- \\
\cmidrule(lr){4-4}
\cmidrule(lr){6-10}
 &  &  & \makecell{H-UCSF- \\ BERT + \\ only \\ nearby \\ SAEs} & & \makecell{Task 2: 92\% \\
Task 3: 91\%}	& \makecell{Task 2: 68\% \\
Task 3: 61\%}	& \makecell{Task 2: 36\% \\
Task 3: 44\%}	& \makecell{ask 2: 45\% \\
Task 3: 19\%} &	-\\

\hline
\makecell{\textbf{Bețianu et} \\ \textbf{al.} (2024)} &	Conference	& English	& \makecell{DALLMi- \\ BERT} & \makecell{Research/Literature \\ Analysis} &	-	& -	& -	& \makecell{mAP: \\ 58.20\%}	& - \\

\hline

\makecell{\textbf{Farruque} \\ \textbf{et al.} \\ (2024)} & Journal	& English &	\makecell{Mental-BERT} &	\makecell{Clinical Decision \\ Support}	& -	& \makecell{DSD \\ (Macro): \\  45\% \\
DSD \\ (Weighted): \\ 56\%} &	\makecell{DSD \\ (Macro): \\ 51\% \\
DSD \\ (Weighted): \\ 68\%} &	\makecell{DSD \\ (Macro): \\ 49\%\\ 
DSD \\ (Weighted):\\  51\%}	- \\
\cmidrule(lr){6-10}
& & & & & -	& DPD: 89\%	& DPD: 97\%	& DPD: 83\%	& - \\

\bottomrule
\end{tabular}
\end{table}
\end{landscape}

\begin{landscape}
\begin{table}[h]
\captionsetup{labelformat=empty}
\centering
\caption*{\textbf{Table 4.} Cont.}
\label{tab:classification_models}
\begin{tabular}{c c c c c c c c c c}
 \toprule
\multirow{2}{*}{\centering\textbf{Reference}} & 
\multirow{2}{*}{\centering\textbf{Type}} & 
\multirow{2}{*}{\centering\textbf{Language}} & 
\multirow{2}{*}{\centering\textbf{Best LLM}} & 
\multirow{2}{*}{\centering\textbf{Application}} & 
\multicolumn{5}{c}{\centering\textbf{Performance Evaluation Metrics}} \\
\cmidrule(lr){6-10}
& & & & & \textbf{Accuracy} & \textbf{F1 Score} & \textbf{Precision} & \textbf{Recall} & \textbf{AUC Score} \\

\hline
\makecell{\textbf{Uslu et al.} \\ (2024)} &	Journal &	English &	\makecell{CXR- \\ BERT- \\ GENERAL}	& \makecell{Clinical Decision \\ Support} &	-	& \makecell{Macro: 71.61\% \\
Micro: 80.49\% \\
Weighted: 80.14\%} &	\makecell{Macro: \\ 69.91\% \\
Micro: \\ 80.73\% \\
Weighted: \\ 80.73\%} &	\makecell{Macro: \\ 75.44\% \\
Micro: \\ 80.27\% \\
Weighted: \\ 80.22\%} & -	 \\

\hline
\multirow{2}{*}{\makecell{\textbf{Luo et al.} \\ (2024)}} & \multirow{2}{*}{Journal} & 	\multirow{2}{*}{\makecell{English \& Chinese}} & \multirow{2}{*}{\makecell{Taiyi \\ (Fine-tuned \\ Qwen-7B- \\ base)}}	& \makecell{Research/Literature \\ Analysis} &	-	& \makecell{BC7LitCoivd \\ (Micro): 84\% \\
HoC (Micro): 80\%}	& - &	-	& - \\
\cmidrule(lr){5-10}
& & & & \makecell{Patient Query \\ Analysis} & - & \makecell{KUAKE\_QIC \\ (Micro): 77.4\%} &	- &	- &	- \\

\hline
\multirow{2}{*}{\makecell{\textbf{Lehman et} \\ \textbf{al.} (2023)}} & \multirow{2}{*}{Conference} & 	\multirow{2}{*}{\makecell{English}} & \multirow{2}{*}{\makecell{BioClin \\ RoBERTa}}	& \makecell{Information \\ Extraction} &	-	& \makecell{Macro: 70.70\% \\ Micro: 80.50\%}	& - &	-	& - \\
\cmidrule(lr){5-10}
& & & & \makecell{Other Annotations} & 90\% & - &	- &	- &	- \\

\hline
\makecell{\textbf{Cui et al.} \\ (2023)}	& Conference&	English	&BERT&	\makecell{Clinical Decision \\ Support} &	77.56\%	& 92.45\%	&-	&-	&- \\

\hline
\makecell{\textbf{Chen et al.} \\ (2023)}	&Journal	&English	& \makecell{BioLink \\ BERT-Large}	& \makecell{Research/Literature \\ Analysis}	& - &	Micro: 84.90\%	& - &	- &	- \\

\hline
\makecell{\textbf{Bansal et} \\ \textbf{al.} (2023)}	&Conference	&English	& \makecell{DeBERTa \\ Large} &	\makecell{Public Health and \\ Opinion Analysis}	&-	& Macro: 67\%	& - &	- &	- \\

\hline
\makecell{\textbf{Gretz et} \\ \textbf{al.} \\ (2023)} &	Conference &	English	 & \makecell{Flan-T5- \\ XXL}	 & \makecell{Research/Literature \\ Analysis} &	- &	Macro: 60.69\%	 &- &	- &	- \\

\hline
\makecell{\textbf{Savage et} \\ \textbf{al.} (2023)} & 	Journal & 	English & 	\makecell{BioMed- \\ RoBERTa }& 	\makecell{Clinical Decision \\ Support} & 	- & 	- & 	67\% & 	-	 & \makecell{PR: 82\% \\
ROC: 89\% } \\

\hline
\makecell{\textbf{Van} \\ \textbf{Ostaeyen} \\ \textbf{et al.} \\ (2023)} &	Journal &	Dutch &	RobBERT &	Quality and Equity &	- &	\makecell{Quality: 76\% \\
CanMEDS: 72\%} &	- &	- &	- \\

\hline
\makecell{\textbf{Ge et al.} \\ (2023)} &	Conference &	Chinese &	\makecell{Baichuan2- \\ 13B} &	\makecell{Patient Query \\ Analysis} &	92.10\% &	- &	- &	- &	-  \\

\hline
\makecell{\textbf{Ren et al.} \\ (2023)} &	Conference &	English	& BERTweet	& \makecell{Patient Query \\ Analysis} &	78\%	& Macro: 75\% & 	-	& -& 	- \\
\hline
\makecell{\textbf{Tan et al.} \\ (2023)} &	Journal	&English	&GatorTron	& \makecell{Clinical Decision \\ Support}	&86.09\%	&-	&-	&-	&- \\
\hline
\makecell{\textbf{Shiju et al.} \\ (2022)} &	Conference &	English	& \makecell{Bio\_Clinical \\ BERT} &	\makecell{Public Health and \\ Opinion Analysis}&	87\%	&87\%&	-	&-&	- \\

\bottomrule
\end{tabular}
\end{table}
\end{landscape}

\begin{landscape}
\begin{table}[h]
\captionsetup{labelformat=empty}
\centering
\caption*{\textbf{Table 4.} Cont.}
\label{tab:classification_models}
\begin{tabular}{c c c c c c c c c c}
 \toprule
\multirow{2}{*}{\centering\textbf{Reference}} & 
\multirow{2}{*}{\centering\textbf{Type}} & 
\multirow{2}{*}{\centering\textbf{Language}} & 
\multirow{2}{*}{\centering\textbf{Best LLM}} & 
\multirow{2}{*}{\centering\textbf{Application}} & 
\multicolumn{5}{c}{\centering\textbf{Performance Evaluation Metrics}} \\
\cmidrule(lr){6-10}
& & & & & \textbf{Accuracy} & \textbf{F1 Score} & \textbf{Precision} & \textbf{Recall} & \textbf{AUC Score} \\

\hline
\multirow{2}{*}{\makecell{\textbf{Kementchedjhieva} \\ \textbf{and} \\ \textbf{Chalkidis} (2023)}} & \multirow{2}{*}{Conference} & 	\multirow{2}{*}{\makecell{English}} & \multirow{2}{*}{T5Enc}	& \makecell{Clinical Decision \\ Support} &	-	& \makecell{Micro: 60.50\% \\
Macro: 31.10\%}	& - &	-	& - \\
\cmidrule(lr){5-10}
& & & & \makecell{Research/Literature \\ Analysis} & - & \makecell{Micro: 75.10\% \\
Macro: 66\%} &	- &	- &	- \\

\hline
\makecell{\textbf{Qi et al.} \\ (2023)} &	Conference& 	English	& SaFER-BERT	& \makecell{Research/Literature \\ Analysis} & 	\makecell{Task 1: \\ 80\% \\ 
Task 2: \\ 94\%} & 	-& 	-& 	-& 	- \\
\hline
\makecell{\textbf{Ciobotaru} \\ \textbf{and Dinu} \\ (2023)} &	Conference& 	Romanian& 	Romanian BERT	& \makecell{Public Health and \\ Opinion Analysis}	& 84\%	& \makecell{Macro: \\ 84\%} & \makecell{Macro: \\ 84\%}	& \makecell{Macro: \\ 85\%}	& - \\
\hline
\makecell{\textbf{Kersting} \\ \textbf{et al.} \\ (2023)}  &	Journal& 	German& 	\makecell{XLM- \\ RoBERTa- \\ large}	& \makecell{Public Health and \\ Opinion Analysis}	& \makecell{A: 88\% \\
B: 86\%\\
C: 88\%\\
D: 89\%}	& \makecell{A: 60\%\\
B: 45\%\\
C: 57\%\\
D: 68\%}	& \makecell{A: 57\%\\
B: 44\%\\
C: 55\%\\
D: 71\%}	& \makecell{A: 72\%\\
B: 50\%\\
C: 60\%\\
D: 67\%}	& - \\

\hline
\makecell{\textbf{Yogarajan} \\ \textbf{et al.} \\ (2022a)}	 & Conference	 & English	 & TransformerXL	 & \makecell{Clinical Decision \\ Support} & 	-	 & \makecell{Micro: 72.30\% \\
Macro: 67.70\%}	 & -	 & -	 & - \\
\hline
\makecell{\textbf{Chen et al.} \\  (2022)} & 	Journal	 & English	 & LitMC-BERT	 & \makecell{Research/Literature \\ Analysis}	 & \makecell{LitCovid: \\ 80.22\%  \\
HoC: \\ 68.54\%} & 	\makecell{LitCovid: \\ 93.14\% \\
HoC: \\ 90.36\%} & 	\makecell{LitCovid: \\ 92.12\% \\
HoC: \\ 90.38\%}	 & \makecell{LitCovid: \\ 94.18\% \\
HoC: \\ 90.35\%}	 & - \\
\hline
\makecell{\textbf{Yogarajan} \\ \textbf{et al.} \\ (2022b)}	 & Conference	 & English	 & TransformerXL	 & \makecell{Patient Safety and \\ Risk Assessment} & 	- & 	- & \makecell{MIMIC-III \\ (Macro): \\ 51\%\\ 
MIMIC-III \\ (Micro): 65\%\\ 
eICU \\ (Macro): \\ 40\%\\ 
eICU \\ (Micro): 63\%}	 & - & 	- \\

\hline
\makecell{\textbf{Schneider} \\ \textbf{et al.} \\ (2021)} &	Conference	 & Portuguese	 & GPT2-Bio-Pt	 & \makecell{Patient Safety and \\ Risk Assessment} & 	-	 & Weighted: 90.09\% & 	- & 	-	 & - \\

\bottomrule
\end{tabular}
\end{table}
\end{landscape}

\begin{landscape}
\begin{table}[h]
\captionsetup{labelformat=empty}
\centering
\caption*{\textbf{Table 4.} Cont.}
\label{tab:classification_models}
\begin{tabular}{c c c c c c c c c c}
 \toprule
\multirow{2}{*}{\centering\textbf{Reference}} & 
\multirow{2}{*}{\centering\textbf{Type}} & 
\multirow{2}{*}{\centering\textbf{Language}} & 
\multirow{2}{*}{\centering\textbf{Best LLM}} & 
\multirow{2}{*}{\centering\textbf{Application}} & 
\multicolumn{5}{c}{\centering\textbf{Performance Evaluation Metrics}} \\
\cmidrule(lr){6-10}
& & & & & \textbf{Accuracy} & \textbf{F1 Score} & \textbf{Precision} & \textbf{Recall} & \textbf{AUC Score} \\

\hline
\multirow{2}{*}{\makecell{\textbf{Yogarajan} \\ \textbf{et al.} \\ (2021)}} & \multirow{2}{*}{Conference} & 	\multirow{2}{*}{\makecell{English}} & PubMedBERT	& \makecell{Clinical Decision \\ Support} &	-	& \makecell{MIMIC-III (Micro): \\ 65\% \\
MIMIC-III (Macro): \\ 41\%}	& - &	-	& - \\
\cmidrule(lr){4-4}
\cmidrule(lr){6-10}
& & & \makecell{BioMed- \\ RoBERTa-base} &  & - & \makecell{eICU (Micro): 60\% \\
eICU (Macro): 32\%} &	- &	- &	- \\

\hline
\makecell{\textbf{Pan et al.} \\ (2020)} &	Conference	 & Chinese	 & FAMLC-BERT	 & \makecell{Clinical Decision \\ Support} & 	-	& 100\%& 	65.39\%	& 79.07\%	& 99.90\% \\

\bottomrule
\end{tabular}
\end{table}
\end{landscape}

\begin{landscape}
\begin{table}[h]
\captionsetup{labelformat=empty}
\centering
\caption*{\textbf{Table 5.} Pre-training-based Reviewed Literature Categorization}
\label{tab:classification_models}
\begin{tabular}{c c c c c c c c c c}
 \toprule
\multirow{2}{*}{\centering\textbf{Reference}} & 
\multirow{2}{*}{\centering\textbf{Type}} & 
\multirow{2}{*}{\centering\textbf{Language}} & 
\multirow{2}{*}{\centering\textbf{Best LLM}} & 
\multirow{2}{*}{\centering\textbf{Application}} & 
\multicolumn{5}{c}{\centering\textbf{Performance Evaluation Metrics}} \\
\cmidrule(lr){6-10}
& & & & & \textbf{Accuracy} & \textbf{F1 Score} & \textbf{Precision} & \textbf{Recall} & \textbf{AUC Score} \\

\hline
\makecell{\textbf{McMaster et} \\ \textbf{al.} \\ (2023)} &	Conference	 & English	 & MeDeBERTa	 & \makecell{Patient Safety and \\ Risk Assessment} & 	-	& 61.10\%& 	-	& -	& \makecell{PR: 89.70\% \\
ROC: 95.59\%} \\

\hline
\multirow{2}{*}{\makecell{\textbf{Li et al} \\ (2023b)}} & \multirow{2}{*}{Journal} & 	\multirow{2}{*}{\makecell{English}} & \makecell{Clinical- \\ Longformer}	& \multirow{2}{*}{\makecell{Clinical Decision \\ Support}} &	OpenI: 97.70\%	& \makecell{MIMIC- \\ AKI:48.40\%}	& - &	-	& \makecell{MIMIC- \\ AKI: \\ 76.20\%} \\
\cmidrule(lr){5-10}
& & & &  Other Annotations & MedNLI: 84.20\%	& - &	- &	- &	- \\

\hline
\makecell{\textbf{Yang et al.} \\ (2022)} &	Journal	 & English & \makecell{GatorTron-\\large} & Other Annotations & 	90.20\%	& - & 	-	& -	& - \\

\hline
\makecell{\textbf{Bressem et al.} \\ (2020)} &	Journal	 & German & RAD-BERT & \makecell{Clinical Decision \\ Support} & 	89\% & 90\% & 	-	& -	& \makecell{PR:93\% \\
ROC: 98\%} \\

\hline
\makecell{\textbf{Blinov et al.} \\ (2020)} & Conference	 & Russian & RuPool-BERT & \makecell{Clinical Decision \\ Support} & - & \makecell{Macro: \\ 29.83\% \\
Weighted: \\ 47.13\%} & 	-	& -	& - \\

\bottomrule
\end{tabular}
\end{table}
\end{landscape}

\begin{landscape}
\begin{table}[h]
\captionsetup{labelformat=empty}
\centering
\caption*{\textbf{Table 6.} Prompt-tuning-based Reviewed Literature Categorization}
\label{tab:classification_models}
\begin{tabular}{c c c c c c c c c c}
 \toprule
\multirow{2}{*}{\centering\textbf{Reference}} & 
\multirow{2}{*}{\centering\textbf{Type}} & 
\multirow{2}{*}{\centering\textbf{Language}} & 
\multirow{2}{*}{\centering\textbf{Best LLM}} & 
\multirow{2}{*}{\centering\textbf{Application}} & 
\multicolumn{5}{c}{\centering\textbf{Performance Evaluation Metrics}} \\
\cmidrule(lr){6-10}
& & & & & \textbf{Accuracy} & \textbf{F1 Score} & \textbf{Precision} & \textbf{Recall} & \textbf{AUC Score} \\

\hline
\multirow{2}{*}{\makecell{\textbf{Wang et} \\ \textbf{al.} \\ (2023)}} & \multirow{2}{*}{Journal} & 	\multirow{2}{*}{\makecell{Chinese}} & \multirow{2}{*}{ERNIE-Health}	& \makecell{Patient Query \\ Analysis} &	86.60\%	& -	& - &	-	& - \\
\cmidrule(lr){5-10}
& & & &  \makecell{Research/ \\ Literature Analysis} & 86.10\%	& - &	- &	- &	- \\

\hline
\multirow{4}{*}{\makecell{\textbf{Peng et al.} \\ (2023)}} & \multirow{4}{*}{Journal} & 	\multirow{4}{*}{\makecell{English}} & \makecell{GatorTronGPT\\-5B}	& \multirow{4}{*}{Information Extraction} &	\makecell{Clinical Abbreviation \\ Disambiguation}	& -	& 98.42\% &	98.32\%	& 98.54\% \\
\cmidrule(lr){4-4}
\cmidrule(lr){6-10}
& & &  \makecell{GatorTronGPT\\-20B}	&  &	\makecell{Natural Language \\ Inference}	& 89.46\%	& - &	-	& - \\
\cmidrule(lr){4-4}
\cmidrule(lr){6-10}
& & &   GatorTronGPT &  &	\makecell{Medication Attribute \\ Filling}	& -	& \makecell{Event: \\ 93.79\% \\
Context: \\ 91.26\%} &	-	& - \\
\cmidrule(lr){4-4}
\cmidrule(lr){6-10}
& & &   \makecell{GatorTronGPT\\-20B}	&  &	\makecell{Progress Note \\ Understanding}	& -	& 79.54\% &	-	& - \\

\hline
\makecell{\textbf{Gu et al.} \\ (2024)} &	Journal	 & Chinese	 & \makecell{T5-AGCVT-\\Prompt} & \makecell{Public Health and \\ Opinion Analysis} & 	79\% & 75\% & 	-	& -	& - \\

\bottomrule
\end{tabular}
\end{table}
\end{landscape}

\begin{landscape}
\begin{table}[h]
\captionsetup{labelformat=empty}
\centering
\caption*{\textbf{Table 7.} Text Classification and LLM Type Categorization: The Case of Fine-tuning}
\label{tab:classification_models}
\begin{tabular}{c c c c c c c c c}
 \toprule
\multirow{2}{*}{\centering\textbf{Reference}} & 
\multicolumn{3}{c}{\makecell{\textbf{Text Classification Type}}} & 
\multicolumn{5}{c}{\makecell{\textbf{Large Language Model Type}}}
\\
\cmidrule(lr){2-9}

& \textbf{Multi-class}	& \textbf{Multi-label}	& \textbf{Binary} & \textbf{	BERT (or variant)}	& \textbf{Closed-source} &	\textbf{Open-source}	& \makecell{\textbf{Pre-trained} \\ \textbf{Transformer}}	& \textbf{BART} \\

\hline
\textbf{Ohse et al.} (2024) &  &  &  \checkmark  &  &  \checkmark  &  &  &   \\
\hline
\textbf{Raja et al.} (2024) &  &  \checkmark  &   &  \checkmark  &  &  &  &   \\
\hline
\textbf{Xu et al.} (2024) & \checkmark &  &  \checkmark  &  &  &  \checkmark  &  \checkmark  &   \\
\hline
\makecell{\textbf{Guevara et al.} (2024)} &  &  \checkmark   &  &  &  &  &  \checkmark  &   \\
\hline
\textbf{Li et al.} (2024b) & \checkmark &  \checkmark  &   \checkmark  &  &  &  \checkmark  &  &   \\
\hline
\textbf{Luo et al.} (2024) &  &  \checkmark  &   &  &  &  \checkmark  &  &   \\
\hline
\textbf{Xie et al.} (2024) &  &  &  \checkmark  &  \checkmark  &  &  &  &   \\
\hline
\textbf{Uslu et al.} (2024) &  &  \checkmark  &   &  \checkmark  &  &  &  &   \\
\hline
\makecell{\textbf{Bumgardner et al.} (2024)} &  &  \checkmark  &   &  &  &  \checkmark  &  &   \\
\hline
\textbf{Wang et al.} (2024) & \checkmark &  &  &  &  &  \checkmark  &  &   \\
\hline
\makecell{\textbf{Chen et al.} (2024b)} & \checkmark &  &  &  \checkmark  &  &  &  &   \\
\hline
\makecell{\textbf{Silverman et al.} (2024)} &  &  &  \checkmark  &  \checkmark  &  &  &  &   \\
\hline
\makecell{\textbf{Bețianu et al.} (2024)} &  &  \checkmark   &  &  \checkmark  &  &  &  &   \\
\hline
\makecell{\textbf{Farruque et al.} (2024)} & \checkmark &  \checkmark   &  \checkmark  &  \checkmark  &  &  &  &   \\
\hline
\textbf{Qi et al.} (2023) &  &  &  \checkmark  &  \checkmark  &  &  &  &   \\
\hline
\makecell{\textbf{Ciobotaru and Dinu} \\ (2023)} & \checkmark &  &  &  \checkmark  &  &  &  &   \\
\hline
\textbf{Gretz et al.} (2023) & \checkmark &  &  &  &  &  &  \checkmark  &   \\
\hline
\makecell{\textbf{Kersting et al.} (2023)} & \checkmark &  &  &  \checkmark  &  &  &  &   \\

\hline
\makecell{\textbf{Lehman et al.} (2023)} &  \checkmark  &   \checkmark  &  &  \checkmark  &  &  &  & \\
\hline
\makecell{\textbf{Bansal et al.} (2023)} &  &   \checkmark  &  &  \checkmark &   &  &  & \\
\hline
\textbf{Cui et al.} (2023) &  &   &  \checkmark  &  \checkmark  &  &  &  & \\
\hline
\makecell{\textbf{Van Ostaeyen et al.} (2023)} &  \checkmark  &   \checkmark  &  &  \checkmark  &  &  &  & \\
\hline
\textbf{Ge et al.} (2023) &  \checkmark  &  &  &   &  &  \checkmark  &  & \\
\hline
\textbf{Ren et al.} (2023) &  \checkmark  &   &  &  \checkmark &   &  &  & \\
\hline
\makecell{\textbf{Savage et al.} (2023)} &  &   &  \checkmark  &  \checkmark  &  &  &  & \\
\hline
\textbf{Tan et al.} (2023) &  \checkmark  &   &  &  &  \checkmark  &  &  & \\
\hline
\makecell{\textbf{Kementchedjhieva and} \\ \textbf{Chalkidis} (2023)} &  &   \checkmark  &  &  &  &  &  \checkmark  & \\
\hline
\textbf{Chen et al.} (2023) &  &   \checkmark  &  &  \checkmark  &  &  &  & \\
\hline
\makecell{\textbf{Yogarajan et al.} (2022a)} &  &   \checkmark  &  &  &  &  &  \checkmark  & \\
\hline
\makecell{\textbf{Yogarajan et al.} (2022b)} &  &   \checkmark  &  &  &  &  &  \checkmark  & \\

\bottomrule
\end{tabular}
\end{table}
\end{landscape}

\begin{landscape}
\begin{table}[h]
\captionsetup{labelformat=empty}
\centering
\caption*{\textbf{Table 7.} Cont.}
\label{tab:classification_models}
\begin{tabular}{c c c c c c c c c}
 \toprule
\multirow{2}{*}{\centering\textbf{Reference}} & 
\multicolumn{3}{c}{\makecell{\textbf{Text Classification Type}}} & 
\multicolumn{5}{c}{\makecell{\textbf{Large Language Model Type}}}
\\
\cmidrule(lr){2-9}

& \textbf{Multi-class}	& \textbf{Multi-label}	& \textbf{Binary} & \textbf{	BERT (or variant)}	& \textbf{Closed-source} &	\textbf{Open-source}	& \makecell{\textbf{Pre-trained} \\ \textbf{Transformer}}	& \textbf{BART} \\

\hline
\textbf{Shiju et al.} (2022) &  &   &  \checkmark  &  \checkmark  &  &  &  & \\
\hline
\textbf{Chen et al.} (2022) &  &   \checkmark  &  &  \checkmark  &  &  &  & \\
\hline
\makecell{\textbf{Schneider et al.} (2021)} &  &   &  \checkmark  &  &  \checkmark  &  &  & \\
\hline
\makecell{\textbf{Yogarajan et al.} (2021)} &  &   \checkmark  &  &   \checkmark  &  &  &  & \\
\hline
\textbf{Pan et al.} (2020) &  &   \checkmark  &  &  \checkmark  &  &  &  & \\

\bottomrule
\end{tabular}
\end{table}

\begin{table}[h]
\captionsetup{labelformat=empty}
\centering
\caption*{\textbf{Table 8.} Text Classification and LLM Type Categorization: The Case of Pre-training}
\label{tab:classification_models}
\begin{tabular}{c c c c c c c c c}
 \toprule
\multirow{2}{*}{\centering\textbf{Reference}} & 
\multicolumn{3}{c}{\makecell{\textbf{Text Classification Type}}} & 
\multicolumn{5}{c}{\makecell{\textbf{Large Language Model Type}}}
\\
\cmidrule(lr){2-9}

& \textbf{Multi-class}	& \textbf{Multi-label}	& \textbf{Binary} & \textbf{	BERT (or variant)}	& \textbf{Closed-source} &	\textbf{Open-source}	& \makecell{\textbf{Pre-trained} \\ \textbf{Transformer}}	& \textbf{BART} \\

\hline
\textbf{McMaster et al.} (2023) &  &  &  \checkmark  &  \checkmark  &  &  &  &  \\
\hline
\textbf{Li et al.} (2023b) &  \checkmark  &  \checkmark  &   \checkmark  &  &  &  &  \checkmark  &  \\
\hline
\textbf{Yang et al.} (2022) &  \checkmark  &  &  &  &  &  &  \checkmark  &  \\
\hline
\textbf{Bressem et al.} (2020) &  &  \checkmark   &  &  \checkmark  &  &  &  &  \\
\hline
\textbf{Blinov et al.} (2020) &  \checkmark  &  &  &  \checkmark  &  &  &  &  \\

\bottomrule
\end{tabular}
\end{table}

\begin{table}[h]
\captionsetup{labelformat=empty}
\centering
\caption*{\textbf{Table 9.} Text Classification and LLM Type Categorization: The Case of Prompt-tuning}
\label{tab:classification_models}
\begin{tabular}{c c c c c c c c c}
 \toprule
\multirow{2}{*}{\centering\textbf{Reference}} & 
\multicolumn{3}{c}{\makecell{\textbf{Text Classification Type}}} & 
\multicolumn{5}{c}{\makecell{\textbf{Large Language Model Type}}}
\\
\cmidrule(lr){2-9}

& \textbf{Multi-class}	& \textbf{Multi-label}	& \textbf{Binary} & \textbf{	BERT (or variant)}	& \textbf{Closed-source} &	\textbf{Open-source}	& \makecell{\textbf{Pre-trained} \\ \textbf{Transformer}}	& \textbf{BART} \\

\hline
\textbf{Peng et al.} (2024) &  \checkmark  &  &  &  &  \checkmark  & &  & \\
\hline
\textbf{Gu et al.} (2024) &  \checkmark  &  &  &  &  & & \checkmark  & \\
\hline
\textbf{Wang et al.} (2023) &  \checkmark  &  &  &  &  & & \checkmark  & \\

\bottomrule
\end{tabular}
\end{table}

\end{landscape}

\subsubsection{Other LLM-based Approaches}
While reviewing the research papers considered, in addition to prompt engineering, (further/from scratch) pre-training, fine-tuning, and prompt-tuning, other approaches were found to contribute to improving the healthcare text classification task. These LLM-based approaches include ensemble learning, data augmentation, and Retrieval-Augmented Generation (RAG).
\\[0.25cm]
Resorting to an ensemble learning-based approach for an LLM-based healthcare text classification consists of combining the strengths of multiple models to enhance performance in handling complex and nuanced healthcare texts. These methods involve leveraging various LLMs, such as BERT, GPT-4, or RoBERTa, to improve the robustness and accuracy of classifying unstructured healthcare data like patient comments or electronic health records (EHRs). In some cases, the base models also include other conventional machine learning methods. The ensemble approach typically involves techniques like bagging, boosting, or stacking, where outputs of individual models are combined, either through majority voting, weighted averaging, or meta-learning, to produce final predictions. This approach helps mitigate the biases and limitations of single models, resulting in higher classification performance. 
\\[0.25cm]
Five research studies resorted to ensemble learning, leveraging the strengths of multiple traditional machine learning models and LLMs. Li et al. (2023a) used an ensemble learning method that combined three LLMs (BERT, RoBERTa, and ClinicalBERT) with a majority voting for final classifications of Alzheimer's Disease (AD) signs and symptoms in EHRs. These tasks included a binary classification (i.e., whether the text contains AD signs and symptoms) and a multi-class classification (MC) based on nine pre-defined AD signs and symptoms categories. This ensemble was trained (i.e., each LLM was fine-tuned) on a combination of three datasets: human-annotated “gold” data, LLM-annotated “silver” data from public EHRs (using Llama 65B for annotation), and synthetic “bronze” data generated by GPT-4. The methodology demonstrated that combining multiple pre-trained models in an ensemble while leveraging both LLM-annotated real clinical text and LLM-generated synthetic data for training could effectively enhance the detection of medical conditions from clinical notes. Wu et al. (2023) achieved SOTA performance in healthcare text classification through a proposed ensemble learning approach combining three LLMs. The methodology focuses on diabetes-related patient queries, using ChatGLM2-6B, Qwen-7B-Chat, and MacBERT in an ensemble architecture, with each model fine-tuned using different techniques (e.g., LoRA, QLoRA, Fast Gradient Method (FGM)). Additionally, ChatGPT and Claude were leveraged for data augmentation, mainly focusing on difficult-to-classify text documents. The system processed the patient queries through all LLMs and employed a majority voting mechanism to determine the final classification among six medical categories (i.e., Diagnosis, Treatment, Common Knowledge, Healthy Lifestyle, Epidemiology, and Other). Jiang et al. (2023) proposed ALEX-L, an unconventional form of ensemble learning that analyzes healthcare-related social media text data that combines BERT models with LLM-based validation. This resulted in three tasks: COVID-19 Diagnosis, Sentiment Analysis, and Social Anxiety Analysis. The methodology uses BERT variants to make initial classifications and then leverages GPT-3.5 as a verification layer. The GPT-3.5 component works by taking BERT's predictions and the original healthcare text, combining them with task-specific instructions and examples into a prompt, and then asking the GPT-3.5 to verify if there's evidence supporting the predicted label. If the GPT-3.5 determines there's insufficient evidence (returns “False”), the prediction is corrected either through manual review or automatic conversion to another class. Chaichulee et al.’s (2022) research achieved its best performance in analyzing bilingual (i.e., Thai and English) ADR reports through an ensemble approach that combined six distinct models: NB-SVM (Naive Bayes - Support Vector Machine), ULMFiT (Universal Language Model Fine-tuning), and four BERT variants (mBERT, XLM-RoBERTa, WangchanBERTa, and AllergyRoBERTa). This comprehensive ensemble method was applied to classify free-text clinical descriptions of drug allergies into the 36 most frequently coded symptom terms from EHR in a Thai hospital setting. The methodology used majority voting to aggregate predictions from all models, with each model bringing unique strengths. Yang et al. (2024) used GPT-4 with a ZS-CoT prompting technique combined with ensemble querying to identify potential drug targets from biomedical literature. The authors tested this on COVID-19 and Nipah virus literature, where the system analyzed papers' titles, abstracts, and keywords to determine if they contained relevant drug target information. The methodology involved crafting specific prompts that included a system message, key definitions, and carefully framed questions to guide GPT-4's analysis. The ensemble approach involved sending the same query three times and using majority voting to make the final decision.  
\\[0.25cm]
Using LLMs for data augmentation in healthcare text classification can significantly improve the model’s performance, particularly when dealing with limited labeled data. This approach involves leveraging LLMs to generate synthetic data, which can then be used to increase the training dataset’s size. These synthetic examples mimic real-world healthcare texts, adding more diversity and robustness to the training process. By introducing variation in the data, LLM-generated augmentation can potentially help with overfitting reduction, which is a common issue when working with smaller datasets. Different augmentation strategies can be employed using LLMs, including document paraphrasing, alternative descriptions of medical conditions, or even the generation of new, plausible clinical notes or patients’ comments. As a result, once the dataset is augmented, a machine learning model or LLM can be trained following a supervised learning paradigm. 
\\[0.25cm]
Only one paper from the eligible ones was identified as using LLMs for data augmentation to classify text. Yuan et al. (2023) proposed LLM-PTM to improve patient-trial matching. The methodology aims to match patients’ EHR with clinical trials while preserving patient privacy. LLM-PTM uses a data augmentation pipeline powered by LLMs. It employed ChatGPT to generate augmented versions of clinical trial eligibility criteria using desensitized patient data to maintain privacy. These versions are semantically equivalent but linguistically diverse while maintaining the exact medical meaning. Subsequently, the augmented data, alongside patients’ EHRs, are embedded into a shared latent space using pre-trained BERT models. The embedding of patient records is enhanced with a memory network to capture the sequence of medical visits, diagnoses, and treatments, while a highway network is employed for encoding the eligibility criteria. Once both the patient data and clinical trial criteria are embedded, the model performs patient-trial matching. It uses a contrastive loss function to maximize the similarity between patient embeddings and inclusion criteria while minimizing the similarity with exclusion criteria. This enables the model to accurately determine whether a patient matches the eligibility requirements of a clinical trial. The proposed approach led to better performance while maintaining patient privacy by only augmenting the publicly available trial criteria rather than sensitive patient data.
\\[0.25cm]
RAG is a recently introduced technique by Lewis et al. (2020), leveraging the combination of LLMs and information retrieval techniques. The aim is to enhance the performance of NLP tasks, including text classification. It is particularly useful when the internal knowledge of the LLM might be revealed as insufficient to handle the task in a specific domain. The idea behind this framework consists of retrieval and generation. It consists of augmenting the input of the LLM using external data sources to be used as additional context during the generation phase. It is designed to help mitigate LLM hallucinations and output inconsistencies, especially in precise areas such as healthcare. For text classification, RAG can potentially offer multiple advantages, such as (1) enhancing the context to ensure a better understanding of the input text and thus improve the classification accuracy, (2) adapting to a specific domain by providing domain-specific information to retrieve from, this is particularly useful when implementing general-purpose LLMs, and (3) ameliorating explainability since the retrieved information can serve for a better classification process. Typically, an RAG pipeline would consist of processing the input text, retrieving the relevant information from external knowledge using retrieval techniques such as BM25 (Robertson and Zaragoza, 2009), followed by augmenting the original input with the retrieved information and, therefore, using the created augmented context to generate the classification output using an LLM.  
\\[0.25cm]
The eligible studies in this systematic review include one research paper where RAG was leveraged for healthcare text classification. Ramteke and Khandelwal (2023) explored using LLMs from the GPT family for stress detection in a binary manner from social media posts. The co-authors compared the performance of GPT-4 and GPT-3.5 (zero-shot and few-shot) with the results of conventional machine learning methods. Multiple vectorization techniques (e.g., TF-IDF, OpenAI Embedding) were also experimented with for the latter.  To enhance few-shot learning, RAG was employed to identify semantically similar examples from the training data. This improved the prompt by retrieving relevant information with K-Nearest Neighbors (KNN), which was used for efficient search.  LLM-based approaches, notably GPT-4, significantly outperformed traditional machine learning models by reaching a recall exceeding 99\% for stress detection, which is particularly valuable in clinical settings. 

\begin{landscape}
\begin{table}[h]
\captionsetup{labelformat=empty}
\centering
\caption*{\textbf{Table 10.} Ensemble Learning-based Reviewed Literature Categorization}
\label{tab:classification_models}
\begin{tabular}{c c c c c c c c c c}
 \toprule
\multirow{2}{*}{\centering\textbf{Reference}} & 
\multirow{2}{*}{\centering\textbf{Type}} & 
\multirow{2}{*}{\centering\textbf{Language}} & 
\multirow{2}{*}{\centering\textbf{Best LLM}} & 
\multirow{2}{*}{\centering\textbf{Application}} & 
\multicolumn{5}{c}{\centering\textbf{Performance Evaluation Metrics}} \\
\cmidrule(lr){6-10}
& & & & & \textbf{Accuracy} & \textbf{F1 Score} & \textbf{Precision} & \textbf{Recall} & \textbf{AUC Score} \\

\hline
\makecell{\textbf{Yang et al.} \\ (2024)} & Journal	& English & GPT-4 & \makecell{Research/Literature \\ Analysis} &	\makecell{SARS-CoV-2: \\ 92.87\% \\
Nipah: \\ 87.40\%} &	\makecell{SARS-CoV-2: \\ 88.43\% \\
Nipah: \\ 73.90\%} & \makecell{SARS-CoV-2: \\ 83.38\% \\
Nipah: \\ 74.72\%} & 	- &	- \\

\hline
\makecell{\textbf{Li et al.} (2023a)}	& Conference &	English	 &	\makecell{BERT-based \\ Ensemble} &	\makecell{Clinical Decision \\ Support}	& \makecell{Binary: \\ 94\% \\
MC: \\ 72\%} &	\makecell{Binary: \\ 80\% \\
MC: -} &	\makecell{Binary: \\ 74\% \\
MC: -}	& \makecell{Binary: \\ 86\% \\
MC: -} &	- \\

\hline
\makecell{\textbf{Wu et al.} (2023)} &	Conference &	Chinese	& \makecell{MacBERT, \\ ChatGLM2-\\6B, and \\ Qwen-7B-Chat \\ Ensemble} &	\makecell{Patient Query \\ Analysis}	& 92\%	&-	&-	&-&	- \\

\hline
\multirow{3}{*}{\makecell{\textbf{Jiang et al.} \\ (2023)}} & \multirow{3}{*}{Conference} & \multirow{3}{*}{\makecell{English}} & ALEX-L & \multirow{2}{*}{\makecell{Clinical Decision \\ Support}} & Task 1: 96.71\% & Task 1: 94.97\% & - & - & - \\
\cmidrule(lr){6-10}
 &  &  &  &  & Task 4: 89.84\% & Task 4: 88.17\% & - & - & - \\
\cmidrule(lr){5-10}
 &  &  &  & \multirow{1}{*}{\makecell{Public Health and \\ Opinion Analysis}} & 77.84\% & 89.13\% & - & - & - \\

\hline
\makecell{\textbf{Chaichulee et al.} \\ (2022)}	& 	Journal		& \makecell{Thai \& English} 	& 	\makecell{NB-SVM, \\ ULMFiT, \\ and BERT\\-based Models \\ Ensemble}	& \makecell{Patient Safety and \\ Risk Assessment}	& 98.37\%	& 98.88\%	& 98.72\%	& 99.31\%	& - \\

\bottomrule
\end{tabular}
\end{table}
\end{landscape}

\begin{landscape}
\begin{table}[h]
\captionsetup{labelformat=empty}
\centering
\caption*{\textbf{Table 11.} Data Augmentation-based Reviewed Literature Categorization}
\label{tab:classification_models}
\begin{tabular}{c c c c c c c c c c}
 \toprule
\multirow{2}{*}{\centering\textbf{Reference}} & 
\multirow{2}{*}{\centering\textbf{Type}} & 
\multirow{2}{*}{\centering\textbf{Language}} & 
\multirow{2}{*}{\centering\textbf{Best LLM}} & 
\multirow{2}{*}{\centering\textbf{Application}} & 
\multicolumn{5}{c}{\centering\textbf{Performance Evaluation Metrics}} \\
\cmidrule(lr){6-10}
& & & & & \textbf{Accuracy} & \textbf{F1 Score} & \textbf{Precision} & \textbf{Recall} & \textbf{AUC Score} \\

\hline
\makecell{\textbf{Yuan et al. } (2023)} & Conference & English & ChatGPT & \makecell{Clinical Decision \\ Support} & - & \makecell{Criteria: 91\% \\
Trial:81.50\%} & \makecell{Criteria: 86.20\% \\
Trial: 83\%} & \makecell{Criteria: 96.40\% \\
Trial:80.10\%}	& - \\

\bottomrule
\end{tabular}
\end{table}

\begin{table}[h]
\captionsetup{labelformat=empty}
\centering
\caption*{\textbf{Table 12.} Retrieval Augmented Generation-based Reviewed Literature Categorization}
\label{tab:classification_models}
\begin{tabular}{c c c c c c c c c c}
 \toprule
\multirow{2}{*}{\centering\textbf{Reference}} & 
\multirow{2}{*}{\centering\textbf{Type}} & 
\multirow{2}{*}{\centering\textbf{Language}} & 
\multirow{2}{*}{\centering\textbf{Best LLM}} & 
\multirow{2}{*}{\centering\textbf{Application}} & 
\multicolumn{5}{c}{\centering\textbf{Performance Evaluation Metrics}} \\
\cmidrule(lr){6-10}
& & & & & \textbf{Accuracy} & \textbf{F1 Score} & \textbf{Precision} & \textbf{Recall} & \textbf{AUC Score} \\

\hline
\makecell{\textbf{Ramteke and} \\ \textbf{Khandelwal} (2023)} &	Conference & English	 &	GPT-4	 &	\makecell{Clinical Decision \\ Support} &		-	 &	>99\%	 &	- &		- & -\\

\bottomrule
\end{tabular}
\end{table}

\end{landscape}

\begin{landscape}
\begin{table}[h]
\captionsetup{labelformat=empty}
\centering
\caption*{\textbf{Table 13.} Text Classification and LLM Type Categorization: The Case of Ensemble Learning}
\label{tab:classification_models}
\begin{tabular}{c c c c c c c c c}
 \toprule
\multirow{2}{*}{\centering\textbf{Reference}} & 
\multicolumn{3}{c}{\makecell{\textbf{Text Classification Type}}} & 
\multicolumn{5}{c}{\makecell{\textbf{Large Language Model Type}}}
\\
\cmidrule(lr){2-9}

& \textbf{Multi-class}	& \textbf{Multi-label}	& \textbf{Binary} & \textbf{	BERT (or variant)}	& \textbf{Closed-source} &	\textbf{Open-source}	& \makecell{\textbf{Pre-trained} \\ \textbf{Transformer}}	& \textbf{BART} \\

\hline
\textbf{Yang et al.} (2024) &  &  &  \checkmark  &  &  \checkmark  & & & \\
\hline
\textbf{Li et al.} (2023a) &  \checkmark  &  &  \checkmark  &  \checkmark  &  & & & \\
\hline
\textbf{Wu et al. }(2023) &  \checkmark  &  &  &  \checkmark  &  &  \checkmark & & \\
\hline
\textbf{Jiang et al.} (2023) &  \checkmark  &  &  \checkmark  &  \checkmark  &  \checkmark  & & & \\
\hline
\textbf{Chaichulee et al.} (2022) &  &  \checkmark  &   &  \checkmark  &  & & & \\

\bottomrule
\end{tabular}
\end{table}

\begin{table}[h]
\captionsetup{labelformat=empty}
\centering
\caption*{\textbf{Table 14.} Text Classification and LLM Type Categorization: The Case of Data Augmentation}
\label{tab:classification_models}
\begin{tabular}{c c c c c c c c c}
 \toprule
\multirow{2}{*}{\centering\textbf{Reference}} & 
\multicolumn{3}{c}{\makecell{\textbf{Text Classification Type}}} & 
\multicolumn{5}{c}{\makecell{\textbf{Large Language Model Type}}}
\\
\cmidrule(lr){2-9}

& \textbf{Multi-class}	& \textbf{Multi-label}	& \textbf{Binary} & \textbf{	BERT (or variant)}	& \textbf{Closed-source} &	\textbf{Open-source}	& \makecell{\textbf{Pre-trained} \\ \textbf{Transformer}}	& \textbf{BART} \\

\hline
\textbf{Yuan et al.} (2023) &    \checkmark  &  & &  &  \checkmark  & & & \\
\bottomrule
\end{tabular}
\end{table}

\begin{table}[h]
\captionsetup{labelformat=empty}
\centering
\caption*{\textbf{Table 15.} Text Classification and LLM Type Categorization: The Case of Retrieval Augmented Generation}
\label{tab:classification_models}
\begin{tabular}{c c c c c c c c c}
 \toprule
\multirow{2}{*}{\centering\textbf{Reference}} & 
\multicolumn{3}{c}{\makecell{\textbf{Text Classification Type}}} & 
\multicolumn{5}{c}{\makecell{\textbf{Large Language Model Type}}}
\\
\cmidrule(lr){2-9}

& \textbf{Multi-class}	& \textbf{Multi-label}	& \textbf{Binary} & \textbf{	BERT (or variant)}	& \textbf{Closed-source} &	\textbf{Open-source}	& \makecell{\textbf{Pre-trained} \\ \textbf{Transformer}}	& \textbf{BART} \\

\hline
\makecell{\textbf{Ramteke and} \\ \textbf{Khandelwal} (2023)} &   &  &   \checkmark &  &  \checkmark  & & & \\

\bottomrule
\end{tabular}
\end{table}

\end{landscape}

\subsection{Performance Evaluation}
In the reviewed healthcare text classification literature, researchers employed various performance evaluation metrics, with accuracy-related measures being predominant. These include standard accuracy (or balanced accuracy), F1 score, recall, precision, and AUC score. Some studies also presented confusion matrices to provide a more detailed view of classification performance. All the reviewed papers used at least one of these conventional metrics to validate their proposed approach, with the exception of Xie et al. (2024) and Carneros-Prado et al. (2023), where the first opted for the Positive Class Balance (PCB) and the Negative Class Balance (NCB) and the second used confusion matrices.

\begin{table}[H]
\captionsetup{labelformat=empty}
\centering
\renewcommand{\arraystretch}{1.2}
\caption{\textbf{Table 16.} Performance Evaluation Metrics}
\begin{tabular}{c l l l}
\hline
\textbf{Performance Evaluation}	& \multicolumn{3}{c}{\textbf{Research Papers}} \\
\hline
\multirow{3}{*}{\textbf{Compute Time}} & • Gu et al. (2024) & • Chen et al. (2024b) & • Chen et al. (2022) \\

 & • Raja et al. (2024) & • Bețianu et al. (2024) & • Yogarajan et al. (2022a) \\

 & • Guo et al. (2024) & • Gretz et al. (2023) & • Yogarajan et al. (2022b) \\
\hline
\textbf{Perplexity} &  • Li et al. (2024b)	& • Li et al. (2023b)\\
\hline
\textbf{FLOPS} &  \multicolumn{3}{c}{Lehman et al. (2023)}\\
\hline
\end{tabular}
\end{table}

In addition to using accuracy metrics, researchers often evaluated their approaches using computational efficiency measures such as Computation Time, Perplexity, or FLOPS (Floating Point Operations Per Second). As shown in Table 16, Computation Time emerged as the most common of these metrics, appearing in nine papers. Although some studies discussed the Computation Time, detailed time comparisons were not provided, and, therefore, the studies were not included in the table. Perplexity was less frequently used, appearing in only two papers—one for pre-training evaluation and another for fine-tuning assessment. FLOPS received minimal attention, with only Lehman et al. (2023) examining this aspect. This limited focus on FLOPS is understandable given the healthcare domain context, as this metric primarily serves model architecture comparison and hardware optimization purposes, which often require detailed model architecture information that may be unavailable for many LLMs. While some authors mentioned implementation costs, these references typically lacked explicit comparisons and were therefore excluded from the analysis in Table 16.

\subsection{Gaps and Limitations}
Despite the significant advances and insights in the reviewed literature, several significant gaps and limitations require careful consideration. This section summarizes these limitations across four key dimensions: data-related challenges that impact the quality and reliability of findings, model-related constraints that affect the computational approaches employed, methodological limitations that influence the implementation, and ethical and privacy considerations that raise essential concerns for future research and applications in healthcare.

\newpage

\subsubsection{Data-related}
\begin{figure}[H]
    \centering
    \includegraphics[scale=0.6]{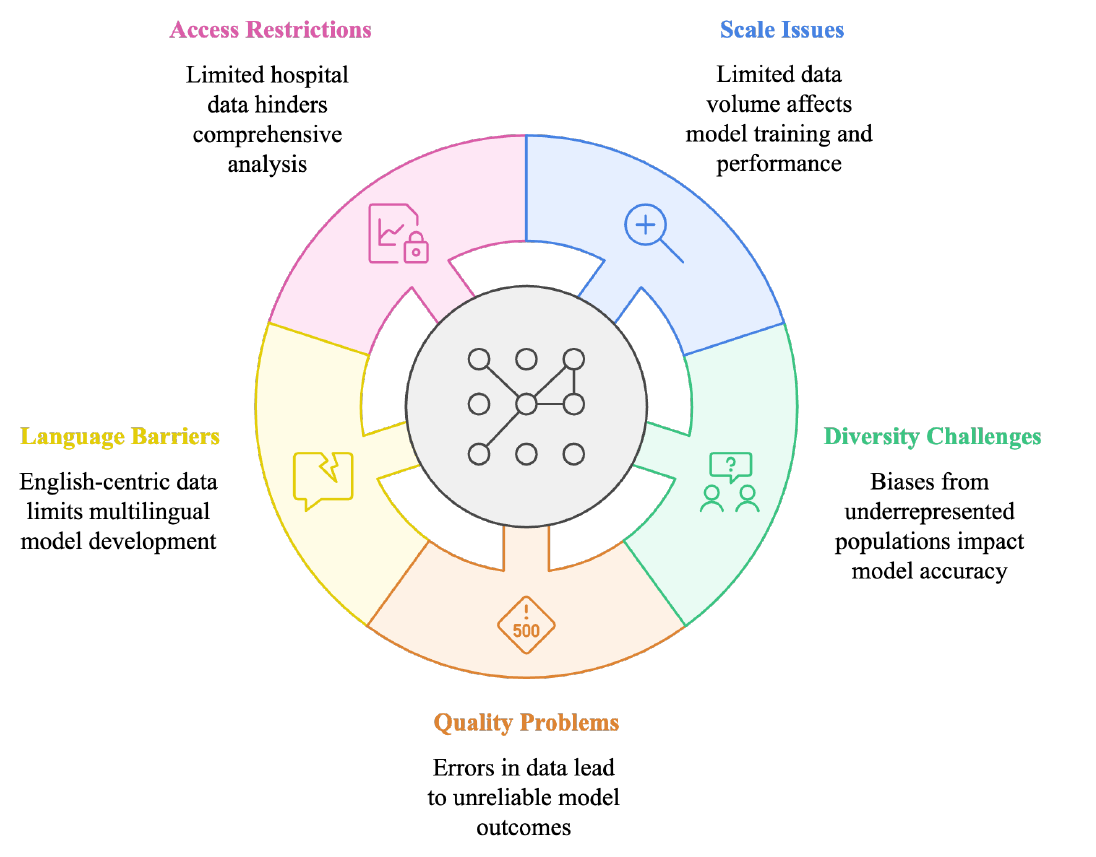}
    \caption*{\textbf{Figure 12.} Data-related Limitations}
    \label{fig:mltc-approaches}
\end{figure}

Healthcare text classification using LLMs faces several significant data-related limitations, primarily concerning scale, diversity, and quality. Many research studies rely on limited datasets or healthcare documents from single institutions, potentially introducing bias and restricting the generalizability of results across different healthcare settings, particularly in pre-training or fine-tuning approaches. The lack of demographic diversity presents another crucial challenge. Underrepresented populations may experience reduced model performance due to training data biases. Data quality issues further aggravate these limitations, resulting from patient de-identification errors, transcription inaccuracies, and inconsistent user-generated content for instance. These challenges become particularly critical when dealing with minority classes representing rare conditions, especially in fine-tuning scenarios where imbalanced datasets can significantly skew the LLM-based approach performance. Language constraints also pose significant barriers, as research often focuses on English-language data, hindering the development of multilingual approaches. Additionally, researchers often rely on publicly available datasets (e.g., MIMIC-III) due to limited access to hospital-specific data. This restriction prevents access to comprehensive patient histories and complete clinical contexts, therefore affecting the development, validation, and practical implementation of the LLM-based approaches proposed.

\subsubsection{Model-related}
\begin{figure}[H]
    \centering
    \includegraphics[scale=0.5]{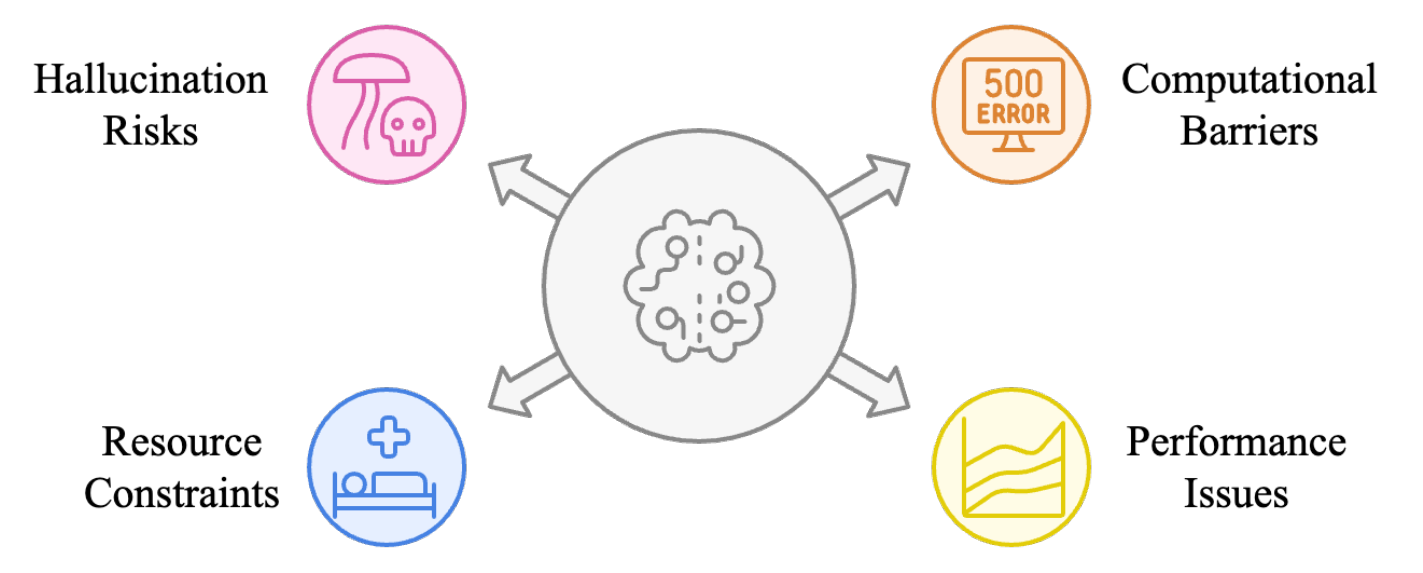}
    \caption*{\textbf{Figure 13.} Model-related Limitations}
    \label{fig:mltc-approaches}
\end{figure}

\newpage

Among the key limitations from which LLMs suffer, significant challenges such as computational and architectural constraints can be highlighted. The complex architecture of LLMs, especially when it comes to fine-tuning and pre-training, requires substantial computational resources. Therefore, this creates significant implementation barriers in healthcare settings and justifies the prevalence of using prompt engineering and BERT (or variant) fine-tuning approaches. However, even these less computationally extensive approaches encounter challenges. For instance, processing long healthcare documents (especially clinical) due to context window limitations requires introducing a chunking strategy, potentially resulting in slower inference speeds that can delay real-time applications. Performance issues can be found through various findings, including lower accuracy for rare medical conditions and challenges with (extreme) multi-label classification. LLMs’ tendency to hallucinate information is particularly concerning in clinical settings where accuracy is crucial. Furthermore, resource constraints in healthcare organizations further complicate implementation. Therefore, memory limitations often force researchers to use smaller LLM variants, thus potentially compromising performance. Additionally, the high operational costs of advanced LLMs like OpenAI o1 can create a barrier for practical applications. The discussed technical limitations, combined with the difficulties encountered when handling the medical jargon, further highlight the gap between current technical capabilities and the demanding requirements of healthcare applications.

\subsubsection{Methodology-related}
\begin{figure}[H]
    \centering
    \includegraphics[scale=0.6]{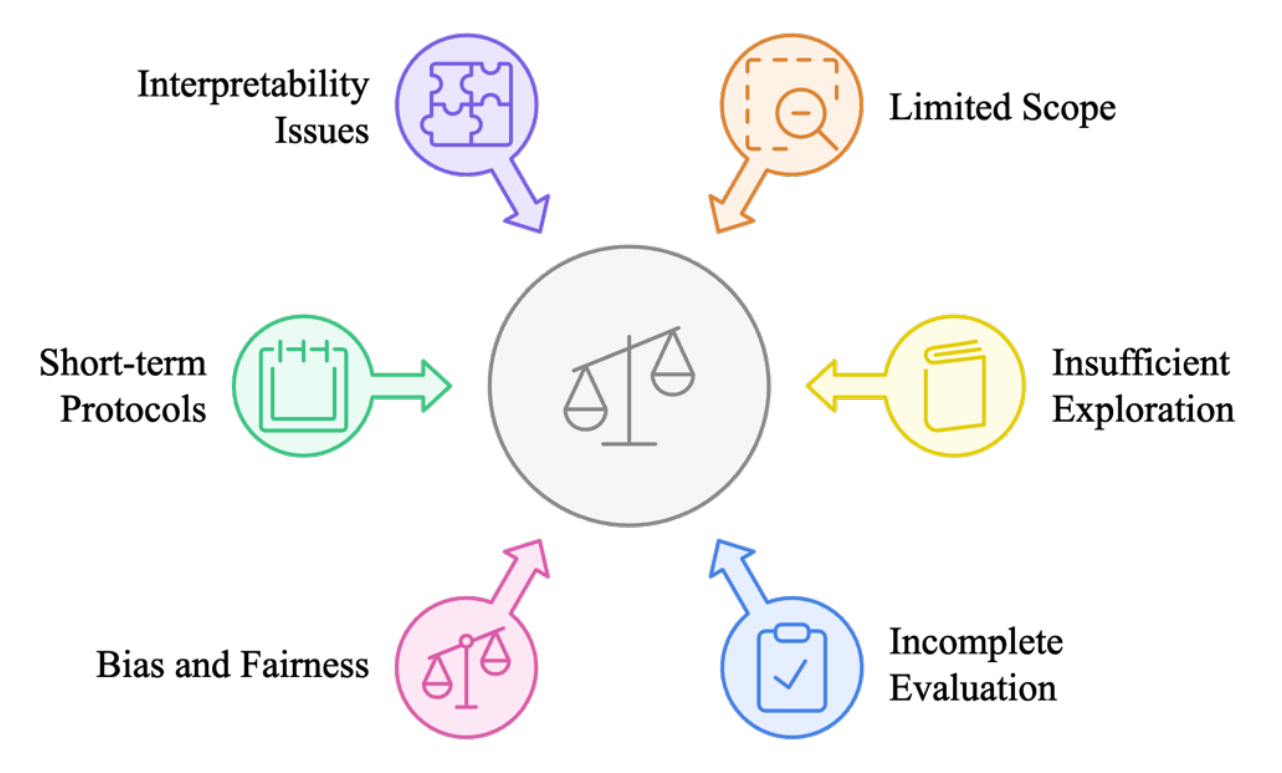}
    \caption*{\textbf{Figure 14.} Methodology-related Limitations}
    \label{fig:mltc-approaches}
\end{figure}

Methodology-related limitations constitute another challenge facing LLM-based healthcare text classification. These encompass gaps in the research approach’s design, evaluation, and validation. Many studies demonstrate limited scope by focusing on single-text classification types and specific healthcare applications rather than exploring a broader range of classification tasks and applications. This is mainly due to the research being driven by specific institutional needs, such as addressing individual healthcare facility requirements. Additionally, there is often insufficient exploration of advanced techniques and limited comparison with traditional machine learning methods, thus making it difficult to establish true performance benchmarks. Evaluation frameworks often lack comprehensiveness, with many studies neglecting crucial metrics like computation time and cost, which are particularly important considerations when combining healthcare applications with LLMs. Notable gaps also include minimal attention to bias and fairness assessment. Some studies rely solely on comparisons with other LLM outputs rather than human annotations, making the assessment of real-world implementation challenging. Furthermore, studies frequently lack robust, relatively long-term evaluation protocols, making it challenging to assess the model’s stability and reliability over time. Additionally, there's often insufficient investigation of the LLMs’ interpretability, which is crucial for clinical applications and also opens the door to an active research area. These methodological limitations eventually impact the LLM-based approach's reliability, validity, and practical applicability for healthcare text classification.

\newpage

\subsubsection{Ethics and Privacy-related}
\begin{figure}[H]
    \centering
    \includegraphics[scale=0.6]{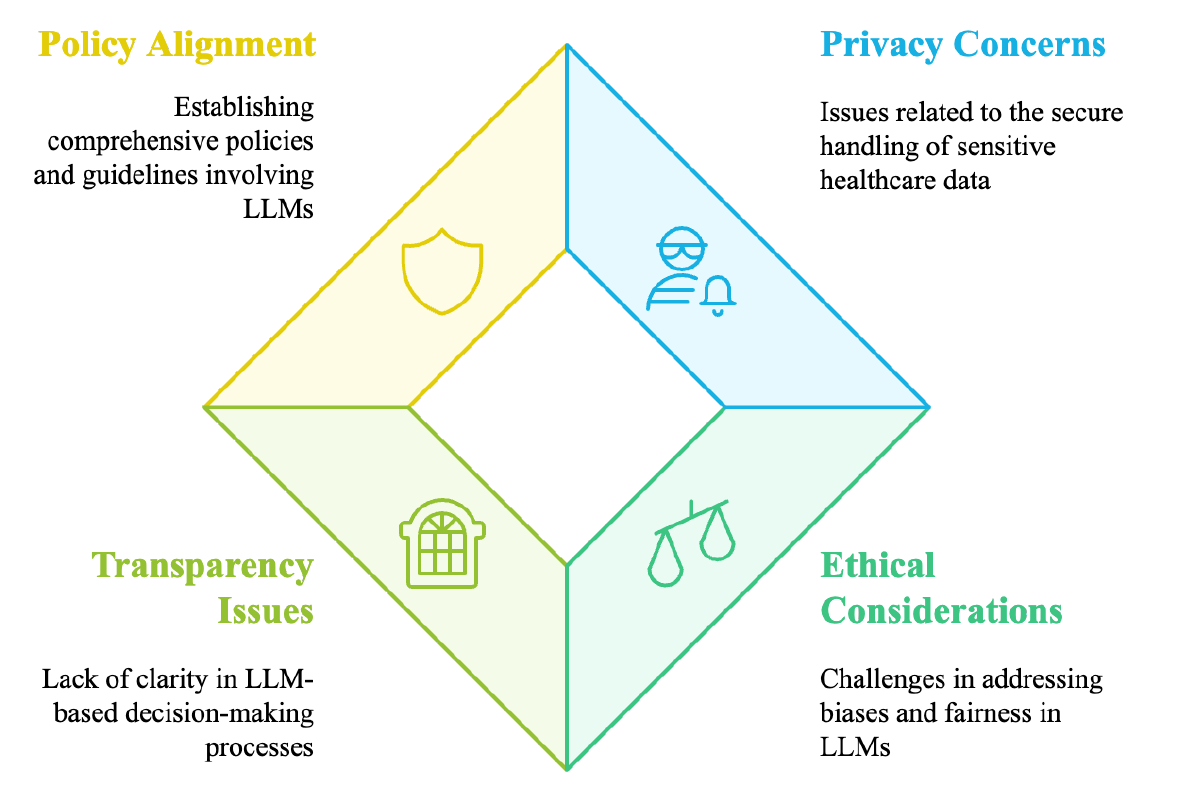}
    \caption*{\textbf{Figure 15.} Ethics and Privacy-related Limitations}
    \label{fig:mltc-approaches}
\end{figure}

Two key challenges and priority considerations when handling healthcare text data classification are protecting patient privacy and ensuring ethical implementation. Privacy concerns are particularly critical when dealing with sensitive clinical notes through cloud-based LLMs that rely on shared computing resources and API requests. One fundamental gap revolves around the lack of attention to data protection requirements and limited exploration of privacy-preserving techniques, creating potential vulnerabilities in patient data security. Furthermore, since the primary concern in this type of research is usually ensuring high text classification accuracy, ethical considerations frequently receive insufficient attention. As a result, the majority of the studies fail to address the potential biases in the models and datasets used, which can lead to healthcare disparities. Additionally, there is a clear lack of transparency and interpretability in LLM-based decision-making processes, which raises ethical concerns about accountability and trust in clinical settings. Also, there is a need to address the ethical implications of automated medical decision support via LLMs, including issues of consent for data sharing and the appropriate balance between the results obtained from an LLM-based approach and human judgment. Moreover, given the novelty of these approaches, there is a need to establish comprehensive policies and guidelines governing LLM use in healthcare settings.

\newpage

\section{Future Research Directions}
The previous sections show that LLMs have demonstrated remarkable capabilities in the healthcare text classification task; however, several gaps and limitations were revealed. Thanks to the rapid evolution of LLM architectures and the development of computational resources, further investigation could be enabled, and the doors to exploring areas revolving around advancing LLM robustness can be opened. This includes approach-based improvements, efficiency optimization, data-related contributions, and clinical practical implementation, as shown in Figure 16. In this section, these key research directions are examined.

\begin{figure}[H]
    \centering
    \includegraphics[scale=0.75]{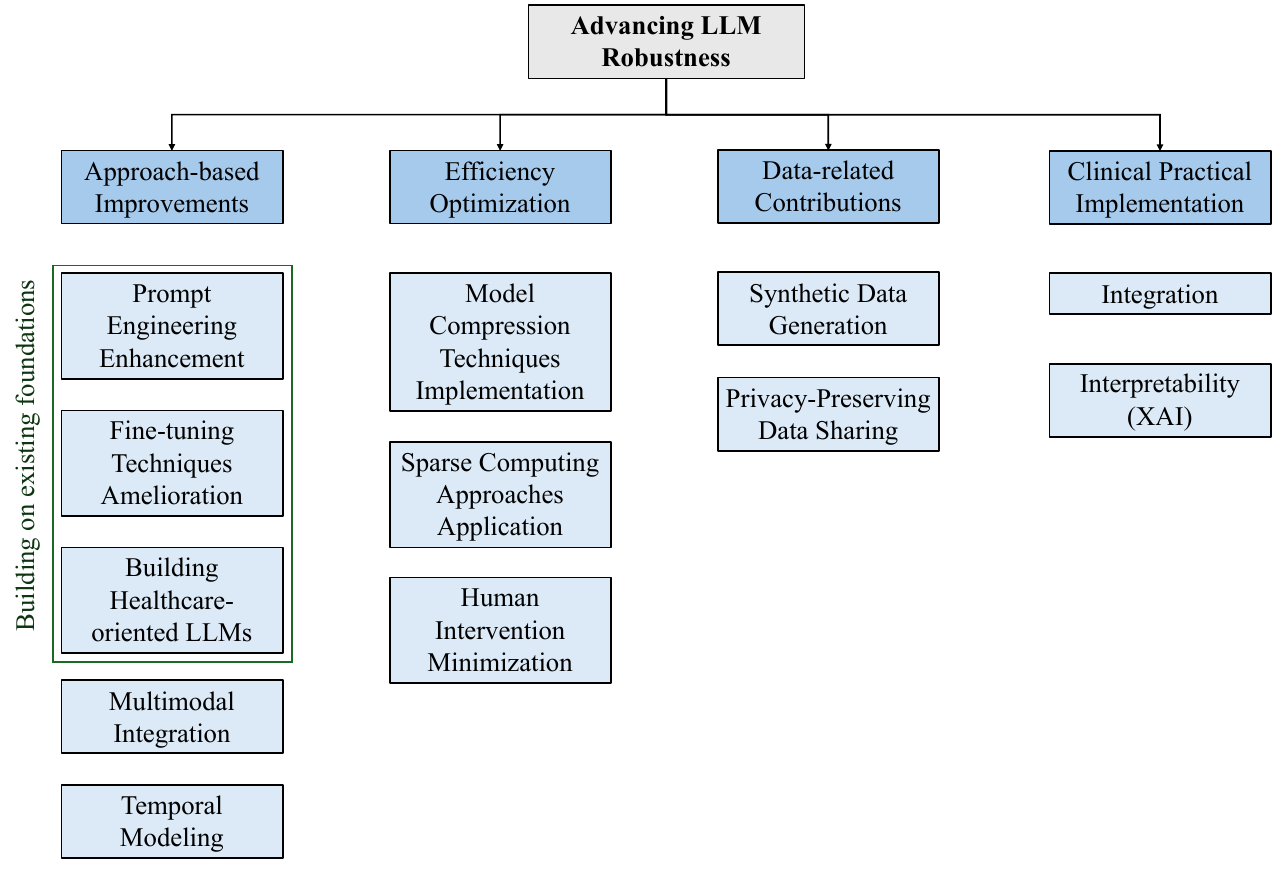}
    \caption*{\textbf{Figure 16.} Key Future Research Directions}
    \label{fig:mltc-approaches}
\end{figure}

\subsection{Approach-based Improvements}
When considering future research directions, building on the existing research foundation is typically the first priority. This involves focusing on approach-based improvements, including advanced investigation of prompt engineering and fine-tuning techniques, and ideally, developing healthcare-oriented LLMs from scratch (e.g., GatorTron). Regarding prompt engineering enhancement, developing sophisticated prompt strategies by creating templates that incorporate medical terminologies would potentially be beneficial in providing more context to the LLM’s input. Furthermore, smaller LLMs can be used to guide larger LLMs hierarchically, such as using BERT to extract the keywords of the text documents through the attention mechanism to construct an enhanced prompt to thereafter feed to GPT-4o, for instance.  Additionally, carrying out a text classification requires first to identify the labels. In multiple real-world applications, such as the case of patient comments collected from surveys, labels are at first missing. Topic modeling is usually required before proceeding with a classification. Therefore, an exciting research area would be to leverage LLMs for end-to-end text classification, including topic modeling. Fine-tuning, the most commonly used LLM-based approach in the reviewed literature is useful when the labeled data is limited, which is often the case. Thus, it is a solution that can achieve high performance with minimal data requirements. Therefore, more attention is needed to develop efficient fine-tuning techniques that can leverage the potential of LLMs that, unfortunately, cannot be adapted using standard supervised fine-tuning due to their large number of parameters. Researchers are currently focusing on tackling this challenge through PEFT (Liu et al., 2022). Further exploring parameter-efficient techniques suitable for resource-constrained healthcare environments and investigating transfer learning approaches that can effectively bridge different medical domains and tasks would undoubtedly be valuable. It is an active research area with the presence of PEFT techniques such as LoRA, which was occasionally resorted to in the literature when larger models were involved. Further down the road of research, building more LLMs from scratch exclusively trained on healthcare data and collected from various sources and institutions will potentially reduce bias and ensure fairness. Eventually, variants will be tuned to different NLP tasks, such as text classification and named entity recognition. Researchers can also consider developing smaller, more focused models that can excel in specific medical areas or tasks, potentially offering more practical solutions for real-world healthcare applications. Targeting Small Language Models (SLMs) would allow efficient fine-tuning that can be conducted on-premises using their data in a secure environment. Future research must also prioritize developing robust frameworks to evaluate both accuracy and cost-efficiency.
\\[0.25cm]
As the collected healthcare data grows, thanks to the digitalization of healthcare processes, it increasingly spans multiple modalities beyond just text. As a result, multimodal integration starts gaining attraction as a promising research area similar to other industries. Although this systematic review excluded the few cases where multimodal data was used, this approach turns out to recognize that the healthcare information, particularly in the case of clinical decision support application, can often be found in various forms (e.g., clinical notes, medical images, lab results, vital sign readings, and structured EHR data) and that once integrated they can provide a more comprehensive context to the LLM’s text classification. A key research challenge in multimodal integration involves developing effective architectures that can handle the heterogeneity of different data types while maintaining their semantic relationships, aiming for a contextually rich classification. This includes exploring various fusion strategies, such as early fusion, where raw inputs are combined, and late fusion, consisting of processing each modality separately and combining their outputs at the decision level. However, broadening the range of healthcare data modality may raise the missing data challenge that would require special handling. 
\\[0.25cm]
One research paper in the reviewed literature highlighted the study of temporal relationships in discharge summaries (Cui et al., 2023), revealing an exciting research area focusing on capturing the dynamic and sequential nature of healthcare data, potentially enabling LLMs to better understand text classification applications such as disease progression or treatment responses. The temporal dimension would add a crucial context that can significantly impact the accuracy and utility of classification tasks in these specific cases. One key aspect of temporal modeling is the emphasis on enhancing LLMs' ability to understand and process temporal expressions in clinical texts, such as the temporal markers (i.e., dates, times, durations) and the implicit temporal relationships (i.e., before, after, during) that are very present in clinical notes for instance. Researchers could explore techniques to modify LLM architectures to better capture these temporal dependencies, such as incorporating temporal attention mechanisms that would enable maintaining the chronological order of clinical events while performing classification tasks. Another important research avenue involves developing methods for handling longitudinal patient data. Traditional LLM approaches often treat each clinical note independently, but temporal modeling seeks to maintain continuity across the patient’s medical history, requiring sophisticated architectures to track the evolution of the patient’s medical status while maintaining consistent good classification performance. For this purpose, incorporating Recurrent Neural Network components or Temporal Convolution Layers within the LLM’s architecture to capture long-term dependencies in patient histories may be worth exploring. 

\subsection{Efficiency Optimization}
LLMs' computational demands and resource requirements present significant challenges for practical implementation in healthcare settings, especially in pre-training and fine-tuning cases. Although the literature provides proof of concepts for the success of LLMs application in healthcare text classification, several key research directions emerge as the field evolves, particularly in efficiency optimization. This section explores potential solutions, including model compression techniques, sparse computing approaches, and the minimization of human intervention. 
\\[0.25cm]
Implementing model compression techniques is crucial in optimizing LLMs for healthcare text classification. Model compression refers to techniques that reduce models' size and computational demands while maintaining as much of their accuracy as possible. Researchers can explore specialized knowledge distillation approaches that can effectively transfer healthcare domain knowledge from a larger model (teacher) to a more compact one (student) while maintaining accuracy. This area requires careful investigation of healthcare-specific teacher-student architectures and domain-adapted distillation techniques that can preserve critical healthcare knowledge. Additionally, quantization can be leveraged for efficiency optimization. Quantization consists of reducing the precision of the LLM's parameters to optimize for memory and computational efficiency, often without a significant loss in performance. Future studies can examine mixed-precision techniques that can adapt to varying healthcare text classification tasks, with particular attention to their impact on rare medical condition detection and diagnostic confidence scores, for instance. Post-training quantization research can also focus on developing calibration methods designed specifically for healthcare text data. Furthermore, pruning strategies, consisting of removing less important weights from the LLM, need to be developed with healthcare-aware metrics that consider the unique requirements of healthcare applications, particularly for maintaining accuracy in detecting minority classes (e.g., rare conditions).
\\[0.25cm]
Another approach that can help with optimizing LLMs for healthcare text classification is sparse computing. Sparse computing refers to computational methods and hardware architectures optimized for handling sparse data (e.g., text, recommendation systems). It is particularly potentially useful when dealing with healthcare text data since it often contains specialized vocabulary. Still, most words from the entire medical vocabulary are usually absent in each text, which results in sparse representation matrices. Additionally, in terms of distribution, healthcare documents can be considered following Zipf's law, where few words/tokens appear frequently while most appear rarely. Therefore, using sparse computing techniques would particularly be helpful in the case of clinical notes, which are more prone to include medical jargon and be lengthy. Dynamic Sparse Attention mechanisms (Liu et al., 2021) could help LLMs focus on clinically relevant terms and relationships while ignoring irrelevant text, potentially improving both efficiency and accuracy. Conditional computation approaches, where only relevant parts of the LLM are activated based on the specific healthcare classification task or document type, could significantly reduce computational consumption.
\\[0.25cm]
Minimizing human intervention presents an opportunity and challenge that needs careful investigation. It would enable focusing on developing robust self-verification mechanisms, uncertainty quantification, and automated quality control processes that would allow LLMs to handle healthcare text classifications that are routinely conducted independently. Integrating continuous learning mechanisms would enable LLMs to adapt to evolving medical knowledge, terminology, and classification requirements. Active learning strategies could also optimize human involvement by intelligently selecting the most notable cases for expert review. This approach would enable a targeted use of limited healthcare professional time by identifying cases where the LLM’s confidence is low.

\subsection{Data-related Contributions}
As previously discussed, LLM-based healthcare text classification faces several data challenges, leading to the introduction of multiple research directions, including synthetic data generation and privacy-preserving data sharing. Synthetic data generation using LLMs represents a transformative approach to potentially addressing critical challenges in healthcare text classification. In healthcare settings, where data accessibility is often constrained by privacy regulations and ethical considerations, LLMs offer a promising solution by generating realistic textual data that can supplement existing datasets while maintaining patient confidentiality. Once again, this research direction would be particularly useful for pre-training and fine-tuning approaches. The primary advantage of LLM-based synthetic data generation lies in its ability to address data scarcity, particularly for the minority class (e.g., rare medical conditions). Healthcare datasets frequently suffer from imbalanced distributions, where certain conditions or topics are underrepresented. LLMs can generate additional examples of these less frequent cases for data augmentation, helping to create more balanced datasets that would help improve the overall classification performance. Another crucial benefit of synthetic data generation in healthcare would be privacy preservation. Leveraging synthetic text data in the LLM-based approach development would eliminate the risk of exposing patients’ sensitive information and would not require their formal consent. However, synthetic data generation comes with various challenges, including its validity in terms of medical accuracy and proper use of terminology, in addition to the bias that may be inherited or even amplified from the LLM that would only lead to perpetuating existing disparities in healthcare text data. Techniques aiming to mitigate these biases would be worth investigating. Moreover, recursively training the LLM on synthetically generated data would most likely lead to model collapse (Shumailov et al., 2024), causing a degradation in performance mainly due to the LLM gradually losing information about the data distribution’s tails (i.e., minority classes) and shifting toward a distribution with reduced variance. Therefore, careful consideration should be taken to advance healthcare-oriented LLMs for text classification without necessarily being limited by access to the data through synthetically generated healthcare text.
\\[0.25cm]
One of the previously identified gaps in the literature refers to limited datasets lacking diversity since the texts are usually collected from a single institution. Privacy-preserving data sharing between healthcare facilities represents a critical research direction for advancing LLM applications in healthcare text classification while protecting sensitive patient information. Traditional data-sharing approaches often involve direct exposure of raw patient records, which raises significant privacy concerns and legal compliance issues. Promising techniques are being developed, making privacy-preserving data sharing an active area of investigation. Federated learning (Loftus et al., 2022) would allow multiple healthcare institutions to collaboratively train LLMs on their local data without directly sharing patient records. The model parameters are aggregated centrally while keeping the underlying training data distributed and private. Furthermore, differential privacy (Ficek et al., 2021) techniques can be integrated into the training process to add carefully calibrated noise that prevents individual patient re-identification while preserving the population-level patterns that are needed for accurate classification. The ultimate goal is to enable healthcare institutions to safely leverage their collective data resources through LLMs while maintaining rigorous privacy standards. This would accelerate improving LLMs’ performance while ensuring fairness and mitigating biases across different healthcare text classification tasks and applications.

\subsection{Clinical Practical Implementation}
Practically implementing an LLM-based approach for healthcare text classification, especially in a clinical setting, has always encountered challenges. This section discusses clinical integration and interpretability (XAI) as fundamental ongoing and future research directions. Clinical integration represents a critical frontier in translating LLM capabilities into real-world healthcare settings for text classification tasks. One key element is understanding how healthcare professionals would interact with LLM systems if they were to be integrated. This includes investigating how clinicians would interpret and use LLM-generated classifications and how clinical decision-making would be influenced. This would lead to identifying interaction patterns that would be used to design interfaces and workflows that would enhance the overall process without being subject to significant resistance. Moreover, real-time performance will be required if the LLM outputs must be used to assist decision-making. Research in this area can examine how to optimize LLM-based approach inference times while maintaining accuracy, ensuring that text classification results are available when needed for time-sensitive clinical decisions. Additionally, researchers must investigate approaches for monitoring system performance over time and detecting potential degradation in classification accuracy. Another critical element is ensuring LLM-based classification systems can effectively communicate with existing healthcare IT infrastructure, guaranteeing a seamless information flow across different platforms and departments that would benefit from the LLM’s outputs. Addressing the organizational impacts in advance regarding cost-effectiveness, return on investment, and the broader organizational changes required to integrate LLM-based approaches into healthcare delivery systems successfully should also not be neglected.
\\[0.25cm]
The practical clinical implementation of LLMs for text classification in healthcare settings requires established interpretability and explainability frameworks, as these systems directly impact patient care decisions. The black-box nature of these approaches makes it challenging for healthcare professionals to interpret the output since healthcare applications require transparent reasoning processes that clinicians can trust and validate. Multiple approaches should be investigated in an attempt to provide more explainable generative artificial intelligence (XAI) for healthcare text classification outputs.  To not only explain the classifications but also to create explanations that can be used for LLM’s performance and trustworthiness enhancement, Wu et al. (2024) summarized several key strategies that researchers can resort to for explaining LLM text classification decisions. Attribution methods such as gradient-based and perturbation-based approaches would help identify important input features. At the same time, component-level interpretation would serve to analyze self-attention patterns to understand how the model processes input text or interpret feed-forward networks to understand how information flows through the model. Furthermore, CoT prompting, used by some of the reviewed research studies, can be used to add explicit reasoning steps to prompts to make classification decisions more interpretable while getting the model to explain its classification reasoning in a step-by-step manner. From a practical implementation standpoint, XAI solutions must be integrated into healthcare workflows through interactive explanation interfaces allowing healthcare professionals to understand the model’s decisions at varying levels of detail. Additionally, the explanations must be generated in real-time to support time-sensitive clinical decisions for instance.

\section{Conclusions}
Shifting from traditional machine learning approaches leveraging healthcare structured data (Sakai et al. (2022a), Sakai et al. (2022b), Sakai et al. (2023b)), LLMs have demonstrated remarkable potential in healthcare text classification, achieving high-performance metrics (mainly accuracy-related). The evidence reveals that these models consistently outperform traditional machine learning approaches in handling complex medical text classification tasks, with particular strength in understanding context and medical terminology. Researchers have explored implementing various approaches, from lightweight prompt engineering to moderately intensive prompt-tuning and fine-tuning to resource-intensive methods like pre-training, in addition to other LLM-based approaches such as ensemble learning and RAG. This diversity of approaches highlights healthcare text classification as an active LLM research area. A key advantage of LLM-based approaches is their ability to enable rapid deployment without requiring extensive labeled datasets or numerous contributors. This is particularly valuable in healthcare settings, where annotated training data is often scarce and computational resources may be limited. However, multiple gaps and limitations must be acknowledged. For instance, the black box nature of these models presents challenges for interpretability, a crucial factor in healthcare applications where understanding decision-making processes is essential, especially when it comes to clinical applications. Additionally, the sensitive nature of patients’ data presents a constant challenge that will always require careful consideration. These represent only a subset of the identified limitations. Potential future research lines were also discussed in this systematic review, attempting to address the identified gaps. As research continues, focusing on addressing current limitations of healthcare text classification while maintaining high performance will not only be crucial for realizing the full potential of LLMs but also a trade-off that researchers and healthcare professionals should carefully navigate.

\section*{References}
\begin{enumerate}[label={[\arabic*]}, leftmargin=*]
    \item Zhao, W. X., Zhou, K., Li, J., Tang, T., Wang, X., Hou, Y., Min, Y., Zhang, B., Zhang, J., Dong, Z., Du, Y., Yang, C., Chen, Y., Chen, Z., Jiang, J., Ren, R., Li, Y., Tang, X., Liu, Z., Liu, P., Nie, J.-Y., \& Wen, J.-R. (2024). A survey of large language models. arXiv. https://arxiv.org/abs/2303.18223.
    \item Brown, T. B., Mann, B., Ryder, N., Subbiah, M., Kaplan, J., Dhariwal, P., Neelakantan, A., Shyam, P., Sastry, G., Askell, A., Agarwal, S., Herbert-Voss, A., Krueger, G., Henighan, T., Child, R., Ramesh, A., Ziegler, D. M., Wu, J., Winter, C., Hesse, C., Chen, M., Sigler, E., Litwin, M., Gray, S., Chess, B., Clark, J., Berner, C., McCandlish, S., Radford, A., Sutskever, I., Amodei, D. (2020). Language models are few-shot learners. arXiv. https://arxiv.org/abs/2005.14165.
    \item Thoppilan, R., De Freitas, D., Hall, J., Shazeer, N., Kulshreshtha, A., Cheng, H., Jin, A., Rishi, H., Scott, H., Zeng, L., Han, Y., Laudon, T., Ma, Y., Ravindran, D., Winsor, P., Duke, J., Soraker, B., Zevenbergen, B., Riesa, J., Saeta, P., Wang, Y., Merrill, J., Prabhakaran, V., Huang, Y., Hutchinson, M., Nilforoshan, H., Molina, A., Stüker, S., Liao, H., Ganesh, A., Liu, W., Kitaev, O., Tsueng, Z., Cable, X., Coté, F., Kuzmina, V., Christmas, J., Kumar, S., McAllister, R., Tanskanen, J., Tang, Z., Deng, L., Malysheva, A., Schoelkopf, J., Tamblyn, E., Kerr, T., Croak, M., Chi, E., \& Le, Q. (2022). LaMDA: Language models for dialog applications. arXiv preprint arXiv:2201.08239.
    \item Touvron, H., Lavril, T., Izacard, G., Martinet, X., Lachaux, M., Lacroix, T., Rozière, B., Goyal, N., Hambro, E., Azhar, F., Rodriguez, A., Joulin, A., Grave, E., \& Lample, G. (2023). LLaMA: Open and Efficient Foundation Language Models. arXiv preprint arXiv:2302.13971.
    \item Chowdhery, A., Narang, S., Devlin, J., Bosma, M., Mishra, G., Roberts, A., Barham, P., Chung, H. W., Sutton, C., Gehrmann, S., \& Schuh, P. (2023). PaLM: Scaling language modeling with pathways. Journal of Machine Learning Research, 24(240), 1-113.
    \item OpenAI. (2020, November 30). "Introducing ChatGPT." OpenAI. https://openai.com/blog/chatgpt.
    \item Vaswani, A. (2017). Attention is all you need. Advances in Neural Information Processing Systems.
    \item Raschka, S. (2024). Build a large language model (from scratch). Manning Publications.
    \item Ghali, M. K., Farrag, A., Sakai, H., Baz, H. E., Jin, Y., \& Lam, S. (2024). Gamedx: Generative ai-based medical entity data extractor using large language models. arXiv preprint arXiv:2405.20585.
    \item Sebastiani, F. (2002). Machine learning in automated text categorization. ACM computing surveys (CSUR), 34(1), 1-47.
    \item Pang, B., Lee, L., \& Vaithyanathan, S. (2002). Thumbs up? Sentiment classification using machine learning techniques. arXiv preprint cs/0205070.
    \item Dada, E. G., Bassi, J. S., Chiroma, H., Adetunmbi, A. O., \& Ajibuwa, O. E. (2019). Machine learning for email spam filtering: review, approaches and open research problems. Heliyon, 5(6).
    \item Nobata, C., Tetreault, J., Thomas, A., Mehdad, Y., \& Chang, Y. (2016, April). Abusive language detection in online user content. In Proceedings of the 25th international conference on world wide web (pp. 145-153).
    \item Spasic, I., \& Nenadic, G. (2020). Clinical text data in machine learning: systematic review. JMIR medical informatics, 8(3), e17984.
    \item Karimi, S., Dai, X., Hassanzadeh, H., \& Nguyen, A. (2017, August). Automatic diagnosis coding of radiology reports: a comparison of deep learning and conventional classification methods. In BioNLP 2017 (pp. 328-332).
    \item Bittar, A., Velupillai, S., Roberts, A., \& Dutta, R. (2019). Text classification to inform suicide risk assessment in electronic health records. In MEDINFO 2019: Health and Wellbeing e-Networks for All (pp. 40-44). IOS Press.
    \item Sakai, H., Mikaeili, M., Lam, S. S., \& Bosire, J. (2023a). Text Classification for Patient Experience Improvement: A Neural Network Approach. In IISE Annual Conference. Proceedings (pp. 1-6). Institute of Industrial and Systems Engineers (IISE).
    \item Sakai, H., Lam, S. S., Mikaeili, M., \& Bosire, J. (2024a). Patient Experience Feedback Sentiment Analysis: Combining BERT and LSTM with Genetic Algorithm Optimization. In IISE Annual Conference. Proceedings (pp. 1-6). Institute of Industrial and Systems Engineers (IISE).
    \item Sakai, H., Lam, S. S., Mikaeili, M., Bosire, J., \& Jovin, F. (2024b). Large Language Models for Patient Comments Multi-Label Classification. arXiv preprint arXiv:2410.23528.
    \item Sakai, H., \& Lam, S. S. (2025). QUAD-LLM-MLTC: Large Language Models Ensemble Learning for Healthcare Text Multi-Label Classification. arXiv preprint arXiv:2502.14189.
    \item Apté, C., Damerau, F., \& Weiss, S. M. (1994). Automated learning of decision rules for text categorization. ACM Transactions on Information Systems (TOIS), 12(3), 233-251.
    \item McCallum, A., \& Nigam, K. (1998, July). A comparison of event models for naive bayes text classification. In AAAI-98 workshop on learning for text categorization (Vol. 752, No. 1, pp. 41-48).
    \item Joachims, T. (1998, April). Text categorization with support vector machines: Learning with many relevant features. In European conference on machine learning (pp. 137-142). Berlin, Heidelberg: Springer Berlin Heidelberg.
    \item Lewis, D. D., \& Ringuette, M. (1994, April). A comparison of two learning algorithms for text categorization. In Third annual symposium on document analysis and information retrieval (Vol. 33, pp. 81-93).
    \item Xu, B., Guo, X., Ye, Y., \& Cheng, J. (2012). An improved random forest classifier for text categorization. J. Comput., 7(12), 2913-2920.
    \item Rani, D., Kumar, R., \& Chauhan, N. (2022, October). Study and comparision of vectorization techniques used in text classification. In 2022 13th International Conference on Computing Communication and Networking Technologies (ICCCNT) (pp. 1-6). IEEE.
    \item Socher, R., Pennington, J., Huang, E. H., Ng, A. Y., \& Manning, C. D. (2011, July). Semi-supervised recursive autoencoders for predicting sentiment distributions. In Proceedings of the 2011 conference on empirical methods in natural language processing (pp. 151-161).
    \item Zhang, X., \& LeCun, Y. (2015). Text understanding from scratch. arXiv preprint arXiv:1502.01710.
    \item Lai, S., Xu, L., Liu, K., \& Zhao, J. (2015, February). Recurrent convolutional neural networks for text classification. In Proceedings of the AAAI conference on artificial intelligence (Vol. 29, No. 1).
    \item Devlin, J., Chang, M. W., Lee, K., \& Toutanova, K. (2019). BERT: Pre-training of deep bidirectional transformers for language understanding. Proceedings of the 2019 Conference of the North American Chapter of the Association for Computational Linguistics, 4171–4186.
    \item Radford, A. (2018). Improving language understanding by generative pre-training.
    \item Li, J., Dada, A., Puladi, B., Kleesiek, J., \& Egger, J. (2024a). ChatGPT in healthcare: a taxonomy and systematic review. Computer Methods and Programs in Biomedicine, 108013.
    \item Wang, B., Xie, Q., Pei, J., Chen, Z., Tiwari, P., Li, Z., \& Fu, J. (2023). Pre-trained language models in biomedical domain: A systematic survey. ACM Computing Surveys, 56(3), 1-52.
    \item Sallam, M. (2023). The utility of ChatGPT as an example of large language models in healthcare education, research and practice: Systematic review on the future perspectives and potential limitations. MedRxiv, 2023-02.
    \item Busch, F., Hoffmann, L., Rueger, C., van Dijk, E. H. C., Kader, R., Ortiz-Prado, E., Makowski, M. R., Saba, L., Hadamitzky, M., Kather, J. N., Truhn, D., Cuocolo, R., Adams, L. C., \& Bressem, K. K. (2024). Systematic review of large language models for patient care: Current applications and challenges. medRxiv. https://doi.org/10.1101/2024.03.04.24303733.
    \item Kesiku, C. Y., Chaves-Villota, A., \& Garcia-Zapirain, B. (2022). Natural Language Processing Techniques for Text Classification of Biomedical Documents: A Systematic Review. Information, 13(10), 499.
    \item Hossain, E., Rana, R., Higgins, N., Soar, J., Barua, P. D., Pisani, A. R., \& Turner, K. (2023). Natural language processing in electronic health records in relation to healthcare decision-making: a systematic review. Computers in biology and medicine, 155, 106649.
    \item Moher, D., Liberati, A., Tetzlaff, J., Altman, D. G., \& PRISMA Group, T. (2009). Preferred reporting items for systematic reviews and meta-analyses: the PRISMA statement. Annals of internal medicine, 151(4), 264-269.
    \item Sushil, M., Zack, T., Mandair, D., Zheng, Z., Wali, A., Yu, Y., Quan, Y., Lituiev, D., \& Butte, A. J. (2024). A comparative study of large language model-based zero-shot inference and task-specific supervised classification of breast cancer pathology reports. Journal of the American Medical Informatics Association, ocae146.
    \item Lossio-Ventura, J. A., Weger, R., Lee, A. Y., Guinee, E. P., Chung, J., Atlas, L., Linos, E., \& Pereira, F. (2024). A comparison of ChatGPT and fine-tuned Open Pre-Trained Transformers (OPT) against widely used sentiment analysis tools: Sentiment analysis of COVID-19 survey data. JMIR Mental Health, 11, Article e50150.
    \item Shi, Y., Ma, H., Zhong, W., Tan, Q., Mai, G., Li, X., Liu, T., \& Huang, J. (2023). Chatgraph: Interpretable text classification by converting chatgpt knowledge to graphs. In 2023 IEEE International Conference on Data Mining Workshops (ICDMW) (pp. 515-520). IEEE.
    \item Li, R., Wang, X., \& Yu, H. (2023a). Two directions for clinical data generation with large language models: data-to-label and label-to-data. In Proceedings of the Conference on Empirical Methods in Natural Language Processing. Conference on Empirical Methods in Natural Language Processing (Vol. 2023, p. 7129). NIH Public Access.
    \item Chen, Z., Mao, H., Li, H., Jin, W., Wen, H., Wei, X., Wang, S., Yin, D., Fan, W., Liu, H., \& Tang, J. (2024). Exploring the potential of large language models (LLMs) in learning on graphs. ACM SIGKDD Explorations Newsletter, 25(2), 42-61.
    \item Ohse, J., Hadžić, B., Mohammed, P., Peperkorn, N., Danner, M., Yorita, A., Kubota, N., Rätsch, M., \& Shiban, Y. (2024). Zero-Shot Strike: Testing the generalisation capabilities of out-of-the-box LLM models for depression detection. Computer Speech \& Language, 88, Article 101663.
    \item BT, B., \& Chen, J. M. (2024). Performance Assessment of ChatGPT versus Bard in Detecting Alzheimer’s Dementia. Diagnostics, 14(8), 817.
    \item Aldeen, M., Luo, J., Lian, A., Zheng, V., Hong, A., Yetukuri, P., \& Cheng, L. (2023, December). ChatGPT vs. Human Annotators: A Comprehensive Analysis of ChatGPT for Text Annotation. In 2023 International Conference on Machine Learning and Applications (ICMLA) (pp. 602-609). IEEE.
    \item Liu, Q., Hyland, S. L., Bannur, S., Bouzid, K., Castro, D. C., Wetscherek, M. T., Tinn, R., Sharma, H., Pérez-García, F., Schwaighofer, A., Rajpurkar, P., Khanna, S. T., Poon, H., Usuyama, N., Thieme, A., Nori, A., Lungren, M. P., Oktay, O., \& Alvarez-Valle, J. (2023). Exploring the Boundaries of GPT-4 in Radiology. In Proceedings of the 2023 Conference on Empirical Methods in Natural Language Processing (pp. 14414-14445). Association for Computational Linguistics.
    \item Ramteke, P. S., \& Khandelwal, S. (2023, November). Comparing Conventional Machine Learning and Large-Language Models for Human Stress Detection Using Social Media Posts. In 2023 2nd International Conference on Futuristic Technologies (INCOFT) (pp. 1-8). IEEE.
    \item Lewis, P., Perez, E., Piktus, A., Petroni, F., Karpukhin, V., Goyal, N., Küttler, H., Lewis, M., Yih, W., Rocktäschel, T., Riedel, S., \& Kiela, D. (2020). Retrieval-Augmented Generation for Knowledge-Intensive NLP Tasks. In Advances in Neural Information Processing Systems 34 (NeurIPS 2020). Vancouver, Canada.
    \item Robertson, S., \& Zaragoza, H. (2009). The probabilistic relevance framework: BM25 and beyond. Foundations and Trends® in Information Retrieval, 3(4), 333-389.
    \item Kim, S., Kim, K., \& Jo, C. W. (2024). Accuracy of a large language model in distinguishing anti-and pro-vaccination messages on social media: The case of human papillomavirus vaccination. Preventive Medicine Reports, 42, 102723.
    \item Gu, X., Chen, X., Lu, P., Li, Z., Du, Y., \& Li, X. (2024). AGCVT-prompt for sentiment classification: Automatically generating chain of thought and verbalizer in prompt learning. Engineering Applications of Artificial Intelligence, 132, 107907.
    \item Raja, H., Munawar, A., Mylonas, N., Delsoz, M., Madadi, Y., Elahi, M., Hassan, A., Abu Serhan, H., Inam, O., Hernandez, L., Chen, H., Tran, S., Munir, W., Abd-Alrazaq, A., \& Yousefi, S. (2024). Automated Category and Trend Analysis of Scientific Articles on Ophthalmology Using Large Language Models: Development and Usability Study. JMIR Formative Research, 8, Article e52462. https://doi.org/10.2196/52462.
    \item Guo, E., Gupta, M., Deng, J., Park, Y. J., Paget, M., \& Naugler, C. (2024). Automated paper screening for clinical reviews using large language models: Data analysis study. Journal of Medical Internet Research, 26, e48996.
    \item Yang, J., Walker, K. C., Bekar-Cesaretli, A. A., Hao, B., Bhadelia, N., Joseph-McCarthy, D., \& Paschalidis, I. C. (2024). Automating biomedical literature review for rapid drug discovery: Leveraging GPT-4 to expedite pandemic response. International Journal of Medical Informatics, 105500.
    \item Change, C. H., Lucas, M. M., Lu-Yao, G., \& Yang, C. C. (2024, June). Classifying Cancer Stage with Open-Source Clinical Large Language Models. In 2024 IEEE 12th International Conference on Healthcare Informatics (ICHI) (pp. 76-82). IEEE.
    \item Carneros-Prado, D., Villa, L., Johnson, E., Dobrescu, C. C., Barragán, A., \& García-Martínez, B. (2023, November). Comparative study of large language models as emotion and sentiment analysis systems: A case-specific analysis of GPT vs. IBM Watson. In International Conference on Ubiquitous Computing and Ambient Intelligence (pp. 229-239). Cham: Springer Nature Switzerland.
    \item Peng, C., Yang, X., Chen, A., Yu, Z., Smith, K. E., Costa, A. B., Flores, M. G., Bian, J., \& Wu, Y. (2024). Generative large language models are all-purpose text analytics engines: Text-to-text learning is all your need. Journal of the American Medical Informatics Association, Article ocae078.
    \item Sivarajkumar, S., \& Wang, Y. (2022). HealthPrompt: a zero-shot learning paradigm for clinical natural language processing. In AMIA Annual Symposium Proceedings (Vol. 2022, p. 972). American Medical Informatics Association.
    \item Wang, Y., Wang, Y., Peng, Z., Zhang, F., Zhou, L., \& Yang, F. (2023). Medical text classification based on the discriminative pre-training model and prompt-tuning. Digital Health, 9, 20552076231193213.
    \item Xu, X., Yao, B., Dong, Y., Gabriel, S., Yu, H., Hendler, J., Ghassemi, M., Dey, A. K., \& Wang, D. (2024). Mental-llm: Leveraging large language models for mental health prediction via online text data. Proceedings of the ACM on Interactive, Mobile, Wearable and Ubiquitous Technologies, 8(1), 1-32.
    \item Williams, C. Y., Zack, T., Miao, B. Y., Sushil, M., Wang, M., Kornblith, A. E., \& Butte, A. J. (2024). Use of a large language model to assess clinical acuity of adults in the emergency department. JAMA Network Open, 7(5), e248895-e248895.
    \item Alsentzer, E., Rasmussen, M. J., Fontoura, R., Cull, A. L., Beaulieu-Jones, B., Gray, K. J., Bates, D. W., \& Kovacheva, V. P. (2023). Zero-shot interpretable phenotyping of postpartum hemorrhage using large language models. NPJ Digital Medicine, 6(1), 212. https://doi.org/10.1038/s41746-023-00794-6.
    \item Sarkar, S., Feng, D., \& Santu, S. K. K. (2023, December). Zero-shot multi-label topic inference with sentence encoders and llms. In Proceedings of the 2023 Conference on Empirical Methods in Natural Language Processing (pp. 16218-16233).
    \item Yuan, J., Tang, R., Jiang, X., \& Hu, X. (2023). Large language models for healthcare data augmentation: An example on patient-trial matching. In AMIA Annual Symposium Proceedings (Vol. 2023, p. 1324). American Medical Informatics Association.
    \item Guevara, M., Chen, S., Thomas, S., Chaunzwa, T. L., Franco, I., Kann, B. H., Moningi, S., Qian, J. M., Goldstein, M., Harper, S., Aerts, H. J. W. L., Catalano, P. J., Savova, G. K., Mak, R. H., \& Bitterman, D. S. (2024). Large language models to identify social determinants of health in electronic health records. NPJ Digital Medicine, 7(1), 6. https://doi.org/10.1038/s41746-023-00970-0.
    \item Wu, C., Fang, W., Dai, F., \& Yin, H. (2023, October). A Model Ensemble Approach with LLM for Chinese Text Classification. In China Health Information Processing Conference (pp. 214-230). Singapore: Springer Nature Singapore.
    \item Yang, X., Chen, A., PourNejatian, N., Shin, H. C., Smith, K. E., Parisien, C., Compas, C., Martin, C., Costa, A. B., Flores, M. G., Zhang, Y., Magoc, T., Harle, C. A., Lipori, G., Mitchell, D. A., Hogan, W. R., Shenkman, E. A., Bian, J., \& Wu, Y. (2022). A large language model for electronic health records. npj Digital Medicine, 5(1), 194. https://doi.org/10.1038/s41746-022-00742-2.
    \item McMaster, C., Chan, J., Liew, D. F., Su, E., Frauman, A. G., Chapman, W. W., \& Pires, D. E. (2023). Developing a deep learning natural language processing algorithm for automated reporting of adverse drug reactions. Journal of biomedical informatics, 137, 104265.
    \item Li, Y., Wehbe, R. M., Ahmad, F. S., Wang, H., \& Luo, Y. (2023b). A comparative study of pretrained language models for long clinical text. Journal of the American Medical Informatics Association, 30(2), 340-347. https://doi.org/10.1093/jamia/ocac225
    \item Li, R., Wang, X., \& Yu, H. (2024b). LlamaCare: An Instruction Fine-Tuned Large Language Model for Clinical NLP. In Proceedings of the 2024 Joint International Conference on Computational Linguistics, Language Resources and Evaluation (LREC-COLING 2024) (pp. 10632-10641).
    \item Lehman, E., Hernandez, E., Mahajan, D., Wulff, J., Smith, M. J., Ziegler, Z., Nadler, D., Szolovits, P., Johnson, A., \& Alsentzer, E. (2023). Do We Still Need Clinical Language Models? Proceedings of Machine Learning Research, 209
    \item Schneider, E. T. R., de Souza, J. V. A., Gumiel, Y. B., Moro, C., \& Paraiso, E. C. (2021). A GPT-2 Language Model for Biomedical Texts in Portuguese. In 2021 IEEE 34th International Symposium on Computer-Based Medical Systems (CBMS) (pp. 474-479). IEEE. https://doi.org/10.1109/CBMS52027.2021.00056
    \item Gretz, S., Halfon, A., Shnayderman, I., Toledo-Ronen, O., Spector, A., Dankin, L., Katsis, Y., Arviv, O., Katz, Y., Slonim, N., \& Ein-Dor, L. (2023). Zero-shot Topical Text Classification with LLMs - an Experimental Study. In Findings of the Association for Computational Linguistics: EMNLP 2023 (pp. 9647–9676). Association for Computational Linguistics.
    \item Savage, T., Wang, J., \& Shieh, L. (2023). A large language model screening tool to target patients for best practice alerts: development and validation. JMIR Medical Informatics, 11, e49886.
    \item Shiju, A., \& He, Z. (2022, June). Classifying drug ratings using user reviews with transformer-based language models. In 2022 IEEE 10th International Conference on Healthcare Informatics (ICHI) (pp. 163-169). IEEE.
    \item Xie, K., Ojemann, W.K., Gallagher, R.S., Shinohara, R.T., Lucas, A., Hill, C.E., Hamilton, R.H., Johnson, K.B., Roth, D., Litt, B., \& Ellis, C.A. (2024). Disparities in seizure outcomes revealed by large language models. Journal of the American Medical Informatics Association, 31(6).
    \item Chen, Q., Du, J., Allot, A., \& Lu, Z. (2022). LitMC-BERT: transformer-based multi-label classification of biomedical literature with an application on COVID-19 literature curation. IEEE/ACM transactions on computational biology and bioinformatics, 19(5), 2584-2595.
    \item Bumgardner, V. C., Mullen, A., Armstrong, S. E., Hickey, C., Marek, V., \& Talbert, J. (2024). Local Large Language Models for Complex Structured Tasks. AMIA Summits on Translational Science Proceedings, 2024, 105.
    \item Cui, Y., Han, L., \& Nenadic, G. (2023, July). Medtem2. 0: Prompt-based temporal classification of treatment events from discharge summaries. In Proceedings of the 61st Annual Meeting of the Association for Computational Linguistics (Volume 4: Student Research Workshop) (pp. 160-183).
    \item Van Ostaeyen, S., De Langhe, L., De Clercq, O., Embo, M., Schellens, T., \& Valcke, M. (2023). Automating the identification of feedback quality criteria and the CanMEDS roles in written feedback comments using natural language processing. Perspectives on Medical Education, 12(1), 540.
    \item Jiang, Y., Qiu, R., Zhang, Y., \& Zhang, P. F. (2023, November). Balanced and explainable social media analysis for public health with large language models. In Australasian Database Conference (pp. 73-86). Cham: Springer Nature Switzerland.
    \item Ge, C., Ling, H., Quan, F., \& Zeng, J. (2023, October). Chinese Diabetes Question Classification Using Large Language Models and Transfer Learning. In China Health Information Processing Conference (pp. 205-213). Singapore: Springer Nature Singapore.
    \item Ren, Y., Wu, D., Khurana, A., Mastorakos, G., Fu, S., Zong, N., Fan, J., Liu, H., \& Huang, M. (2023). Classification of Patient Portal Messages with BERT-based Language Models. In 2023 IEEE 11th International Conference on Healthcare Informatics (ICHI) (pp. 176-182). IEEE.
    \item Wang, H., Gao, C., Dantona, C., Hull, B., \& Sun, J. (2024). DRG-LLaMA: tuning LLaMA model to predict diagnosis-related group for hospitalized patients. npj Digital Medicine, 7(1), 16.
    \item Chen, S., Li, Y., Lu, S., Van, H., Aerts, H. J., Savova, G. K., \& Bitterman, D. S. (2024b). Evaluating the ChatGPT family of models for biomedical reasoning and classification. Journal of the American Medical Informatics Association, 31(4), 940-948.
    \item Bressem, K. K., Adams, L. C., Gaudin, R. A., Tröltzsch, D., Hamm, B., Makowski, M. R., Schüle, C.-Y., Vahldiek, J. L., \& Niehues, S. M. (2020). Highly accurate classification of chest radiographic reports using a deep learning natural language model pre-trained on 3.8 million text reports. Bioinformatics, 36(21), 5255-5261.
    \item Tan, R. S. Y. C., Lin, Q., Low, G. H., Lin, R., Goh, T. C., Chang, C. C. E., Lee, F. F., Chan, W. Y., Tan, W. C., Tey, H. J., Leong, F. L., Tan, H. Q., Nei, W. L., Chay, W. Y., Tai, D. W. M., Lai, G. G. Y., Cheng, L. T., Wong, F. Y., Chua, M. C. H., Chua, M. L. K., Tan, D. S. W., Thng, C. H., Tan, I. B. H., \& Ng, H. T. (2023). Inferring cancer disease response from radiology reports using large language models with data augmentation and prompting. Journal of the American Medical Informatics Association, 30(10), 1657-1664.
    \item Silverman, A. L., Sushil, M., Bhasuran, B., Ludwig, D., Buchanan, J., Racz, R., Parakala, M., El-Kamary, S., Ahima, O., Belov, A., Choi, L., Billings, M., Li, Y., Habal, N., Liu, Q., Tiwari, J., Butte, A. J., \& Rudrapatna, V. A. (2024). Algorithmic Identification of Treatment‐Emergent Adverse Events From Clinical Notes Using Large Language Models: A Pilot Study in Inflammatory Bowel Disease. Clinical Pharmacology \& Therapeutics, 115(6), 1391-1399.
    \item Kementchedjhieva, Y., \& Chalkidis, I. (2023). An exploration of encoder-decoder approaches to multi-label classification for legal and biomedical text. In Findings of the Association for Computational Linguistics: ACL 2023 (pp. 5828-5843). Association for Computational Linguistics.
    \item Chen, Q., Sun, H., Liu, H., Jiang, Y., Ran, T., Jin, X., Xiao, X., Lin, Z., Chen, H., \& Niu, Z. (2023). An extensive benchmark study on biomedical text generation and mining with ChatGPT. Bioinformatics, 39(9), btad557. https://doi.org/10.1093/bioinformatics/btad557.
    \item Yogarajan, V., Pfahringer, B., Smith, T., \& Montiel, J. (2022a). Concatenating BioMed-Transformers to Tackle Long Medical Documents and to Improve the Prediction of Tail-End Labels. In International Conference on Artificial Neural Networks (pp. 209-221). Cham: Springer Nature Switzerland.
    \item Bețianu, M., Mălan, A., Aldinucci, M., Birke, R., \& Chen, L. (2024, April). DALLMi: Domain Adaption for LLM-Based Multi-label Classifier. In Pacific-Asia Conference on Knowledge Discovery and Data Mining (pp. 277-289). Singapore: Springer Nature Singapore.
    \item Farruque, N., Goebel, R., Sivapalan, S., \& Zaïane, O. R. (2024). Depression symptoms modelling from social media text: an LLM driven semi-supervised learning approach. Language Resources and Evaluation, 1-29.
    \item Pan, D., Zheng, X., Liu, W., Li, M., Ma, M., Zhou, Y., Yang, L., \& Wang, P. (2020). Multi-label classification for clinical text with feature-level attention. In 2020 IEEE 6th International Conference on Big Data Security on Cloud (BigDataSecurity), IEEE International Conference on High Performance and Smart Computing (HPSC) and IEEE International Conference on Intelligent Data and Security (IDS) (pp. 186-191). IEEE.
    \item Bansal, P., Das, S., Rai, V., \& Kumari, S. (2023). Multi-label Classification of Covid-19 Vaccine Tweet.
    \item Chaichulee, S., Promchai, C., Kaewkomon, T., Kongkamol, C., Ingviya, T., \& Sangsupawanich, P. (2022). Multi-label classification of symptom terms from free-text bilingual adverse drug reaction reports using natural language processing. PLoS One, 17(8), e0270595.
    \item Uslu, E. E., Sezer, E., \& Guven, Z. A. (2024). NLP-Powered Insights: A Comparative Analysis for Multi-Labeling Classification with MIMIC-CXR Dataset. IEEE Access.
    \item Blinov, P., Avetisian, M., Kokh, V., Umerenkov, D., \& Tuzhilin, A. (2020). Predicting clinical diagnosis from patients electronic health records using BERT-based neural networks. In Artificial Intelligence in Medicine: 18th International Conference on Artificial Intelligence in Medicine, AIME 2020, Minneapolis, MN, USA, August 25–28, 2020, Proceedings 18 (pp. 111-121). Springer International Publishing.
    \item Yogarajan, V., Montiel, J., Smith, T., \& Pfahringer, B. (2022b). Predicting COVID-19 Patient Shielding: A Comprehensive Study. In Australasian Joint Conference on Artificial Intelligence (pp. 332-343). Cham: Springer International Publishing.
    \item Qi, Z., Tan, X., Qu, C., Xu, Y., \& Qi, Y. (2023, July). Safer: a robust and efficient framework for fine-tuning bert-based classifier with noisy labels. In Proceedings of the 61st Annual Meeting of the Association for Computational Linguistics (Volume 5: Industry Track) (pp. 390-403).
    \item Ciobotaru, A., \& Dinu, L. P. (2023). SART \& COVIDSentiRo: Datasets for Sentiment Analysis Applied to Analyzing COVID-19 Vaccination Perception in Romanian Tweets. Procedia Computer Science, 225, 1331-1339.
    \item Luo, L., Ning, J., Zhao, Y., Wang, Z., Ding, Z., Chen, P., Fu, W., Han, Q., Xu, G., Qiu, Y., Pan, D., Li, J., Li, H., Feng, W., Tu, S., Liu, Y., Yang, Z., Wang, J., Sun, Y., \& Lin, H. (2024). Taiyi: A Bilingual Fine-Tuned Large Language Model for Diverse Biomedical Tasks. Journal of the American Medical Informatics Association.
    \item Kersting, J., Maoro, F., \& Geierhos, M. (2023). Towards comparable ratings: Exploring bias in German physician reviews. Data \& Knowledge Engineering, 148, 102235.
    \item Yogarajan, V., Montiel, J., Smith, T., \& Pfahringer, B. (2021, June). Transformers for multi-label classification of medical text: an empirical comparison. In International Conference on Artificial Intelligence in Medicine (pp. 114-123). Cham: Springer International Publishing.
    \item Vayena, E., Blasimme, A., \& Cohen, I. G. (2018). Machine learning in medicine: addressing ethical challenges. PLoS medicine, 15(11), e1002689.
    \item Liu, P., Yuan, W., Fu, J., Jiang, Z., Hayashi, H., \& Neubig, G. (2023). Pre-train, prompt, and predict: A systematic survey of prompting methods in natural language processing. ACM Computing Surveys, 55(9), 1-35.
    \item Schick, T., \& Schütze, H. (2020). Exploiting cloze questions for few shot text classification and natural language inference. arXiv preprint arXiv:2001.07676.
    \item Hu, E. J., Shen, Y., Wallis, P., Allen-Zhu, Z., Li, Y., Wang, S., Wang, L., \& Chen, W. (2021). LoRA: Low-rank adaptation of large language models. arXiv preprint arXiv:2106.09685.
    \item Lee, J., Yoon, W., Kim, S., Kim, D., Kim, S., So, C. H., \& Kang, J. (2020). BioBERT: a pre-trained biomedical language representation model for biomedical text mining. Bioinformatics, 36(4), 1234-1240.
    \item Alsentzer, E., Murphy, J. R., Boag, W., Weng, W. H., Jin, D., Naumann, T., \& McDermott, M. (2019). Publicly available clinical BERT embeddings. arXiv preprint arXiv:1904.03323.
    \item Lester, B., Al-Rfou, R., \& Constant, N. (2021). The power of scale for parameter-efficient prompt tuning. Proceedings of the 2021 Conference on Empirical Methods in Natural Language Processing (EMNLP), 3045-3061. https://doi.org/10.18653/v1/2021.emnlp-main.243
    \item Liu, H., Tam, D., Muqeeth, M., Mohta, J., Huang, T., Bansal, M., \& Raffel, C. A. (2022). Few-shot parameter-efficient fine-tuning is better and cheaper than in-context learning. Advances in Neural Information Processing Systems, 35, 1950-1965.
    \item Liu, L., Qu, Z., Chen, Z., Ding, Y., \& Xie, Y. (2021). Transformer acceleration with dynamic sparse attention. arXiv preprint arXiv:2110.11299.
    \item Shumailov, I., Shumaylov, Z., Zhao, Y., Papernot, N., Anderson, R., \& Gal, Y. (2024). AI models collapse when trained on recursively generated data. Nature, 631(8022), 755-759.
    \item Loftus, T. J., Ruppert, M. M., Shickel, B., Ozrazgat-Baslanti, T., Balch, J. A., Efron, P. A., Upchurch, G. R., Jr., Rashidi, P., Tignanelli, C., Bian, J., \& Bihorac, A. (2022). Federated learning for preserving data privacy in collaborative healthcare research. Digital Health, 8, Article 20552076221134455.
    \item Ficek, J., Wang, W., Chen, H., Dagne, G., \& Daley, E. (2021). Differential privacy in health research: A scoping review. Journal of the American Medical Informatics Association, 28(10), 2269-2276.
    \item Wu, X., Zhao, H., Zhu, Y., Shi, Y., Yang, F., Liu, T., Zhai, X., Yao, W., Li, J., Du, M., \& Liu, N. (2024). Usable XAI: 10 strategies towards exploiting explainability in the LLM era. arXiv preprint arXiv:2403.08946.
    \item Sakai, H., Farrag, A., Abubaker, S., AlRawashdeh, S., \& Won, D. (2022a). Healthcare fraud detection using data mining. In IISE Annual Conference. Proceedings (pp. 1-6). Institute of Industrial and Systems Engineers (IISE).
    \item Sakai, H., \& Lam, S. S. (2022b). Diabetic Patients Readmission Decision Support. In IISE Annual Conference. Proceedings (pp. 1-6). Institute of Industrial and Systems Engineers (IISE).
    \item Sakai, H., Farrag, A., Hendaileh, E., \& Lu, S. S. (2023b). Machine Learning Approaches for Stroke Classification. In IISE Annual Conference. Proceedings (pp. 1-6). Institute of Industrial and Systems Engineers (IISE).

\end{enumerate}

\end{document}